\documentclass[12pt]{article}	% The documentclass must be ``report''.

\usepackage{amsmath,amsthm,amsfonts,amscd} 
\usepackage{verbatim}      	% Allows quoting source with commands.
\usepackage{makeidx}       	% Package to make an index.
\usepackage{epsfig}         	% Allows inclusion of eps files.
\usepackage{url}		% Allows good typesetting of web URLs.
\usepackage{graphicx}
\usepackage[caption=false]{subfig}
\graphicspath{{Figures/}} 
\usepackage{amsfonts}
\usepackage{algorithm}
\usepackage{algorithmicx}
\usepackage{algpseudocode}
\usepackage{tikz}
\usepackage{float}
\usepackage{xcolor}
\usepackage{dsfont}

\theoremstyle{definition}

\theoremstyle{remark}

\newcommand\restr[2]{{% we make the whole thing an ordinary symbol
  \left.\kern-\nulldelimiterspace % automatically resize the bar with \right
  #1 % the function
  \vphantom{\big|} % pretend it's a little taller at normal size
  \right|_{#2} % this is the delimiter
  }}

\newcommand{\latexe}{{\LaTeX\kern.125em2%
                      \lower.5ex\hbox{$\varepsilon$}}}

\chardef\bslash=`\\	% \bslash makes a backslash (in tt fonts)
\makeatletter		% Starts section where @ is considered a letter
\def\square{\RIfM@\bgroup\else$\bgroup\aftergroup$\fi
  \vcenter{\hrule\hbox{\vrule\@height.6em\kern.6em\vrule}%
                                              \hrule}\egroup}
\makeatother		% Ends sections where @ is considered a letter.
\makeindex    % Make the index

\usepackage{amsmath,mathtools}
\DeclareMathOperator*{\argmax}{arg\,max}
\DeclareMathOperator*{\argmin}{arg\,min}

\newcommand{\beq}{\begin{equation} \begin{aligned}}
\newcommand{\eeq}{\end{aligned} \end{equation}}
\newcommand{\beqstar}{\begin{equation*} \begin{aligned}}
\newcommand{\eeqstar}{\end{aligned} \end{equation*}}

\newcommand{\map}{\phi} % signed distance function to surface of obstacle
\newcommand{\vis}{\psi} % visibility function
\newcommand{\graze}{\gamma} % 
\newcommand{\aux}{\alpha} % 
\newcommand{\shadow}{\xi}

\newcommand{\domain}{\Omega}
\newcommand{\free }{\Omega_{\text{free}}} % the set of free space
\newcommand{\obs }{\Omega_{\text{obs}}} % the set of free space

\newcommand{\visset }{\mathcal{V}} % the

\usepackage[mathscr]{eucal}

\newcommand{\xp}{{x_P}}
\newcommand{\xe}{{x_E}}
\renewcommand{\xp}{{P}}
\renewcommand{\xe}{{E}}
\newcommand{\bfxp}{{{\bf P}}}
\newcommand{\bfxe}{{{\bf E}}}
\newcommand{\xpinit}{{\xp^0}}
\newcommand{\xeinit}{{\xe^0}}
\newcommand{\val}{V}
\newcommand{\pcontrol}{{\sigma_P}}
\newcommand{\econtrol}{{\sigma_E}}
\newcommand{\admiss}{{\mathcal{A}}}
\newcommand{\cost}{\mathcal{J}}

\newcommand{\pspeed}{{f_P}}
\newcommand{\espeed}{{f_E}}
\newcommand{\failset}{\mathcal{T_\text{end}}}

\newcommand{\sgn}{{\text{sgn}}}
\newcommand{\parrival}[1][{}]{d_{\xp_{#1}}}
\newcommand{\earrival}[1][{}]{d_{\xe_{#1}}}
\newcommand{\pvalid}[1][{}]{A_{\xp_{#1}}}
\newcommand{\evalid}[1][{}]{A_{\xe_{#1}}}
\newcommand{\bfpvalid}[1][{}]{{\bf A_{\xp_{#1}}}}
\newcommand{\bfevalid}[1][{}]{{\bf A_{\xe_{#1}}}}
\newcommand{\tstar}[1][{}]{\tau_{\xe_{#1}}}
\newcommand{\bftstar}[1][{}]{{\bf \uptau_{\xe_{#1}}}}
\newcommand{\norm}{\alpha}

\newcommand{\trans}{T}
\newcommand{\mcts}{\text{MCTS}}

\newcommand{\prior}{{f}}

\newcommand{\bigo}{\mathcal{O}}
\usepackage{upgreek}

\usepackage{graphicx}
\usepackage{comment}
\usepackage{amsmath,amssymb} % define this before the line numbering.
\usepackage{color}

\usepackage{subfig}
\usepackage{xcolor}
\usepackage{paralist}
\usepackage[normalem]{ulem}   % just here for del command, remove later

\usepackage{siunitx}

\usepackage{dblfloatfix} % for two column float placement
\usepackage{balance}  % automatically even out the last page white space

\newcommand{\Colin}[1]{{\textcolor{orange}{ \textbf{cbm:} #1 }}}

\renewcommand{\Colin}[1]{}

\usepackage[doublespacing]{setspace}
\title{{\bf \Large{Visibility Optimization for Surveillance-Evasion Games \vspace{1em} }}}

\author{Louis Ly and Yen-Hsi Richard Tsai\\
\footnotesize{\tt{louisly@utexas.edu, ytsai@math.utexas.edu} }\\[3em]
Oden Institute for Computational Engineering and Sciences \\ 
The University of Texas at Austin} 
\date{}

\begin{document}
\maketitle

\begin{abstract}
We consider surveillance-evasion differential games, where a pursuer must try to constantly maintain visibility of a moving evader.  The pursuer loses as soon as the evader becomes occluded. Optimal controls for game can be formulated as a Hamilton-Jacobi-Isaac equation. We use an upwind scheme to compute the feedback value function, corresponding to the end-game time of the differential game. Although the value function enables optimal controls, it is prohibitively expensive to compute, even for a single pursuer and single evader on a small grid.  We consider a discrete variant of the surveillance-game.  We propose two locally optimal strategies based on the static value function for the surveillance-evasion game with multiple pursuers and evaders. We show that Monte Carlo tree search and self-play reinforcement learning can train a deep neural network to generate reasonable strategies for on-line game play. Given enough computational resources and offline training time, the proposed model can continue to improve its policies and efficiently scale to higher resolutions.
\end{abstract}

\section{Introduction}

We consider a multiplayer \emph{surveillance-evasion} game consisting of two
teams, the pursuers and the evaders. The pursuers must maintain line-of-sight
visibility of the evaders for as long as possible as they move through an
environment with obstacles. Meanwhile, the evaders aim to hide from the
pursuers as soon as possible. The game ends when the pursuers lose sight of the
evaders. We assume all players have perfect knowledge of the obstacles and the
game is closed-loop -- each player employs a feedback strategy, reacting
dynamically to the positions of all other players. 

%The game can be divided into two settings, open loop and closed loop.
%In closed loop, the controls are dynamic and change as the actions occur.
%In open loop, the players select their control at the beginning and cannot change the actions as the game progresses.
%These games are useful in applications such as autonomous target tracking and babysitting.
In section~\ref{sec:hji}, we consider the game in the context of Hamilton-Jacobi-Isaacs
(HJI) equations.  We propose a scheme to compute the value function, which,
informally, describes how "good" it is for each player to be in a specific
state. Then each player can pick the strategy that optimizes the value function
locally. Due to the principal of optimality, local optimization with respect to
the value function is globally optimal. This is because the value function
encodes information from all possible trajectories. As a result, the value
function is also very expensive to compute.

Section~\ref{sec:locally} discusses locally-optimal policies and
section~\ref{sec:learning-policy} presents search-based methods
to learn policies for the multiplayer version of the game.

\subsection{Related works}

The surveillance-evasion game is related to a popular class of games called
pursuit-evasion \cite{ho1965differential,isaacs1965differential}, where the
objective is for the pursuer to physically capture the evader.  Classical
problems take place in obstacle-free space with constraints on the players'
motion. Variants include the lion and man
\cite{karnad2009lion,sgall2001solution}, where both players have the same
maneuverability,  and the homicidal chauffeur \cite{merz1974homicidal}, where
one player drives a vehicle, which is faster, but has constrained mobility. 
Lewin et. al. \cite{lewin1975surveillance} considered a
game in an obstacle-free space, where the pursuer must keep the evader within a
detection circle.

Bardi et. al. \cite{bardi1999numerical} proposed a semi-Lagrangian scheme for
approximating the value function of the pursuit-evasion game as viscosity solution
to the Hamilton-Jacobi-Isaacs equation, in a bounded domain with no obstacles.
In general, these methods are very expensive, with complexity $\bigo(m^{kd})$ where
$k$ is the number of players and $d$ is the dimension. This is because the
value function, once computed, can provide the optimal controls for all
possible player positions. 
A class of methods try
to deal with the curse of dimensionality by solving for the solutions of
Hamilton-Jacobi equations at individual points in space and time. These methods
are causality-free; the solution at one point does not depend on solutions at
other points, making them conveniently parallelizable. They are efficient,
since one only solves for the value function locally, where it is needed,
rather than globally.  Chow et. al.
\cite{chow2017algorithm,chow2018algorithm,chow2019algorithm} use the Hopf-Lax
formula to efficiently solve Hamilton-Jacobi equations for a class of
Hamiltonians.  Sparse grid characteristics, due to Kang et. al. \cite{kang2017mitigating}, is another
causality-free method which finds the solution by solving a boundary value
problem for each point.  Unfortunately, these methods do not apply to domains
with obstacles since they cannot handle boundary conditions.  

% Sparse grid characteristics method requires convexity assumptions on the Hamiltonian.  Also, they use their method to train neural network to do inference efficiently.  

The visibility-based pursuit-evasion game, introduced by Suzuki. et. al
\cite{suzuki1992searching}, is a version where the
pursuer(s) must compute the shortest path to find all hidden evaders in a
cluttered environment, or report it is not possible. The number of evaders is
unknown and their speed is unbounded.  Guibas et. al.
\cite{guibas1997visibility} proposed a graph-based method for polygonal
environments.  Other settings include multiple pursuers
\cite{stiffler2014complete}, bounded speeds \cite{tovar2008visibility},
unknown, piecewise-smooth planar environments \cite{sachs2004visibility}, and
simply-connected two-dimensional curved environments
\cite{lavalle2001visibility}.

The surveillance-evasion game has been studied previously
in the literature. 
LaValle et. al. \cite{lavalle1997motion} use dynamic programming to compute
optimal trajectories for the pursuer, assuming a known evader trajectory. For
the case of an unpredictable evader, they suggest a local heuristic: maximize
the probably of visibility of the evader at the next time step.
They also mention, but do not implement, an idea to locally maximize the evader's time to occlusion.

Bhattacharya et. al. \cite{bhattacharya2008approximation,bhattacharya2011cell}
used geometric arguments to partition the environment into several regions
based on the outcome of the game.  
%evader can successfully escape, for environments with polygonal and smooth convex obstacles.  %They do not suggest controls.
In \cite{bhattacharya2009existence,zou2018optimal}, they use geometry and optimal control to compute
optimal trajectories for a single pursuer and single evader near the corners of a polygon.
The controls are then extended to the whole domain containing polygonal obstacles by partitioning based on the corners
\cite{zou2016optimal},
for the finite-horizon tracking problem \cite{zou2016visibility},
and for multiple players by allocating a pursuer for each evader via the Hungrarian matching algorithm
\cite{zhang2016multi}.
%extends to multiple players by allocating a purser to each evader through a Hungrarian matching algorithm using a designed risk function.
%In \cite{zou2016optimal}, they consider the optimal controls around a corner, and extend to general polygonal environments by designing pursuit partions, ways to weight the strategies around each corner based on distance/time.

Takei et. al. \cite{takei2014efficient} proposed an efficient
algorithm for computing the static value function corresponding to the open loop game,
where each player moves according to a fixed strategy determined at initial time.
Their open loop game is conservative towards the pursuer, since the evader can
optimally counter any of the pursuer's strategies. As a consequence, the game
is guaranteed to end in finite time, as long as the domain is not
star-shaped. %The computational cost is $\bigo(m^{2d})$ in $d$ dimensions, where $m$ is the size of the grid in one dimension.  The algorithm is parallelizable.
In contrast, a closed loop game allows players to react dynamically to each other's actions.

In \cite{cartee2019time,gilles2019evasive,takei2015optimal}, the authors
propose optimal paths for an evader to reach a target destination, while
minimizing exposure to an observer.  In \cite{gilles2019evasive}, the observer
is stationary.  In \cite{cartee2019time}, the observer moves according to a
fixed trajectory.  In \cite{takei2015optimal}, the evader can tolerate brief
moments of exposure so long as the consecutive exposure time does not exceed a
given threshold.  In all three cases, the observer's controls are restricted to
choosing from a known distribution of trajectories; they are not allowed to
move freely.% within the domain.

Bharadwaj et. al. \cite{bharadwaj2018synthesis} use reactive synthesis to
determine the pursuer's controls for the surveillance-evasion game on a discrete
grid. They propose a method of \emph{belief abstraction} to coarsen the state
space and only refine as needed.  The method is quadratic in the number of
states: $\bigo(m^{2kd})$ for $k$ players.  While it is more computationally
expensive than the Hamilton-Jacobi based methods, it is more flexible in being
able to handle a wider class of temporal surveillance objectives, such as
maintaining visibility at all times, maintaining a bound on the spatial
uncertainty of the evader, or guaranteeing visibility of the evader infinitely
often.

Recently, Silver et. al developed the AlphaGoZero and AlphaZero programs that
excel at playing Go, Chess, and Shogi, without using any prior knowledge of the
games besides the rules \cite{silver2017mastering,silver2017mastering2}. They
use Monte Carlo tree search, deep neural networks and self-play reinforcement
learning to become competitive with the world's top professional players.

%Inspired by their work, we apply Monte Carlo tree search and self-play reinforcement learning to the surveillance-evasion games.
%These games are symmetric, since the goals of the players are identical. 
%In our games, the goals for the pursuer and evader differ.

\subsection*{Contributions}
We use a Godunov upwind scheme to compute the value function for the
closed loop surveillance-evasion game with obstacles in two dimensions.  The
state space is four dimensional.  The value function allows us to compute the
optimal feedback controls for the pursuers and evaders. Unlike the static game
\cite{takei2014efficient}, it is possible for the pursuer to win. However, the
computation is $\bigo(m^{kd})$ where $k$ is the number of players and $d$ the
dimensions.

As the number of players grows, computing the value function becomes infeasible.
We propose locally optimal strategies for the multiplayer surveillance-evasion
game, based on the value function for the static game.  In addition, we propose
a deep neural network trained via self play and Monte Carlo tree search to
learn controls for the pursuer.  Unlike Go, Chess, and Shogi, the
surveillance-evasion game is not symmetric; the pursuers and evaders require
different tactics. We use the local strategies to help improve the
efficiency of self-play.

The neural network is trained offline on a class of environments.  Then,
during play time, the trained network can be used to play games efficiently on
previously unseen environments.  That is, at the expense of preprocessing time
and optimality, we present an algorithm which can run efficiently. While the
deviation from optimality may sound undesirable, it actually is reasonable.
Optimality assumes perfect actions and instant reactions.  It real
applications, noise and delays will perturb the system away from optimal
trajectories.  We show promising examples in 2D.  

%Lastly, we show application 
%surveillance-constrained visibility optimization problem.  There, in addition
%to a strict surveillance requirement, the agent must continually patrol the
%environment.

\section{Visibility level sets}
We review our representation of geometry and visibility.
All the functions described below can be computed efficiently in $\bigo(m^d)$, where
$m$ is the number of grid points in each of $d$ dimensions.

\subsection*{Level set functions}
Level set functions \cite{osher1988fronts,sethian1999level,osher2006level} are useful as an implicit
representation of geometry.  Let $\obs \subseteq \mathbb{R}^d$ be a closed set
of a finite number of connected components
representing the obstacle. Denote the occluder function $\map$ with the following properties:
\beq
\begin{cases}
\map(x) < 0 & x \in \obs \\
\map(x) = 0 & x \in \partial \obs \\
\map(x) > 0 & x \notin \obs \\
\end{cases}
\label{eq:occluder}
\eeq

The occluder function is not unique; notice that for any constant $c>0$, the function $c \map$
also satisfies (\ref{eq:occluder}). We use the signed distance function as the occluder function:
\beq
\map(x):= 
\begin{dcases}
-\inf_{y\in \partial \obs} \|x-y\|_2 & x\in \obs \\
\inf_{y\in \partial \obs} \|x-y\|_2 & x \notin \obs
\end{dcases}
\eeq
The signed distance function is a viscosity solution to the Eikonal equation:
\beq
|\nabla \map| &= 1 \\
\map(x) &= 0 \text{ for } x \in \partial \obs
\eeq

It can be computed, for example, using the fast sweeping method \cite{tsai2002rapid} or
the fast marching method \cite{tsitsiklis1995efficient}.

\subsection*{Visibility function}
Let $\free$ be the open set representing free space. Let $\visset_{x_0}$ be the set of points in $\free$
visible from $x_0\in\free$. We seek a function $\vis(x,x_0)$ with properties

\beq
\begin{cases}
\vis(x,y) > 0 & x \in \visset_{x_0} \\
\vis(x,y) = 0 & x \in \partial \visset_{x_0} \\
\vis(x,y) < 0 & x \notin \visset_{x_0} \\
\end{cases}
\label{eq:visibility}
\eeq

Define the visibility level set function $\vis$:
\beq
\vis(x,{x_0}) = \min_{r\in[0,1]} \map(x_0 + r(x-{x_0}))
\eeq

It can be computed efficiently using the fast sweeping method based on the PDE formulation described in \cite{tsai2004visibility}:

\beq
\nabla \vis \cdot \frac{x-{x_0}}{|x-{x_0}|} &= \min \Big\{ H(\vis-\map) \nabla \vis \cdot \frac{x-{x_0}}{|x-{x_0}|},0 \Big\} \\
\vis({x_0},{x_0}) &= \map({x_0})
\eeq
where $H$ is the characteristic function of $[0,\infty)$.

\subsection*{Shadow function}
\label{sec:shadow}
When dealing with visibility, it is useful to represent the shadow regions. 
The gradient of the occluder function $\nabla \map$ is perpendicular to the
level sets $\map$.  The dot product
of $\nabla \map$ and the viewing direction $(x_0-x)$ characterizes the cosine
of the grazing angle $\theta$ between obstacles and viewing direction. In particular,
$|\theta|<\pi/4$ for the portion of obstacles that are directly visible to $x_0$.

Define the grazing function:
\beq
\graze(x,x_0) &= (x_0-x)^T \cdot \nabla \map(x)\\
\eeq

By masking with the occluder function, we can characterize the portion of the
obstacle boundary that is not visible from the vantage point $x_0$.  Define the
auxiliary and auxiliary visibility functions: \beq \aux(x,x_0) &= \max \{
\map(x,x_0), \graze(x,x_0) \} \\ \tilde{\aux}(x,x_0) &= \min_{r\in[0,1]} \aux(x
+ r(x_0-x),x_0) \\ \eeq

By masking the auxilary visibility function with the obstacle, we arrive at the desired
shadow function \cite{takei2014efficient}:
\beq
\shadow(x,x_0) &= \max\{ \tilde{\aux}(x,x_0), -\map(x)\}
\eeq

\captionsetup[subfloat]{labelformat=empty}
\begin{figure}[hptb]
\centering
\subfloat[Occluder function $\map(\cdot)$]{\includegraphics[width=.35\textwidth]{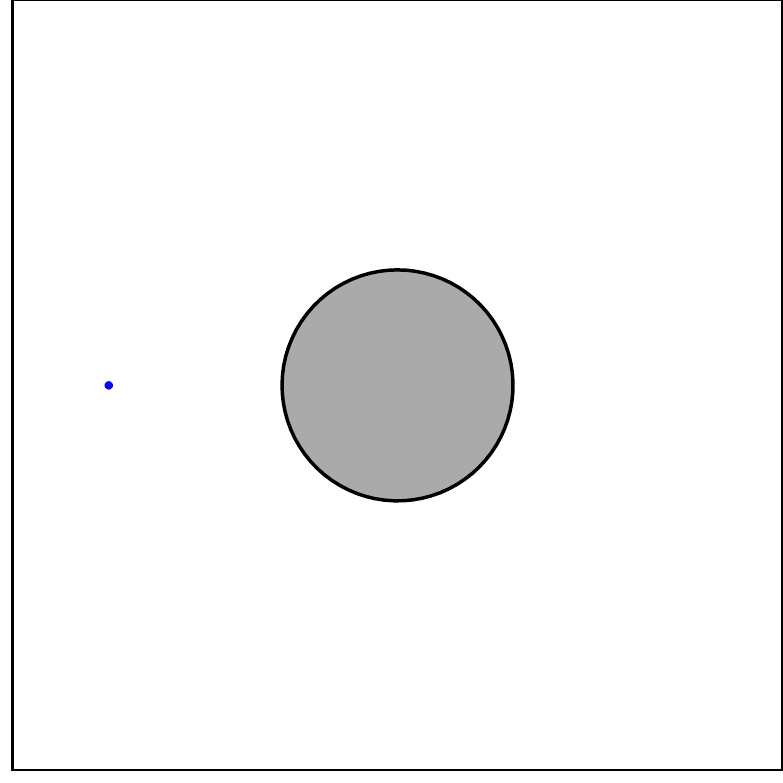}} \quad
\subfloat[Visibility function $\vis(\cdot)$]{\includegraphics[width=.35\textwidth]{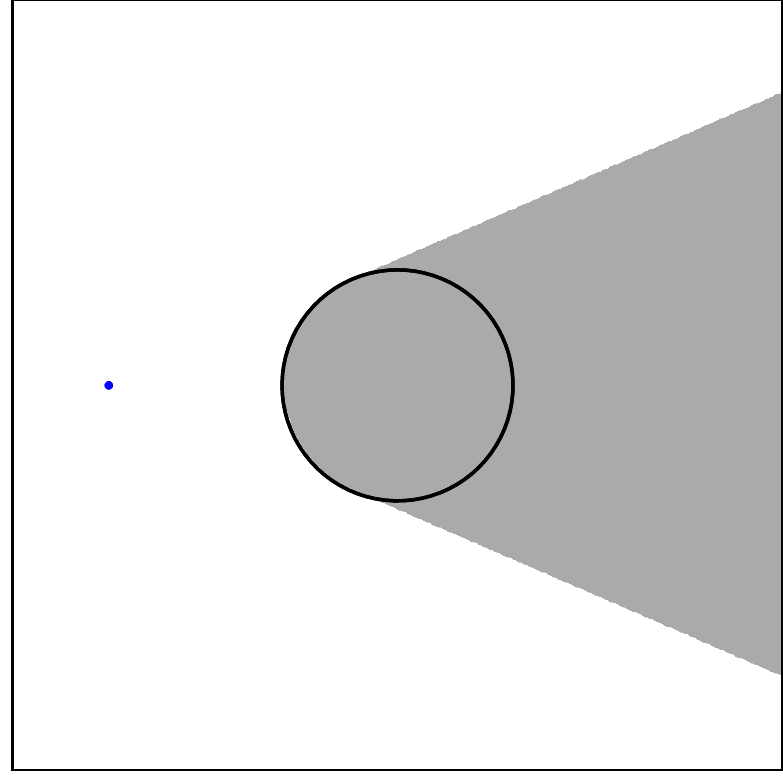}} \\
\subfloat[Grazing function $\graze(\cdot)$]{\includegraphics[width=.35\textwidth]{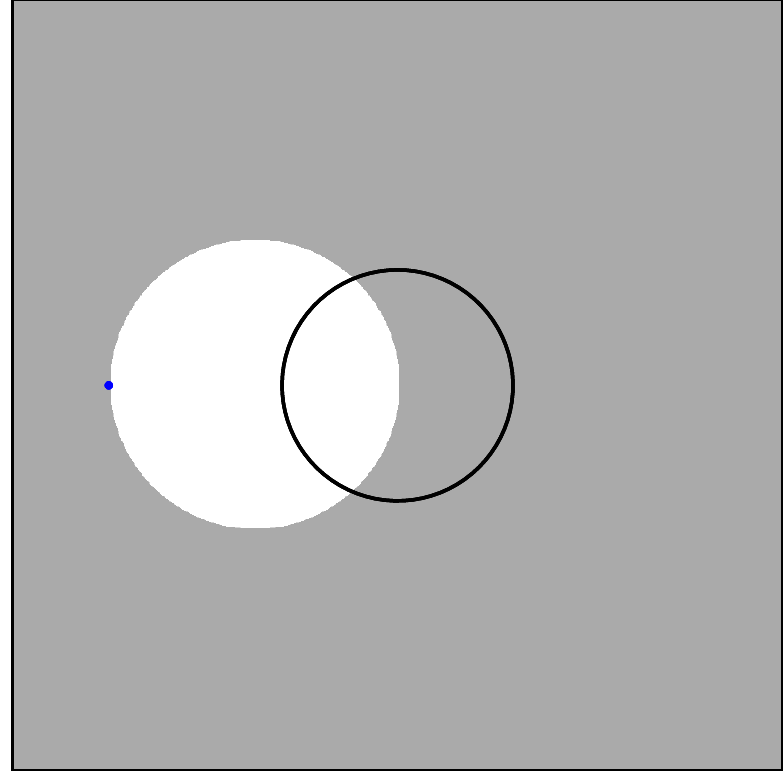}} \quad
\subfloat[Auxiliary function $\aux(\cdot)$]{\includegraphics[width=.35\textwidth]{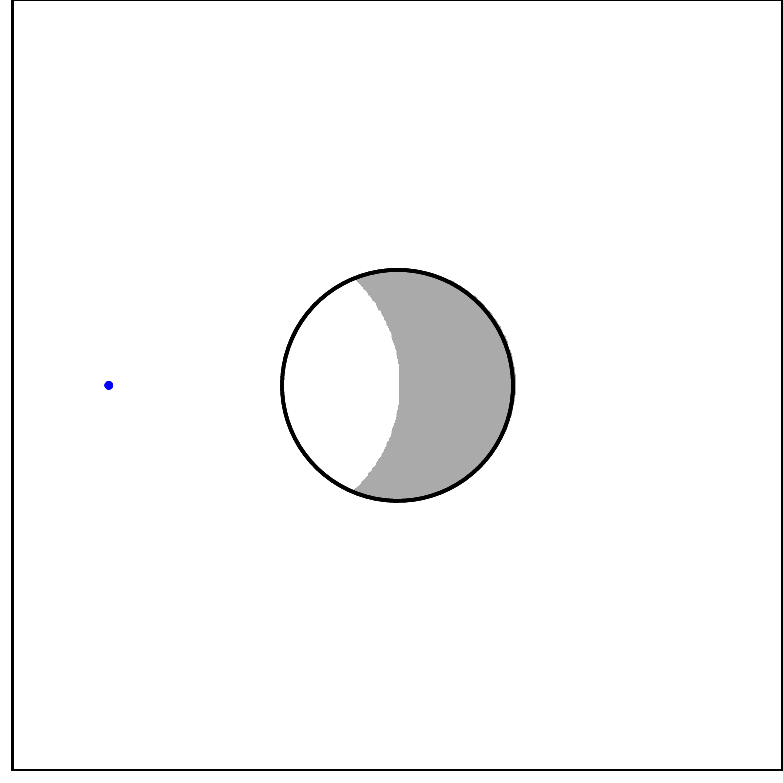}} \\
\subfloat[Auxiliary visibility function $\tilde{\aux}(\cdot)$]{\includegraphics[width=.35\textwidth]{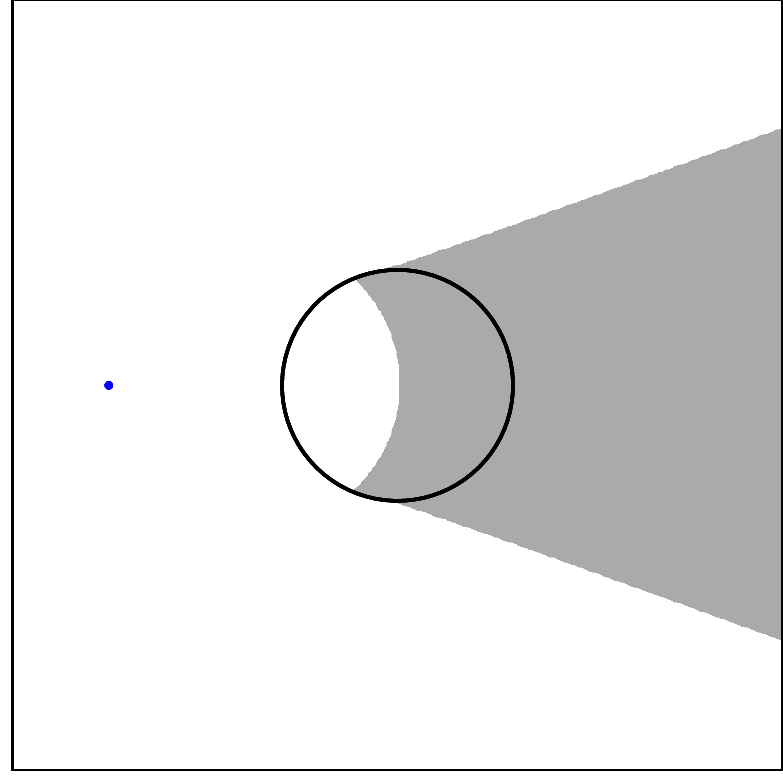}} \quad
\subfloat[Shadow function $\xi(\cdot)$]{\includegraphics[width=.35\textwidth]{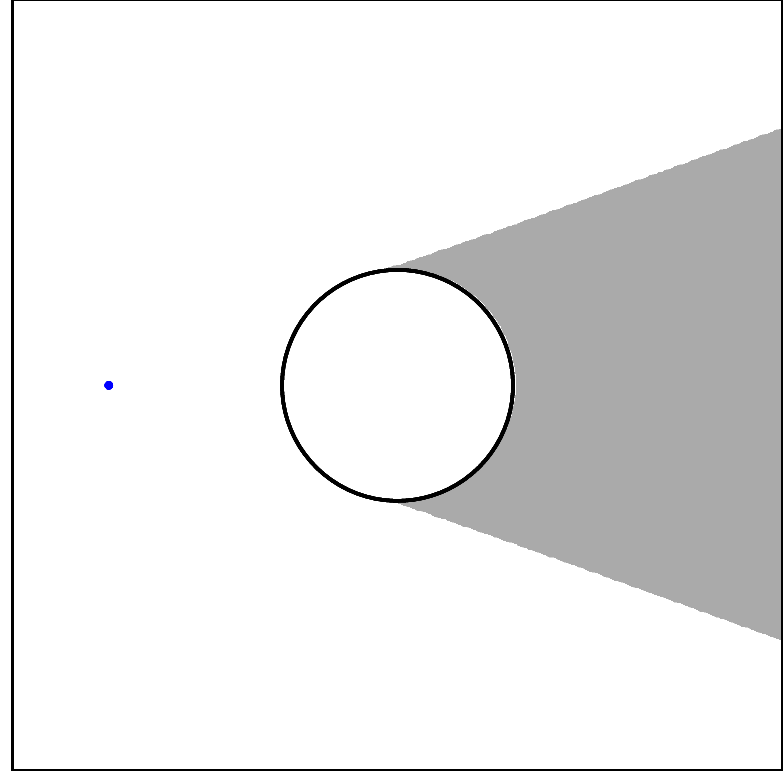}}
\caption{Level set functions from a vantage point $x_0$ (blue dot). Each function is negative in the shaded region. Obstacle boundary shown as black contour.} \label{fig:ls_notation}

\end{figure}

The difference between the shadow function and the visibility function is that
the shadow function excludes the obstacles.
Although 
$$\tilde{\shadow}(x,x_0) = \max\{-\map(x),\vis(x,x_0)\}$$ 
looks like a candidate shadow function, it 
is not correct. In particular 
$$\{ x|\tilde{\shadow}(x,x_0)=0\}$$ 
includes the portion of the obstacle boundary visible to $x_0$.

Figure~\ref{fig:ls_notation} summarizes the relevant level set functions used in this work.

\section{Value function from HJI equation}
\label{sec:hji}

Without loss of generality, we formulate the two player game, with a single pursuer and single evader.
The domain $\domain\subseteq \mathbb{R}^d$ consists of obstacles and free space:
$\domain=\obs\cup\free$.  Consider a pursuer and evader whose positions at a particular time instance
are given by $\xp,\xe:[0,\infty) \to \free$, respectively. 
Let $A:= S^{d-1}\cup \{{\bf 0}\}$ be the compact set of control values.
The feedback controls map the players' positions to a control value:
\beq\pcontrol,\econtrol\in\admiss:=\{\sigma:\free\times\free\to A \ | \ \sigma \text{ measurable}\},\eeq
where $\admiss$ is the set of admissible controls.
The players move with velocities $\pspeed,\espeed:
\domain \to [0,\infty)$ according to the dynamics

\beq
%\dot{\xp}(t) &= \pspeed(\xp(t)) \pcontrol(\xp(t),\xe(t))  \qquad &\dot{\xe}(t) &= \espeed(\xe(t)) \econtrol(\xp(t),\xe(t))\\
\dot{\xp}(t) &= \pspeed(\xp(t)) \pcontrol(\xp(t),\xe(t))  \qquad &\dot{\xe}(t) &= \espeed(\xe(t)) \econtrol(\xp(t),\xe(t))\\
\xp(0)&= \xpinit \qquad &\xe(0)&= \xeinit \\
\eeq

%Notice the controls dynamically respond to the current positions for each player. 
For clarity of notation, we will omit the dependence of the controls on
the players' positions.  For simplicity, we assume velocities are isotropic,
meaning they do not depend on the controls. In real-world scenarios, this may not
be the case.  For example, an airplane's dynamics might be constrained by its
momentum and turning radius.

As a slight relaxation, we consider the finite-horizon version of the game, where the pursuers
win if they can prolong the game past a time threshold $T$.
Let $\failset := \{(\xp(\cdot),\xe(\cdot)) | \shadow(\xp(\cdot),\xe(\cdot))\le 0\}$ be the end-game set of losing positions,
where $\shadow$ is the shadow function defined in section~\ref{sec:shadow}.
Define the payoff function 
\beq 
\cost[\xpinit,\xeinit,t,\pcontrol,\econtrol] :=
\inf\{0\le \tau \le t|(\xp(\tau),\xe(\tau)) \in \failset \},
\eeq
where $\cost[\xpinit,\xeinit,t,\pcontrol,\econtrol]:=t$ if the set $(\xp(\tau),\xe(\tau))\in\failset$ is empty.
The payoff is the minimum time-to-occlusion for given set
initial positions and controls.
Define the finite-horizon value function as:
\beq
\val(\xpinit,\xeinit,t) = \sup_{\pcontrol\in\admiss} \inf_{\econtrol\in\admiss} \cost[\xpinit,\xeinit,t,\pcontrol,\econtrol] 
\eeq

The value function describes the length of the game played to time $t$, starting from all pairs of initial positions, and assuming optimal controls.
We are interested in $\val(\xpinit,\xeinit,T)$ for a sufficiently large $T$, which characterizes
the set of initial positions from which the pursuers can maintain visibility of the evaders
for at least $T$ time units.
As $T\to \infty$, we recover the infinite-horizon value function.

By using the principle of optimality and Taylor expansion, one can derive
the Hamilton-Jacobi-Isaacs equation \cite{bardi2008optimal,crandall1983viscosity,evans1984differential}:
\beq
\val_t + \inf_{\econtrol\in A} \sup_{\pcontrol \in A} \{-\nabla_\xp \val \cdot \pcontrol - \nabla_\xe \val \cdot \econtrol\}
&= 1 \ ,  &\text{ on } \free\setminus\failset \\
\val(\xp,\xe,0) &= 0 \\
\val(\xp,\xe,t) &= 0 \ , & (\xp,\xe)\in\failset \\
\val(\xp,\xe,t) &= \infty \ , & \xp \text{ or }\xe \in \obs \label{eq:hji}
\eeq
It has been shown the value function is the viscosity solution \cite{bardi2008optimal,crandall1983viscosity,evans1984differential} to \eqref{eq:hji}.
For isotropic controls, this simplifies to the following Eikonal equation:
\beq
\val_t -\pspeed |\nabla_\xp \val| +\espeed |\nabla_\xe \val|  &= 1 \ ,  &\text{ on } \free\setminus\failset \\
\val(\xp,\xe,0) &= 0 \\
\val(\xp,\xe,t) &= 0 \ , & (\xp,\xe )\in \failset \\
\val(\xp,\xe,t) &= \infty \ , & \xp \text{ or }\xe \in \obs \label{eq:hji-eikonal}
\eeq
The optimal controls can be recovered by computing the gradient of the value function:
\begin{align} \label{eq:optcontrols}
\pcontrol &=  \frac{\nabla_\xp \val}{|\nabla_\xp \val|} \ , \qquad
\econtrol =  -\frac{\nabla_\xe \val}{|\nabla_\xe \val|} \ .
\end{align}
%Remark: note there are other variants of the value function where the limits of
%integration do not depend on the controls, such as finite-horizon or
%infinite-horizon optimal control problem.  The resulting PDE is similar, but
%perhaps without the boundary conditions.  See, for example, [CITE]. 
%\todo{Or the appendix}.
%(I could technically stick with a time dependent formulation with a large penalty, but there's no guarantee)

\subsection{Algorithm}
\iffalse
We solve (\ref{eq:hji-eikonal}) as the steady state of the time-dependent equation \cite{osher1993level}:
\beq
\val_t -\pspeed |\nabla_\xp \val| +\espeed |\nabla_\xe \val|  &= 1 \qquad &\text{ on } \free\setminus\failset \\
\val(\xp,\xe,t) &= 0 &\qquad (\xp,\xe )\in \failset \\
\val(\xp,\xe) &= \infty &\qquad \xp \text{ or }\xe \in \obs \label{eq:hji-eikonal-time}
\eeq
\fi

Following the ideas in \cite{tsai2003fast}, we discretize the gradient using upwind scheme as follows.
Let $\xp_{i,j},\xe_{k,l}\in\free$ be the discretized positions with grid
spacing $h$.  Denote $\val_{i,j,k,l}$ as the numerical solution to (\ref{eq:hji-eikonal}) for
initial positions $\xp_{i,j},\xe_{k,l}$.  We estimate the gradient using finite
difference. For clarity, we will only mark the relevant subscripts, e.g.
$\val_{i+1}:=\val_{i+1,j,k,l}$.

%\beq
%\partial_{P_x}\val &= \frac{1}{h} \max \Big( \max(\val_{i}-\val_{i-1},0), \min(\val_{i+1}-\val_{i},0) \Big), \\
%\eeq

%\beq
%\xp_{x-} &= \frac{1}{h} \Big(\val_{i,j,k,l} - \val_{i-1,j,k,l}\Big), \qquad  &\xe_{x-} = \frac{1}{h} \Big(\val_{i,j,k,l} - \val_{i,j,k-1,l}\Big),\\
%\xp_{x+} &= \frac{1}{h} \Big(\val_{i+1,j,k,l} - \val_{i,j,k,l}\Big), \qquad  &\xe_{x+} = \frac{1}{h} \Big(\val_{i,j,k+1,l} - \val_{i,j,k,l}\Big),\\
%\xp_{y-} &= \frac{1}{h} \Big(\val_{i,j,k,l} - \val_{i,j-1,k,l}\Big), \qquad  &\xe_{y-} = \frac{1}{h} \Big(\val_{i,j,k,l} - \val_{i,j,k,l-1}\Big),\\
%\xp_{y+} &= \frac{1}{h} \Big(\val_{i,j+1,k,l} - \val_{i,j,k,l}\Big), \qquad  &\xe_{y+} = \frac{1}{h} \Big(\val_{i,j,k,l+1} - \val_{i,j,k,l}\Big).
%\eeq

%\beq
%\xp_{-}^x &= \frac{1}{h} \Big(\val_{i} - \val_{i-1}\Big), \qquad  &\xe_{-}^x &= \frac{1}{h} \Big(\val_{k} - \val_{k-1}\Big),\\
%\xp_{+}^x &= \frac{1}{h} \Big(\val_{i+1} - \val_{i}\Big), \qquad  &\xe_{+}^x &= \frac{1}{h} \Big(\val_{k+1} - \val_{k}\Big),\\
%\xp_{-}^y &= \frac{1}{h} \Big(\val_{j} - \val_{j-1}\Big), \qquad  &\xe_{-}^y &= \frac{1}{h} \Big(\val_{l} - \val_{l-1}\Big),\\
%\xp_{+}^y &= \frac{1}{h} \Big(\val_{j+1} - \val_{j}\Big), \qquad  &\xe_{+}^y &= \frac{1}{h} \Big(\val_{l+1} - \val_{l}\Big).
%\eeq

\beq
\xp_{x^-} &:= \frac{1}{h} \Big(\val_{i} - \val_{i-1}\Big), \qquad  &\xe_{x^-} &:= \frac{1}{h} \Big(\val_{k} - \val_{k-1}\Big),\\
\xp_{x^+} &:= \frac{1}{h} \Big(\val_{i+1} - \val_{i}\Big), \qquad  &\xe_{x^+} &:= \frac{1}{h} \Big(\val_{k+1} - \val_{k}\Big),\\
\xp_{y^-} &:= \frac{1}{h} \Big(\val_{j} - \val_{j-1}\Big), \qquad  &\xe_{y^-} &:= \frac{1}{h} \Big(\val_{l} - \val_{l-1}\Big),\\
\xp_{y^+} &:= \frac{1}{h} \Big(\val_{j+1} - \val_{j}\Big), \qquad  &\xe_{y^+} &:= \frac{1}{h} \Big(\val_{l+1} - \val_{l}\Big).\\
\eeq

Let $a^- := -\min(0,a)$ and
$a^+ := \max(0,a)$. Define
\beq
\sgn\max(a,b) &:= 
\begin{cases}
a^+ & \text{ if } \max(a^+,b^-) = a^+ \\
-b^- & \text{ if } \max(a^+,b^-) = b^- \\
\end{cases} \\
\eeq

and

\beq
\partial \xp_x \val &= \sgn\max (\xp_{x^+}, \xp_{x^-} ) \\
\partial \xp_y \val &= \sgn\max (\xp_{y^+}, \xp_{y^-} ) \\
\partial \xe_x \val &= \sgn\max (\xe_{x^-}, \xe_{x^+} ) \\
\partial \xe_y \val &= \sgn\max (\xe_{y^-}, \xe_{y^+} ) \\
\eeq

Finally, the desired numerical gradients are
\beq
|\nabla_\xp \val| &= \Big( (\partial \xp_x \val )^2 + (\partial \xp_y \val)^2 \Big)^{1/2} \\
|\nabla_\xe \val| &= \Big( (\partial \xe_x \val )^2 + (\partial \xe_y \val)^2 \Big)^{1/2} \\
\eeq

Then we have a simple explicit scheme.
\beq
\val^{n+1} = \val^{n} + \Delta t (1 + \pspeed |\nabla_\xp \val| - \espeed |\nabla_\xe \val| ) 
\label{eq:hji-scheme}
\eeq
The CFL conditions dictate that the time step $\Delta t$ should be
\beq
\Delta t \le \frac{h}{16 \max(\pspeed,\espeed)}
\eeq

For a given environment, we precompute the value function by iteration until convergence.
During play time, we initialize $\xpinit,\xeinit$ and compute the optimal trajectories according to
(\ref{eq:optcontrols}) using $\Delta t$ time increments.

\subsubsection{Boundary conditions}
The obstacles appear in the HJI equation as boundary conditions.
However, direct numerical implementation leads to artifacts near the obstacles.
Instead, we model the obstacles by setting the velocities to be small inside obstacles.
We regularize the velocities by adding a smooth transition \cite{takei2014efficient}:

\begin{align}
v_\epsilon(x) &=
\begin{cases}
v(x) & \map(x)>0 \\
v_\text{min} + \frac{v(x)-v_\text{min}}{2} \Big[ \cos\Big(\frac{\phi(x)\pi}{2\epsilon}\Big)+1 \Big]  & \map(x)\in[-2\epsilon,0]\\
v_\text{min} & \map(x) < -2\epsilon
\end{cases}
\end{align}
where $\map$ is the signed distance function to the obstacle boundaries.
In the numerical experiments, we use $\epsilon=16\Delta x$ and $v_\text{min} = 1/100$.

\subsection{Numerical results}

\subsubsection*{Stationary pursuer}
We verify that the scheme converges numerically for the case in which the pursuer is stationary. 
When $\pspeed=0$, the HJI equation \eqref{eq:hji-eikonal} becomes the time-dependent Eikonal equation:
\beq
\val_t + \espeed |\nabla_\xe \val|  &= 1 \qquad &\text{ on } \free\setminus\failset \\
\val(\xp,\xe,0) &= 0 \\
\val(\xp,\xe,t) &= 0 &\qquad (\xp,\xe )\in \failset \label{eq:hji-stationary}
\eeq
In particular, for sufficiently large $t$, the value function reaches a steady state, and satisfies the Eikonal equation:
\beq
\espeed |\nabla_\xe \val|  &= 1 \qquad &\text{ on } \free\setminus\failset \\
\val(\xp,\xe) &= 0 &\qquad (\xp,\xe )\in \failset \label{eq:hji-stationary-steady}
\eeq
For this special case, the exact solution is known; the solution corresponds to the evader's travel time to
the end-game set $\failset$.
Also, this case effectively reduces the computational cost
from $\bigo(m^4)$ to $\bigo(m^2)$, so that we can reasonably compute solutions at higher resolutions.

We consider a $\domain=[0,1)\times[0,1)$ with a single circular obstacle of radius $0.15$ centered at $(1/2,1/2)$.
The pursuer is stationary at $\xp_0=(1/8,1/2)$. 
We use $\Delta t = \Delta x/20$ and iterate until the solution no longer changes in the $L_1$ sense, using a tolerance of $10^{-5}$.

\begin{figure}[hptb]
      \centering
      \includegraphics[width=.45\textwidth]{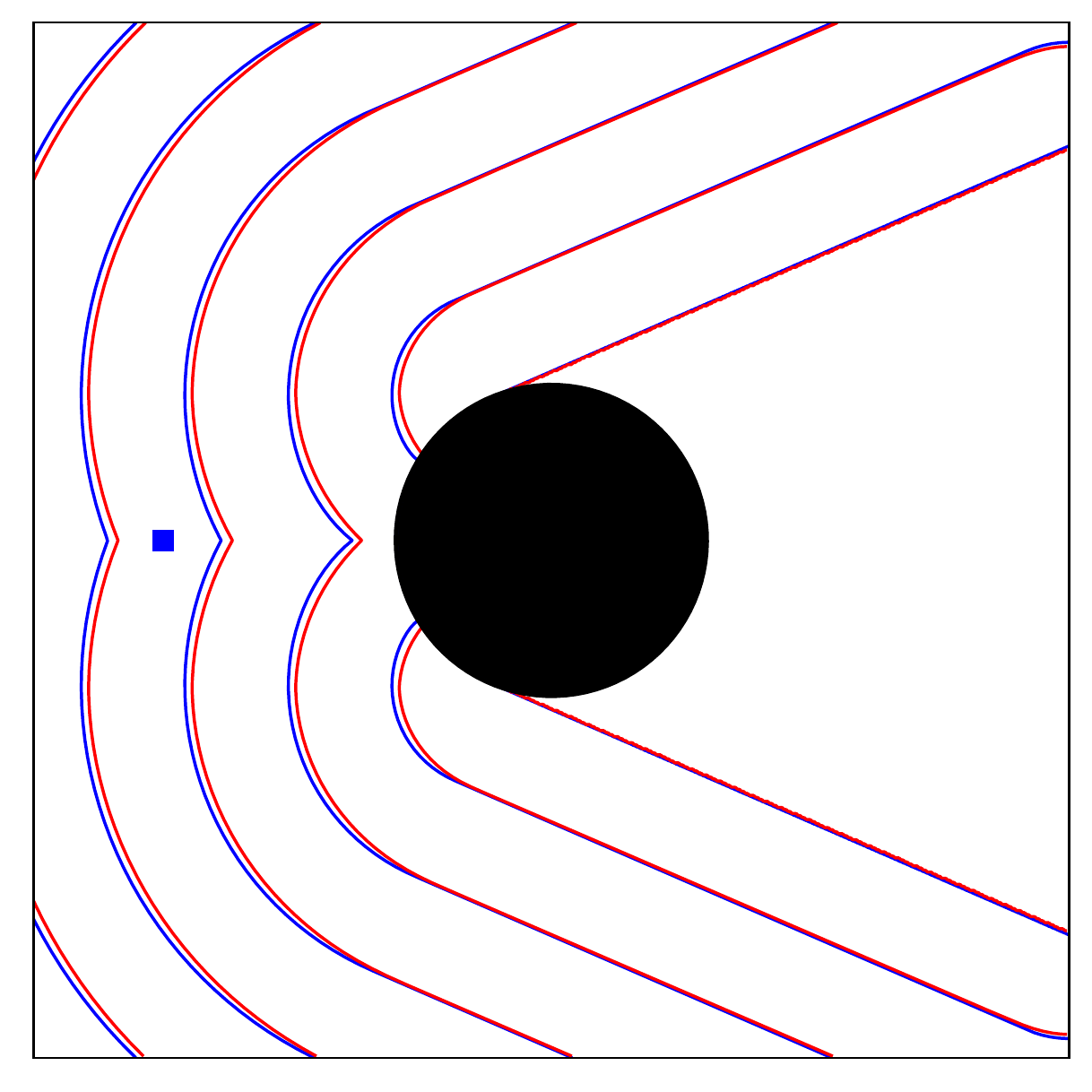} \quad
      \includegraphics[width=.45\textwidth]{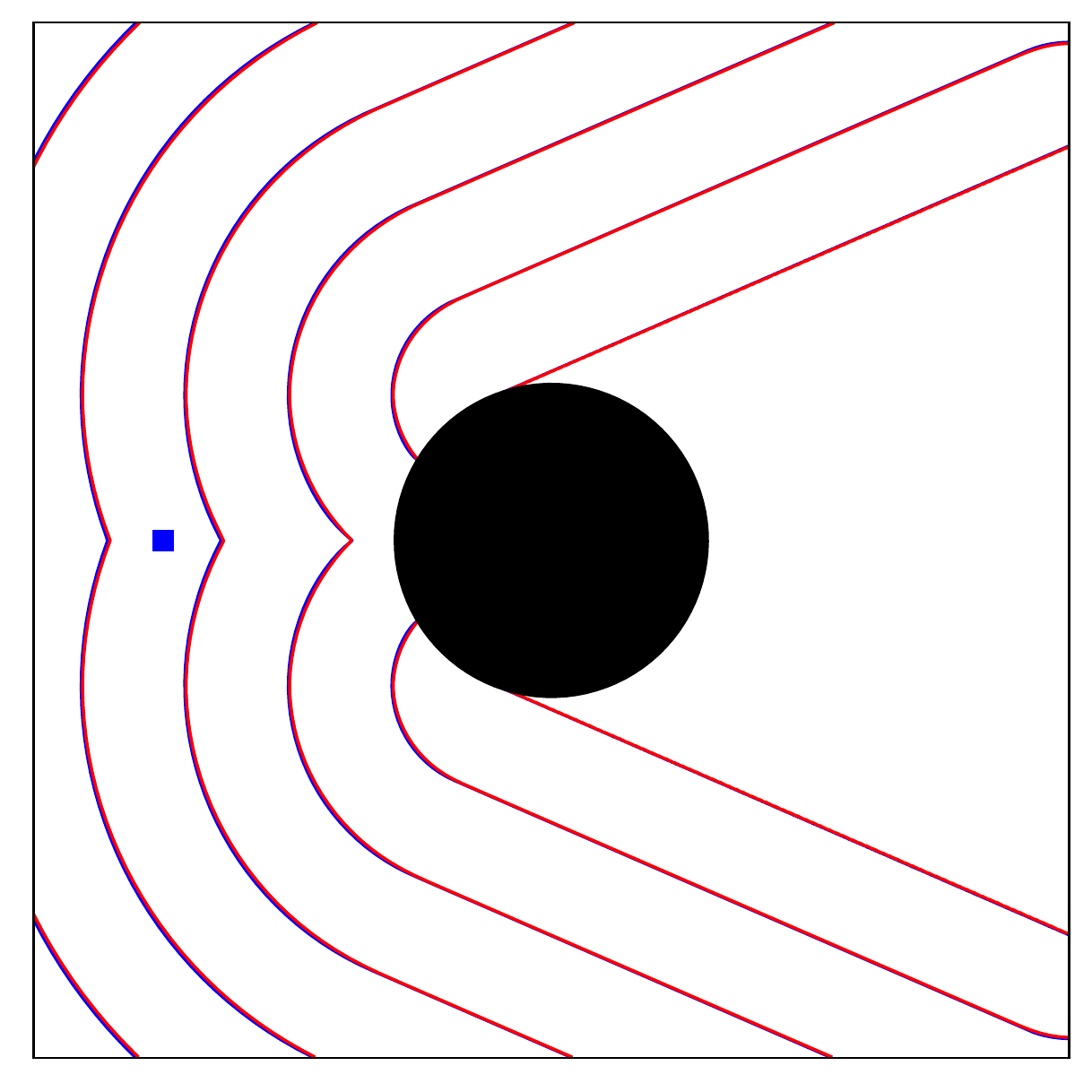}

	  \caption{Comparison of contours of the ``exact'' solution (blue) with those
computed by the scheme (\ref{eq:hji-scheme}) (red) using grid resolutions $m=512$ (left) and
$m=1024$ (right).  The pursuer (blue square) is stationary.  The error emanates from the obstacle
due to boundary conditions, but the scheme converges
as the grid is refined.} \label{fig:hji_contour}

\end{figure}

We compute the ``exact'' solution using fast marching method on high resolution grid $M=2048$.
We vary $m$ from $16$ to $1024$ and observe convergence in the $L_1$ and $L_2$ sense,
as seen in Table~\ref{tab:hji_conv}.
In Figure~\ref{fig:hji_contour}, we plot the level curves comparing the computed solution at $m=512,1024$ to the ``exact'' solution.
Notice the discrepancies are a result of the difficulty in dealing with boundary conditions.
However, these errors decay as the grid is refined.

The case where the evader is stationary is not interesting.

\begin{table}[]
\begin{center}
\caption{Error for the stationary pursuer case, compared to the known solution computed using fast marching method at resolution $M=2048$.\vspace{1em}}
\label{tab:hji_conv}

\begin{tabular}{c|c|c}
$m   $ & $L_1$ error             & $L_2$ error            \\ 
\hline
$16  $ & $0.08972215 $ &$0.01288563 $ \\
$32  $ & $0.03177683 $ &$0.00159669 $ \\
$64  $ & $0.02442984 $ &$0.00111537 $ \\
$128 $ & $0.01059728 $ &$0.00021345 $ \\
$256 $ & $0.00515584 $ &$0.00005214 $ \\
$512 $ & $0.00304322 $ &$0.00001961 $ \\
$1024$ & $0.00086068 $ &$0.00000142 $
\end{tabular}
\end{center}
\end{table}

%$16  $ & $0.08972215762198651  $ &$0.012885633288809298 $ \\
%$32  $ & $0.031776830083754815 $ &$0.001596690139730697 $ \\
%$64  $ & $0.024429847742256806 $ &$0.001115370592972781 $ \\
%$128 $ & $0.010597284307555795 $ &$0.000213453745010690 $ \\
%$256 $ & $0.005155845526871389 $ &$0.000052147495364447 $ \\
%$512 $ & $0.003043227116595554 $ &$0.000019619974499759 $ \\
%$1024$ & $0.000860688403850700 $ &$0.000001424361236251 $

\subsubsection*{A circular obstacle}

The evader has the advantage in the surveillance-evasion game. It is
difficult for the pursuer to win unless it is sufficiently fast.
But once it is fast enough, it can almost always win. 
Define the winning regions for the pursuer and evader, respectively:
\begin{align}
\mathcal{W}_\xp &= \{(\xp,\xe) | \val(\xp,\xe) > T_\text{max}\} \\
\mathcal{W}_\xe &= \{(\xp,\xe) | \val(\xp,\xe) \le T_\text{max}\} 
\end{align}

\begin{figure}[hptb]
      \centering
      \includegraphics[width=.78\textwidth]{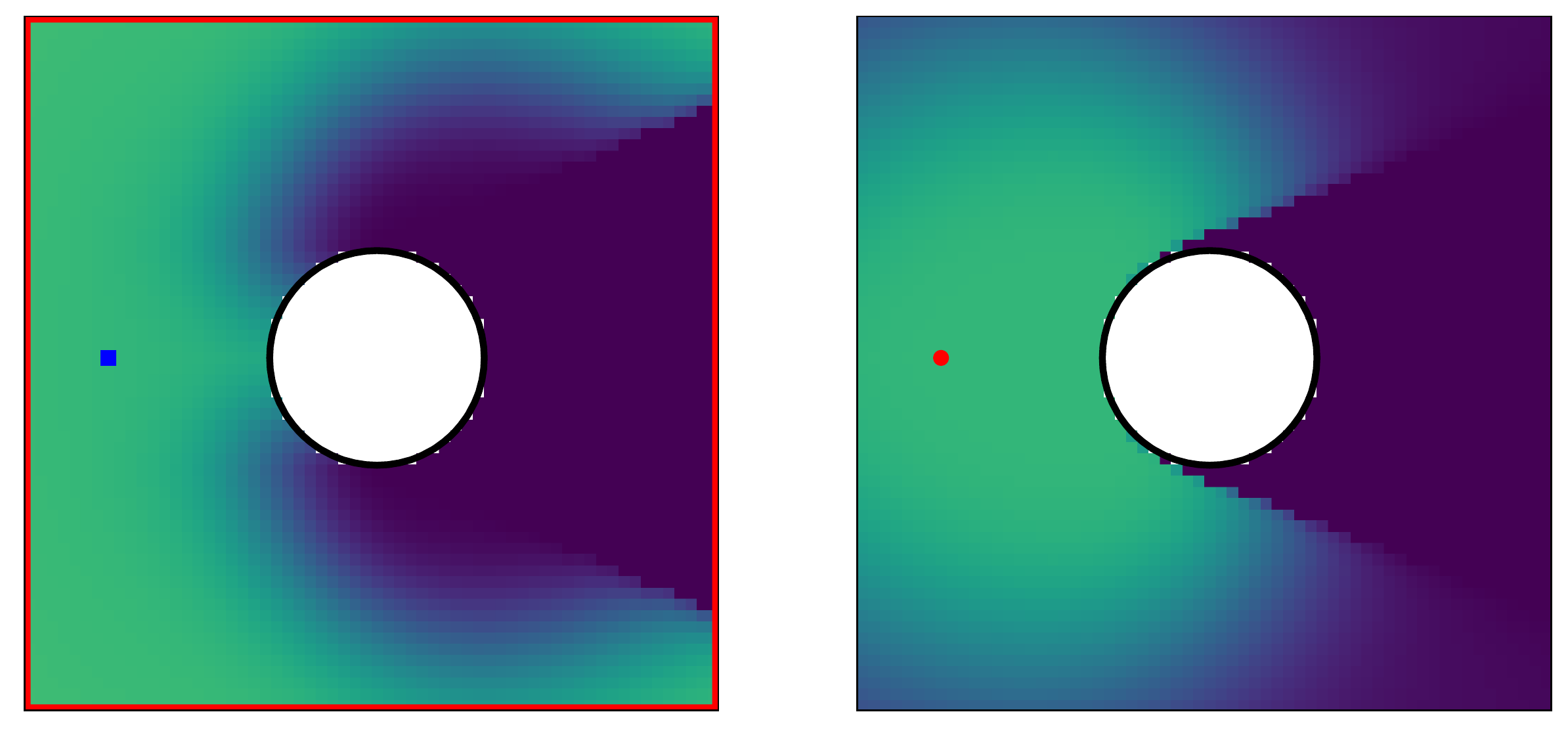}
      \includegraphics[width=.78\textwidth]{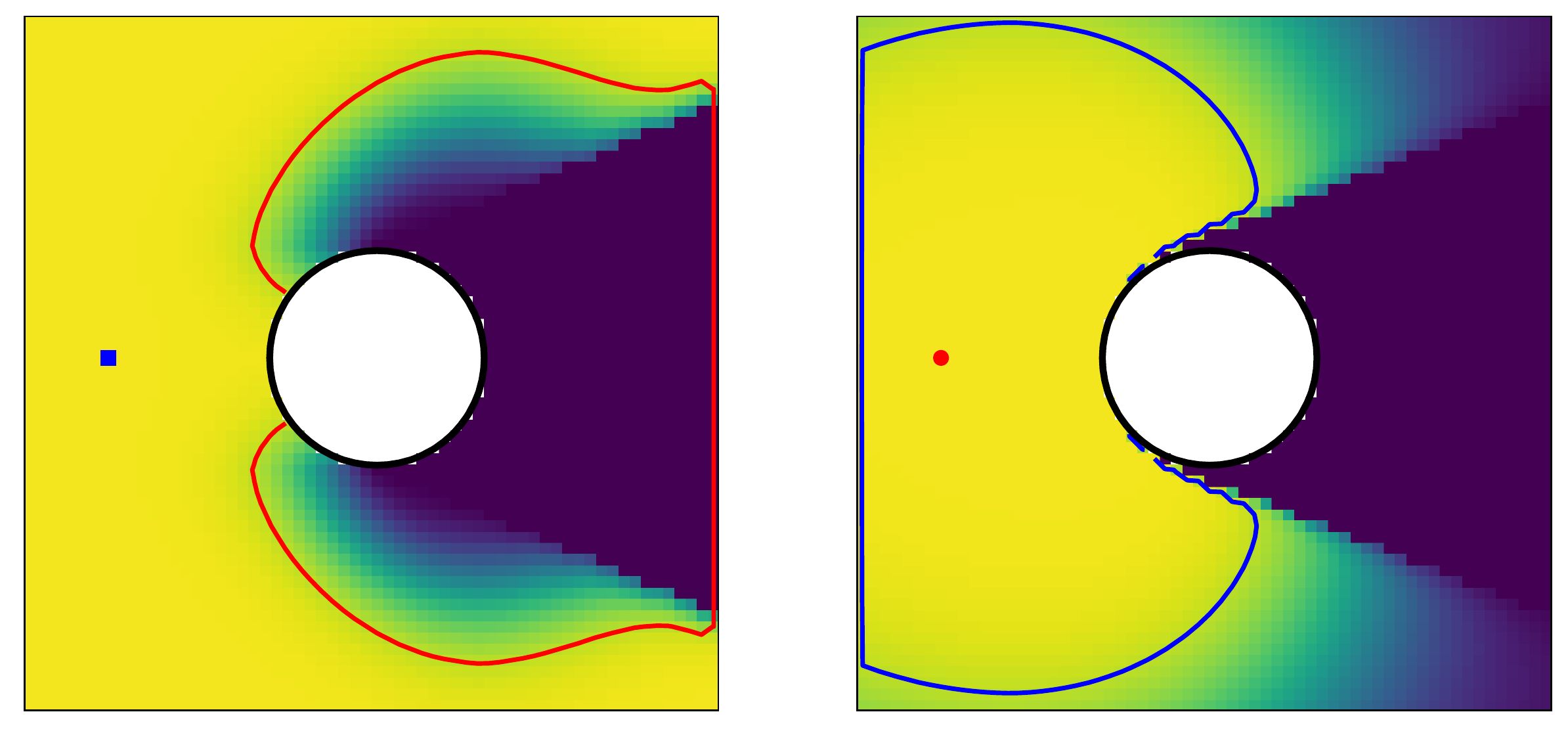}
      \includegraphics[width=.78\textwidth]{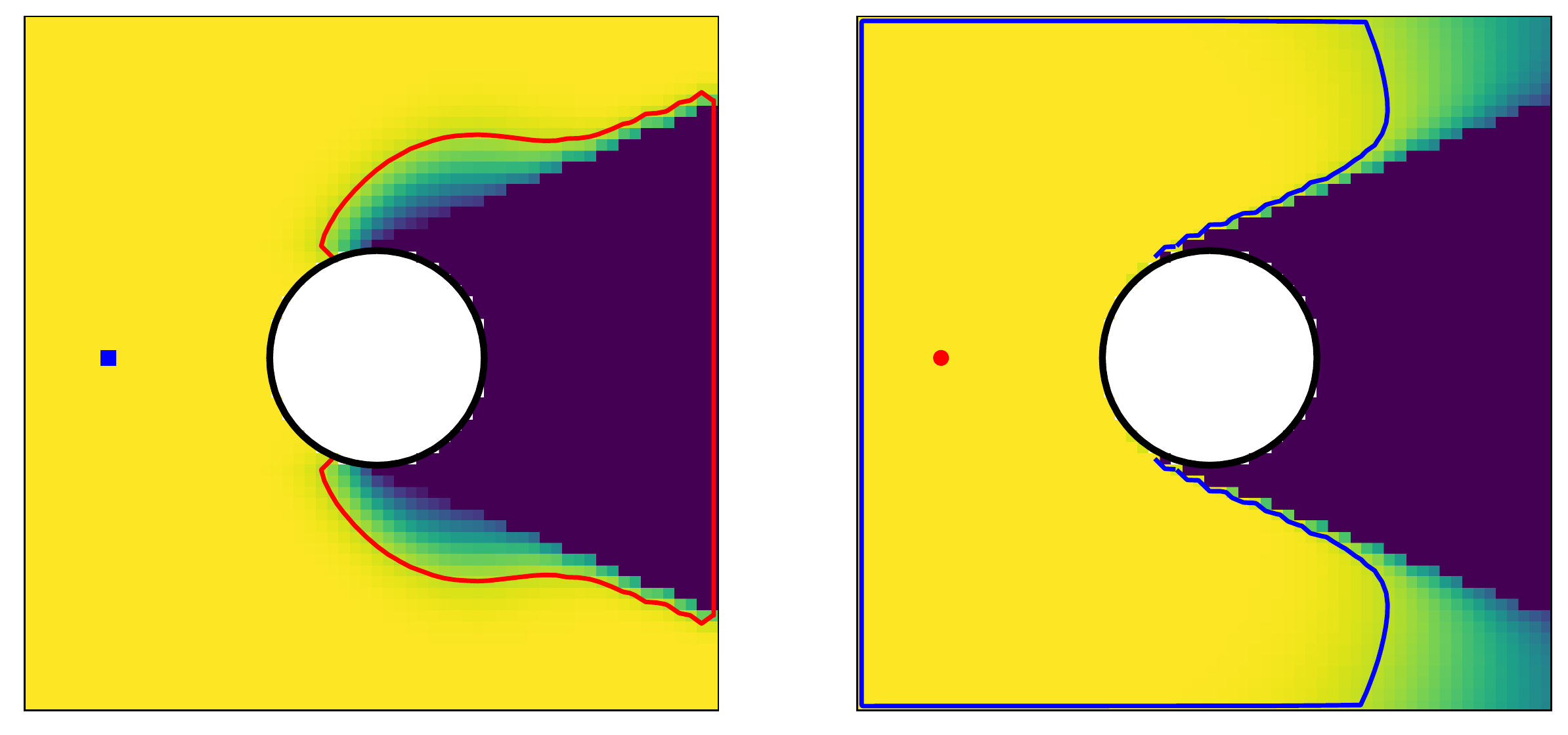}
	  \caption{Comparison of winning initial positions for the evader (left,
red contour) against a pursuer with fixed initial position (blue square) and
vice versa -- winning initial positions for the pursuer (right, blue contour)
against an evader with fixed initial position (red circle).  Left column shows
$\val(\xpinit,\cdot)$ while right column shows $\val(\cdot,\xeinit)$, where higher values of
$\val$ are yellow, while lower values are dark blue. From top to bottom, the pursuer is $1.5$, $2$
and $3$ times faster than the evader.  
The pursuer must be sufficiently fast to have a chance at winning.
} \label{fig:hj_circles_winset} \end{figure}

Here, we use $T_\text{max}=.9 T$ to tolerate numerical artifacts due to boundary conditions.  
In Figure~\ref{fig:hj_circles_winset}, we show how the winning region
for a fixed evader/pursuer position changes as the pursuer's speed increases.
Since it is difficult to visualize data in
4D, we plot the slices $\val(\xpinit,\cdot)$ and $\val(\cdot,\xeinit)$ where
$\xpinit=\xeinit=(1/8,1/2)$. 
We use $m=64$, $\Delta t = \Delta x/20$ and iterate until $T=10$.
We fix $\pspeed=1$ and compute $\val$ for each $\espeed\in\{\frac{1}{3},\frac{1}{2},\frac{2}{3}\}$.
The computation for each value function takes 16 hours.

%As P's speed increases, it becomes less necessary to move, because it can always catch up.
%Also, P can force E into a fixed point, where it becomes impossible to win. In this case,
%E becomes stationary and is essentially \emph{captured}.

\begin{figure}[hptb]
      \centering
      \includegraphics[width=.45\textwidth]{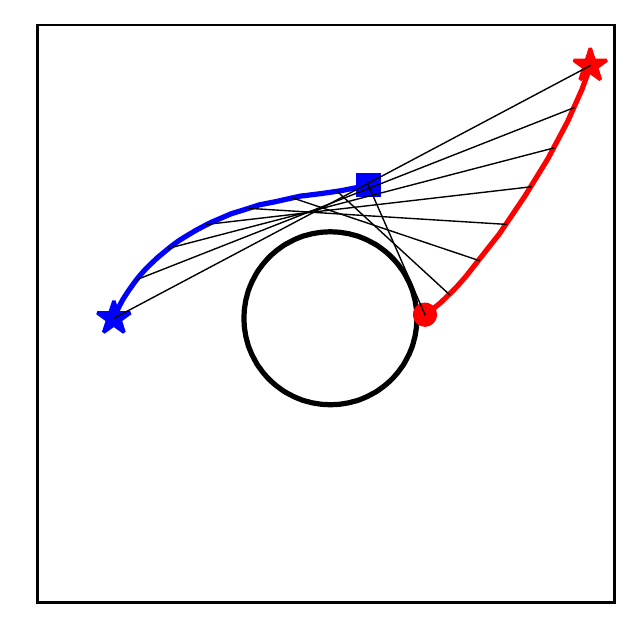}
      \includegraphics[width=.45\textwidth]{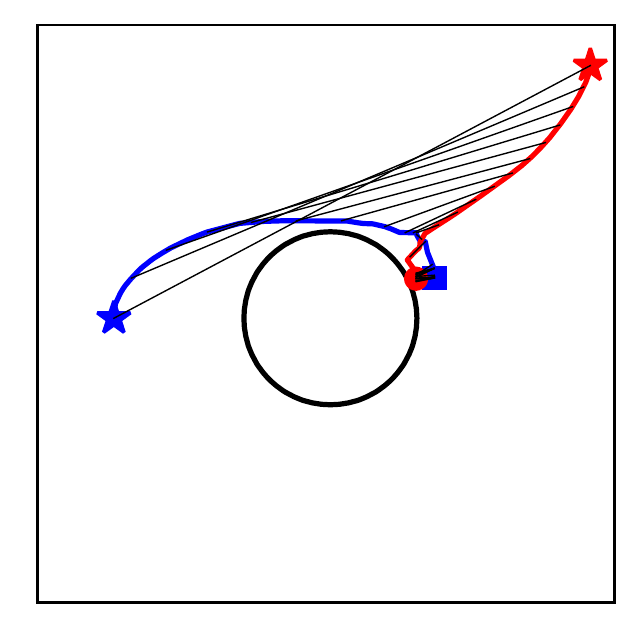}
      \includegraphics[width=.45\textwidth]{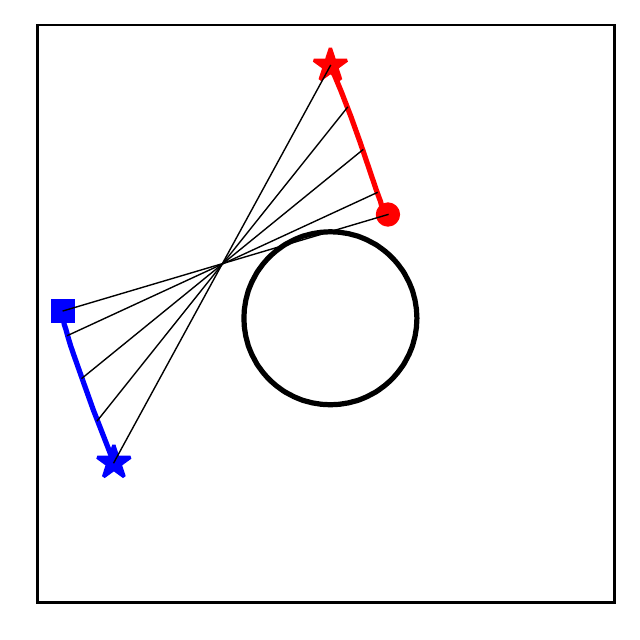}
      \includegraphics[width=.45\textwidth]{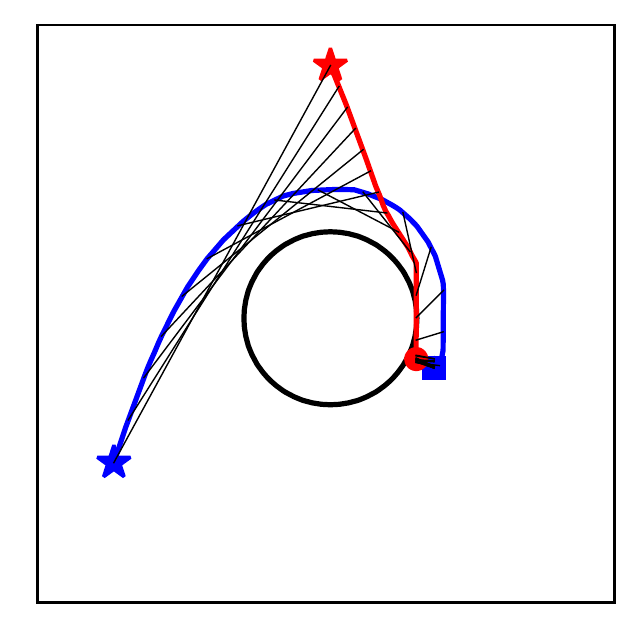}

	  \caption{Trajectories of several games played around a circle. The
pursuer loses when it has same speed as the evader (left column). When the pursuer is 2x faster than the evader,
it is possible to win; the evader essentially gives up once it is cornered, since no controls will change the outcome (right column).
Initial positions are shown as stars. Black lines connect positions at constant time intervals.}

\label{fig:hj_circles_traj} \end{figure}

Figure~\ref{fig:hj_circles_traj} shows trajectories from several initial positions
with various speeds. Interestingly, once the evader is cornered, the optimal controls
dictate that it is futile to move. That is, the value function is locally constant.

\subsubsection*{More obstacles}
\begin{figure}[htb]
      \centering
      \includegraphics[width=.45\textwidth]{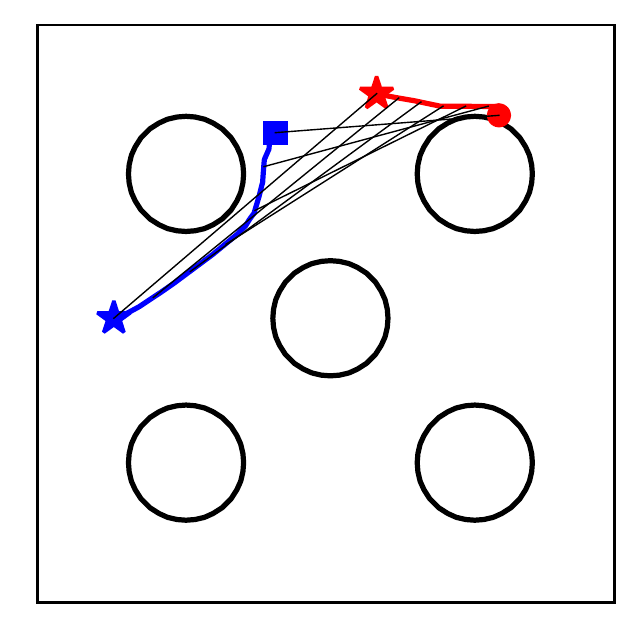}
      \includegraphics[width=.45\textwidth]{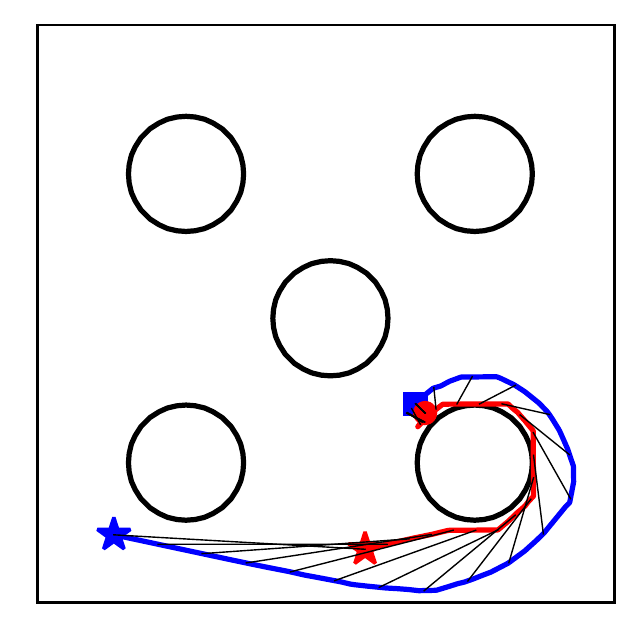}
	  \caption{Trajectories of games played around 5 circular obstacles. Pursuer (blue) is 2x as fast as evader (red).
The evader wins (left) if it can quickly hide. Otherwise it will give up once it is captured (right).
Initial positions are shown as stars. Black lines connect positions at constant time intervals.}
\label{fig:hj_dice_traj} \end{figure}

In Figure~\ref{fig:hj_dice_traj} we consider a more complicated environment with multiple obstacles.
Here, the pursuer is twice as fast as the evader. Although there are many obstacles, the dynamics
are not so interesting in the sense that the evader will generally navigate towards a single obstacle.
Again, the evader tends to give up once the game has been decided.

\begin{figure}[hptb]
      \centering
      \includegraphics[width=.45\textwidth]{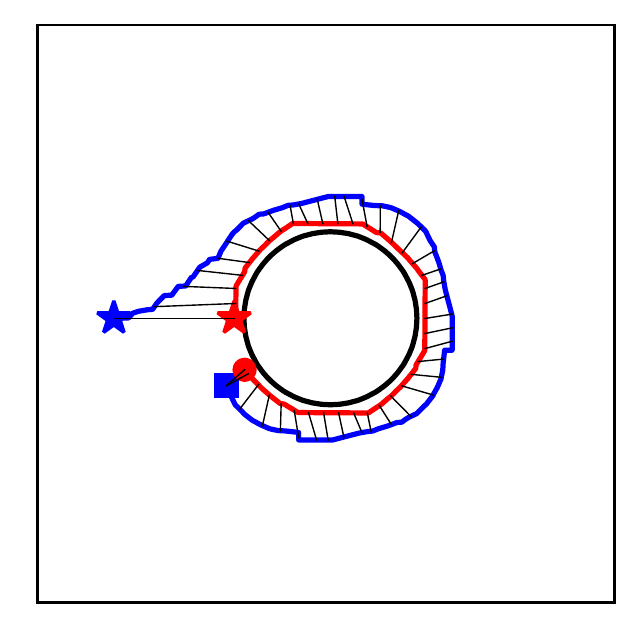}
      \includegraphics[width=.45\textwidth]{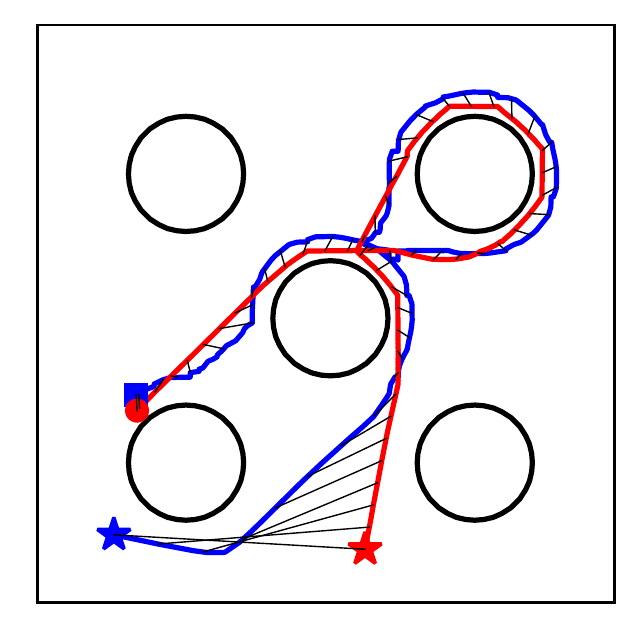}
	  \caption{Manually controlled evader against an optimal pursuer. The evader loses in both cases, but does not give up.}
\label{fig:hj_human_traj} \end{figure}

Finally, in Figure~\ref{fig:hj_human_traj} we show suboptimal controls for the
evader. In particular, the evader is controlled manually. Although manual
controls do not help the evader win, they lead to more interesting
trajectories.


\subsection{Discussion}
Notice that an optimal trajectory for the pursuer balances distance and
visibility.  While moving closer to the evader will guarantee that it can't
``get away'', it is not sufficient, since being close leads to more
occlussions. On the other hand, moving far away gives better visibility of the
environment, but may make it impossible to catch up once E turns the corner.

Although we described the formulation for a single pursuer and evader,
the same scheme holds for multiple pursuers and evaders. The end-game set just
needs to be modified to take into account the multiple players.
That is, the game ends as soon as any evader is occluded from all pursuers.
However, the computational complexity of the scheme is $\bigo(m^{kd})$ which
quickly becomes unfeasible even on small grid sizes.
At the expense of offline compute time, the game can be played efficiently
online. The caveat is that, the value function is only valid for the specific
map and velocities computed offline.

%Note that in the limit of infinite evaders, the game reduces to the surveillance problem from chapter~\ref{chap:exploration}.

\section{Locally optimal strategies}
\label{sec:locally}
%Given enough MCTS iterations and training time, one can expect
%that the neural network trained using self play can continue to improve.
%However, it may only be feasible for large research labs which have access to
%vast computational resources.
%One may consider strategies to help improve the efficiency of
%self-play. 
%Or one may want to train a better evader.
%This can be done using, for example, the pursuer's neural network.

As the number of players increases, computing the value function from the HJI
equations is no longer tractable.
We consider a discrete version of the game, with the aim of feasibly
computing controls for games with multiple pursuers and multiple evaders.  
Each player's position is now restricted on a grid, and at each turn,
the player can move to a new position within a neighborhood determined by its
velocity. Players cannot move through obstacles. 
Formally, define the arrival time function
\beq
\parrival(x,y) &:= \min_{\pcontrol\in\admiss} \min \{t|\xp(0)=x,\xp(t)=y\}\\
\earrival(x,y) &:= \min_{\econtrol\in\admiss} \min \{t|\xe(0)=x,\xe(t)=y\}  .
\eeq
The set of valid actions are the positions $y$ which can be reached from $x$
within a $\Delta t$ time increment:
\beq
\pvalid(t)&:=\{y\in\free | \parrival(\xp(t),y) \le \Delta t\}\\
\evalid(t)&:=\{y\in\free | \earrival(\xe(t),y) \le \Delta t\}.
\eeq
In a $\Delta t$ time step, each player can move to a position
\beq
\xp(t+\Delta t) \in \pvalid(t) \\
\xe(t+\Delta t) \in \evalid(t).
\eeq

Analogously, for multiple players, denote the number of pursuers and evaders as
$k_P$ and $k_E$, respectively. Define
\beq
\bfxp &= (\xp_1,\dots, \xp_{k_P})\\
\bfxe &= (\xe_1,\dots, \xe_{k_E})\\
\bfpvalid(t) &= \{\bfxp| \xp_i \in \pvalid(t) \ , \ i=1,\dots,k_P\} \\
\bfevalid(t) &= \{\bfxe| \xe_j \in \evalid(t) \ , \ j=1,\dots,k_E\}
\eeq
so that in $\Delta t$ time, each team can move to
\beq
\bfxp(t+\Delta t) \in \bfpvalid(t) \\
\bfxe(t+\Delta t) \in \bfevalid(t) \\
\eeq

The game ends as soon as one evader is occluded from all
pursuers. The end-game set is
\beq
\failset = \{ (\bfxp,\bfxe) \ |\ \exists \ j : \shadow(\xp_i,\xe_j)\le 0 \text{ for } i=1,\dots,k_P\} ,
\eeq

We propose two locally optimal strategies for the pursuer.

\subsection{Distance strategy}
The trajectories from the section~\ref{sec:hji} suggest that the pursuer must 
generally remain close to the evader. Otherwise, the evader can quickly hide behind obstacles.
A simple strategy for the pursuer is to move towards the evader:
\beq
\xp(t+\Delta t) &= \argmin_{x\in\pvalid(t)} \parrival\Big(x,\xe(t) \Big).
\eeq
That is, in the time increment $\Delta t$, the pursuer should pick the action that minimizes
its travel time to the evader's current position at time $t$. 

%In the next section, where we use search algorithms to refine policies,
%it is useful to quantify the utility of each action.
%To do so, define $p_\text{distance}:\Omega \times \free \times \free \to \mathbb{R}$ as the policy,
%which outputs a probability the agent should take an action, conditioned on the current player positions.
%We normalize using the \emph{softmax} function to generate the policy:
%\beq
%\norm &= \Big[\displaystyle{\sum_{x\in\pvalid(t)}} e^{ -\parrival(x,\xe(t))} \Big]^{-1}\\
%p_\text{distance}\big(x| (\xp(t),\xe(t))\big) &=  
%\begin{dcases}
%\norm e^{ -\parrival(x,\xe(t))}   & x \in \pvalid(t)\\
%0 & \text{otherwise}
%\end{dcases}
%\eeq

For the multiplayer game,  we propose a variant of the
Hausdorff distance, where max is replaced by a sum:
\begin{align*}
%\min_\pcontrol \sum_{i=1}^{k_P} \min_j \parrival[i]\Big( \xp_i(t+\Delta t),\xe_j(t) \Big)^2  +    \sum_{j=1}^{k_E} \min_i \parrival[i]\Big( \xp_i(t+\Delta t),\xe_j(t)\Big)^2 
%\argmin_{\substack{x_i\in\pvalid[i](t)\\i=1,\dots,k_P}} \   \sum_{i=1}^{k_P} \min_j \parrival[i]\Big( x_i,\xe_j(t) \Big)^2  +    \sum_{j=1}^{k_E} \min_i \parrival[i]\Big( x_i,\xe_j(t)\Big)^2 .
{\bf \parrival}({\bf x, \xe}(t)) := 
\frac{1}{2}\Big[ \sum_{i=1}^{k_P} \min_j \parrival[i]\Big( x_i,\xe_j(t) \Big)^2 \ \Big]^{1/2}  +  \frac{1}{2}\Big[  \sum_{j=1}^{k_E} \min_i \parrival[i]\Big( x_i,\xe_j(t)\Big)^2 \ \Big]^{1/2} .
\end{align*}
Informally, the first term encourages each pursuer to be close to an evader,
while the second term encourages a pursuer to be close to each evader.  The sum helps
to prevent ties. 
The optimal action according to the distance strategy is
\beq
%%%{\bf x^\ast} := (x_1^\ast,\dots,x_{k_P}^\ast)= \argmin_{\substack{x_i\in\pvalid[i](t)\\i=1,\dots,k_P}}{\bf \parrival}({\bf x, \xe}(t))  
%%%{\bf x^\ast} := (x_1^\ast,\dots,x_{k_P}^\ast)= \argmin_{{\bf x} \in \bfpvalid(t)}{\bf \parrival}({\bf x, \xe}(t))  
{\bfxp(t+\Delta t)} = \argmin_{{\bf x} \in \bfpvalid(t)}{\bf \parrival}({\bf x, \xe}(t))  
\eeq

In the next section, we will use search algorithms to refine policies.  Rather
than determining the \emph{best} action, it is useful to quantify the utility
of each action. To do so, define 
$p_\text{distance}:\Omega \times \free \times \dots \times \free \to \mathbb{R}$ as the policy,
which outputs a probability the agent should take an action, conditioned on the current player positions.
We normalize using the \emph{softmax} function to generate the policy:
%The corresponding policy for each pursuer is the marginal distribution corresponding to ${\bf x^\ast}$, in following sense.
%Define ${\bf x}^\ast_{-i} = (x_1^\ast,\dots,x_{i-1}^\ast,x_i,x_{i+1}^\ast,\dots,x_{k_P}^\ast)$. That is, replace the $i^{th}$ element in ${\bf x}^\ast$ with $x_i$.
%Then the distance policy for the $i^{th}$ pursuer is
\beq
\norm &= \Big[\displaystyle{\sum_{{\bf x}\in\bfpvalid(t)}} e^{ -{\bf \parrival}({\bf x},{\bf \xe}(t))}\Big]^{-1}\\
p_\text{distance}\big({\bf x}| ({\bf \xp}(t),{\bf \xe}(t))\big) &= 
\begin{dcases}
 \norm e^{ -{\bf \parrival}({\bf x},{\bf \xe}(t))}   & {\bf x} \in \bfpvalid(t)\\
0 & \text{otherwise}
\end{dcases}
\eeq

%In the case where there are more evaders than pursuers, the square term encourages a pursuer to stand in between evaders, rather than choosing to be close to an one.

For the discrete game with $\Delta t$ time increments,
one can enumerate the possible positions for each $\xp_i$ 
and evaluate the travel time to find the optimal control.
The arrival time function $\parrival[i](\cdot,\xe_j(t))$ to each $\xe_j$ at the current
time $t$ can be precomputed in $\bigo(m^d)$ time.  
In the general case, where each pursuer may have a different velocity field $\pspeed_i$, one
would need to compute $k_P$ arrival time functions.
If $a_{\Delta t}(P_i)$ is the
max number of possible actions each pursuer can make in $\Delta t$ increment, then the total
computational complexity for one move is 
$$ O\Big(k_P k_E m^d + \prod_{i=1}^{k_P} a_{\Delta_t}(P_i)\Big). $$

For the special case when the pursuers have the same $\pspeed$, the complexity reduces to
$$ O\Big(k_E m^d + \prod_{i=1}^{k_P} a_{\Delta_t}(P_i)\Big). $$

\subsection{Shadow strategy}

Recall that, for a stationary pursuer, the value function for the evader becomes
the Eikonal equation:
\beq
\espeed |\nabla_\xe \val|  &= 1 \qquad &\text{ on } \free\setminus\failset \\
\val(\xp,\xe) &= 0 &\qquad (\xp,\xe )\in \failset,
\eeq
whose solution is the travel time to the shadow set.
Define the time-to-occlusion as
\beq
\tstar(\xp,\xe^0) :=  \min_{\econtrol\in\admiss}  \min \{ t\ge 0 \ | \ \xe(0)=\xe^0 \ , \ (\xp,\xe(t)) \in \failset \ \} .
\eeq
It is the shortest time in which an evader at $\xe^0$ can be occluded from a stationary pursuer at $\xp$.
Thus, a reasonable strategy for the evader is to pick the action which brings it
closest to the shadow formed by the pursuer's position:
\beq
\xe(t+\Delta t) = \argmin_{y \in\evalid(t)} \tstar(\xp(t),y) .
\label{eq:evader-action}
\eeq
A conservative strategy for the pursuer, then, is to maximize time-to-occlusion, assuming that the evader can anticipate its actions:
\begin{align}
\tstar^\ast(x,\xe(t))  &= \min_{y \in\evalid(t)} \tstar(x,y)\\
\xp(t+\Delta t) &= \argmax_{x\in\pvalid(t)} \tstar^\ast(x) \label{eq:local-takei}.
\end{align}

%Define $p_\text{shadow}:\Omega \times \free \times \free \to \mathbb{R}$ as the 
%corresponding policy:
%\beq
%\norm &= \Big[\displaystyle{\sum_{x\in\pvalid(t)}} e^{ \tstar^\ast(x,\xe(t))} \Big]^{-1}\\
%p_\text{shadow}\big(x| (\xp(t),\xe(t))\big) &=  
%\begin{dcases}
%\norm e^{ \tstar^\ast(x,\xe(t))}   & x \in \pvalid(t)\\
%0 & \text{otherwise}
%\end{dcases}
%\eeq

\emph{Remark:} The strategy (\ref{eq:local-takei}) is a local variant of the static value
function proposed in \cite{takei2014efficient}.  In that paper, they suggest
using the static value function for feedback controls by moving towards the
globally optimal destination, and then recomputing at $\Delta t$ time
intervals. Here, we use the locally optimal action.

For multiple players, the game ends as soon as any evader is hidden from all pursuers. Define the time-to-occlusion for multiple players:
\beq
\bftstar({\bfxp},\bfxe^0):= \min_{\econtrol_i\in\admiss}  \min \{ t\ge 0 \ | \ \bfxe(0)=\bfxe^0 \ , \ (\bfxp,\bfxe(t)) \in \failset \ \} .
\eeq
Then, the strategy should consider the shortest time-to-occlusion among all possible evaders' actions in the $\Delta t$ time increment:
\beq
\bftstar^\ast({\bf x},{\bf \xe}(t))  &:= \min_{{\bf y} \in\bfevalid(t)} \bftstar({\bf x},{\bf y}) \\
\bfxp(t+\Delta t)  &= \argmax_{{\bf x}\in\bfpvalid(t)} \bftstar^\ast({\bf x},{\bf \xe}(t)).
 %%{\bf x}^\ast &= \argmax_{\substack{x_i\in\pvalid[i](t)\\i=1,\dots,k_P}} \bftstar^\ast({\bf x},{\bf \xe}(t)).
%\sup_{\pcontrol_i} \inf_{\econtrol_j}  \Big[ \inf_{j} \{ t | (\xp_i(t+\Delta t),\xe_j(T)) \in \failset \text{ for } i=1,\dots,k_P \Big]
\eeq
The corresponding shadow policy is:
\beq
\norm &= \Big[\displaystyle{\sum_{x\in\bfpvalid(t)}} e^{ \bftstar^\ast({\bf x},{\bf \xe}(t))}\Big]^{-1}\\
p_\text{shadow}\big({\bf x}| ({\bf \xp}(t),{\bf \xe}(t))\big) &= 
\begin{dcases}
 \norm e^{ \bftstar^\ast({\bf x},{\bf \xe}(t))}   & {\bf x} \in \bfpvalid(t)\\
0 & \text{otherwise}
\end{dcases}
\eeq

This strategy is computationally expensive. 
One can precompute the arrival time to each evader by solving an
Eikonal equation $\bigo(m^d)$.
For each combination of pursuer
positions, one must compute the joint visibility function  and corresponding
shadow function $\bigo(m^d)$.
Then the time-to-occlusion can be found by evaluating the precomputed
arrival times to find the minimum within the shadow set $\bigo(m^d)$.
The computational complexity for one move is
\beq O \Big( k_E m^d + m^d \cdot  \prod_{i=1}^{k_P} a_{\Delta t}(P_i) \Big). \eeq

%This strategy becomes very expensive as one increases the number of pursuers, or the velocity of each pursuer.
%With sufficient memory, one can precompute the visibilities for each pursuer's possible action,
%in which case, the complexity is

One may also consider alternating minimization strategies to achieve
\beq O \Big(k_E m^d + m^d \cdot  \sum_{i=1}^{k_P} a_{\Delta t}(P_i) \Big), \eeq
though we leave that for future work.

\subsection*{Blend strategy}

We have seen from $\ref{sec:hji}$ that optimal controls for the pursuer balance the
distance to, and visibility of, the evader. Thus a reasonable approach would be
to combine the distance and shadow strategies. However, it is not clear how
they should be integrated.  One may consider a linear combination, but 
the appropriate weighting depends on the game settings and environment.
Empirically, we observe that the product
of the policies provides promising results across a range of scenarios. Specifically, 
\beq
p_\text{blend} \propto p_\text{shadow} \cdot p_\text{distance}
\eeq

\subsection{Numerical results}

\begin{figure}[hptb]
\centering
\includegraphics[width=.42\textwidth,trim={0 0 0.1in 0},clip]{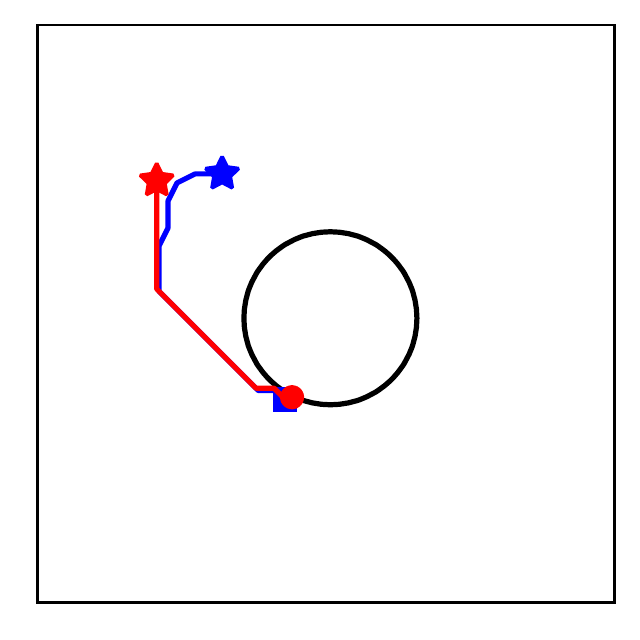}
\includegraphics[width=.42\textwidth,trim={0.1in 0 0 0},clip]{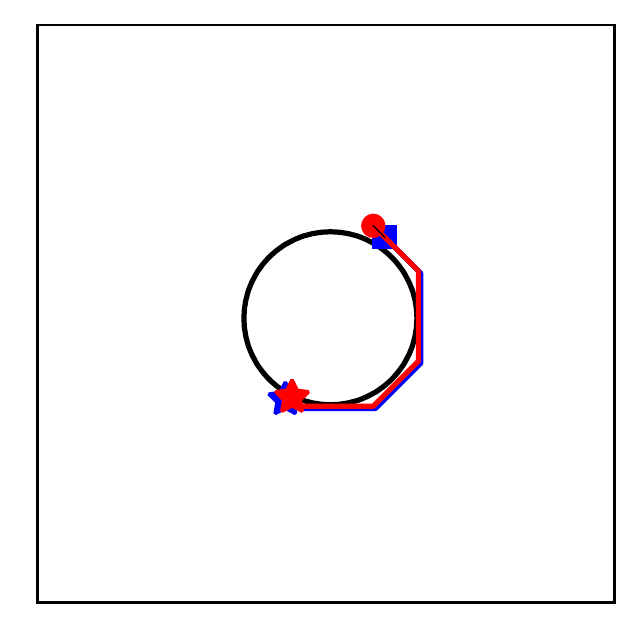}
\includegraphics[width=.42\textwidth,trim={0 0 0.1in 0},clip]{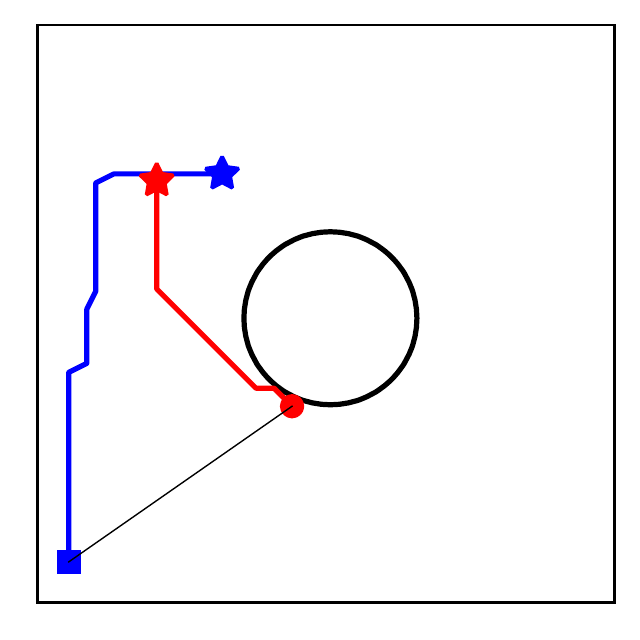}
\includegraphics[width=.42\textwidth,trim={0.1in 0 0 0},clip]{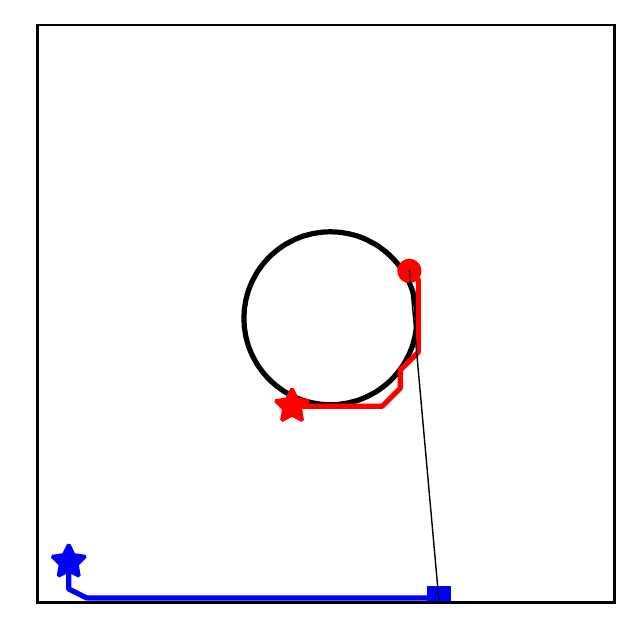}
\includegraphics[width=.42\textwidth,trim={0 0 0.1in 0},clip]{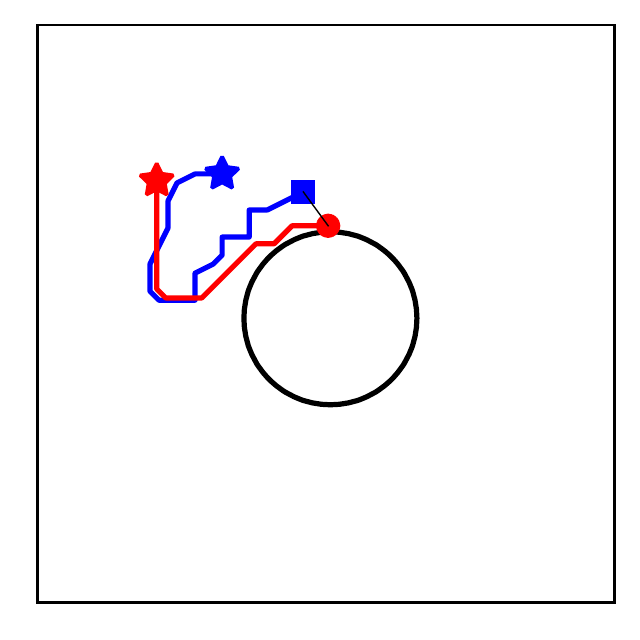}
\includegraphics[width=.42\textwidth,trim={0.1in 0 0 0},clip]{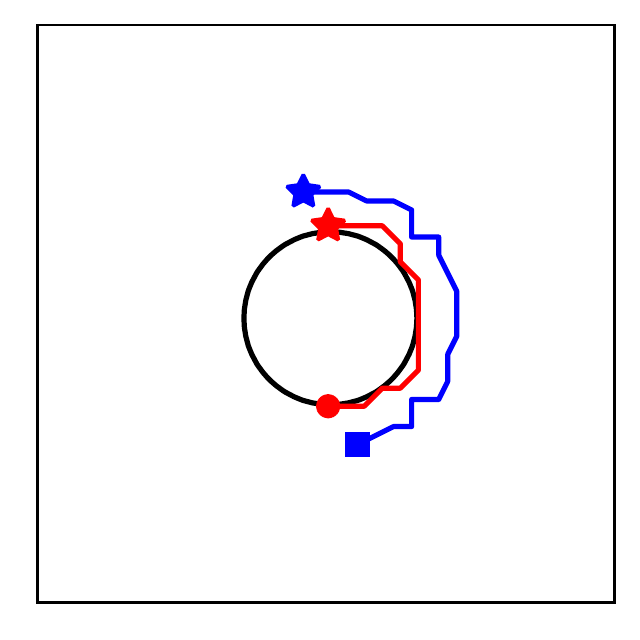}
\caption{Distance strategy (top) follows the evader closely, shadow strategy
(middle) stays far to gain better perspective, while the blend strategy
(bottom) strikes a balance.}

\label{fig:traj-local-circle}

\end{figure}

We present some representative examples of the local policies.
First, we consider the game with a circular obstacle, a single purser and single evader
whose speeds are $\pspeed=3$ and $\espeed=2$, respectively.
Figure~\ref{fig:traj-local-circle} illustrates the typical trajectories for each policy.
In general, the distance strategy leads the pursuer into a cat-and-mouse game
the with evader; the pursuer, when close enough, will jump to the evader's
position at the previous time step. The shadow strategy keeps the pursuer far
away from obstacles, since this allows it to \emph{steer the shadows} in the
fastest way.  The blend strategy balances the two approaches and resembles the
optimal trajectories based on the HJI equation in section~\ref{sec:hji}.

\begin{figure}[hptb]
\centering
\includegraphics[width=.42\textwidth,trim={0 .1in .1in 0},clip]{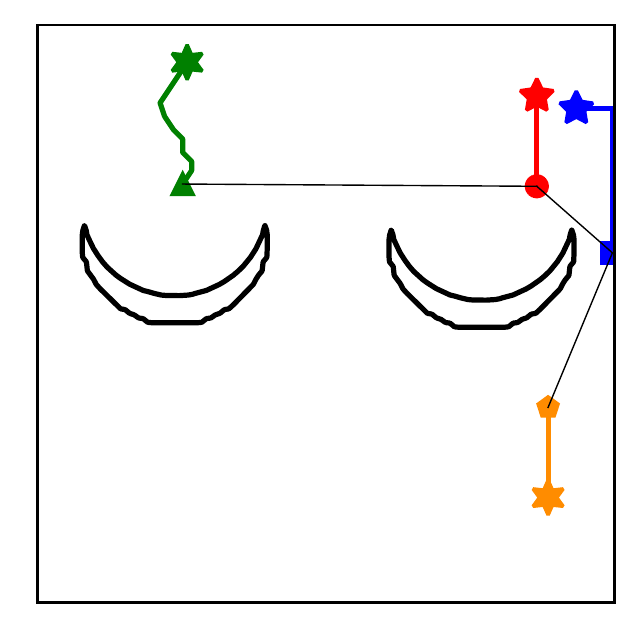}
\includegraphics[width=.42\textwidth,trim={.1in .1in 0 0},clip]{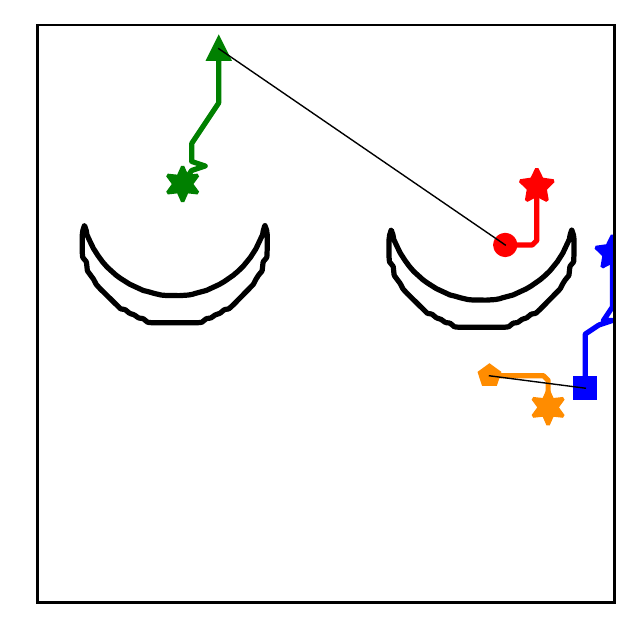}
\includegraphics[width=.42\textwidth,trim={0 .0 .1in .1in},clip]{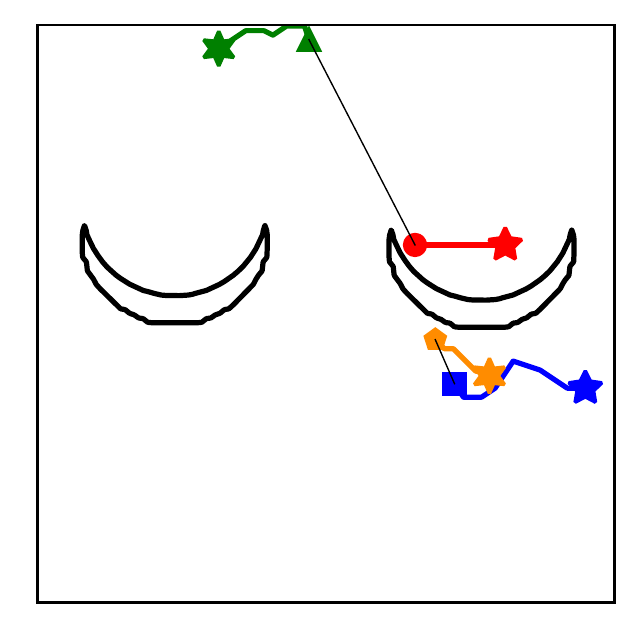}
\includegraphics[width=.42\textwidth,trim={.1in .0 0 .1in},clip]{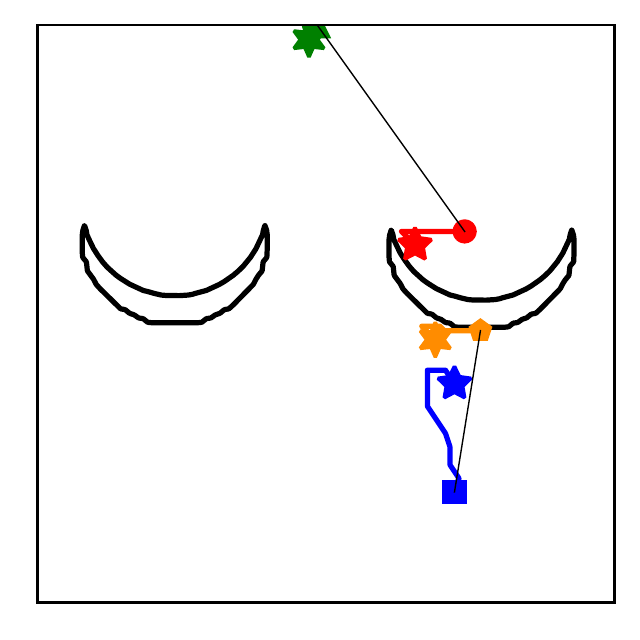}
\includegraphics[width=.42\textwidth]{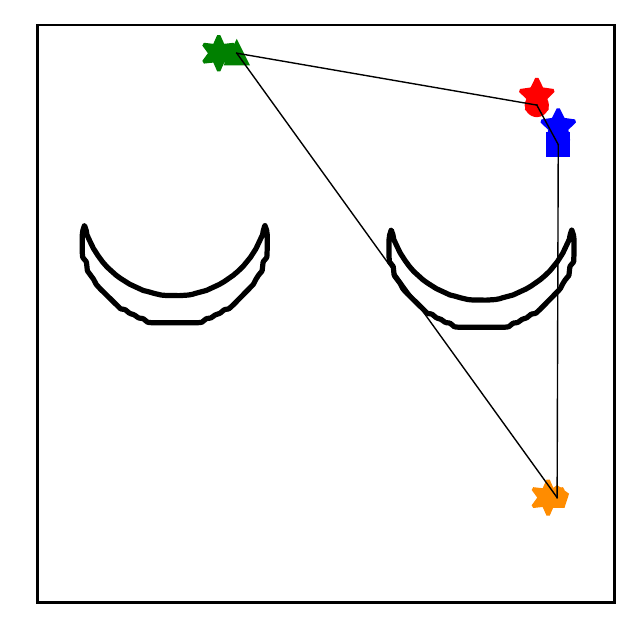}
\caption{(Top 2 rows) The blue and green pursuers cooperate by using the shadow strategy. Green initially has responsibility of the orange
evader, but blue is able to take over. 
(Bottom) The distance strategy loses immediately.
}
\label{fig:traj-shadow-eyes}
\end{figure}

Next, we highlight the advantages of the shadow strategy with a
2 pursuer, 2 evader game on a map with two crescent-shaped obstacles.
The pursuer and evader speeds are $\pspeed=4$ and $\espeed=2$, respectively.
The openness of the environment creates large occlusions. The pursuers use the
shadow strategy to cooperate and essentially corner the evaders.
Figure~\ref{fig:traj-shadow-eyes} shows snapshots of the game. 
The distance strategy loses immediately since the green purser does not 
properly track the orange evader.

\begin{figure}[hptb]
\centering
\includegraphics[width=.45\textwidth]{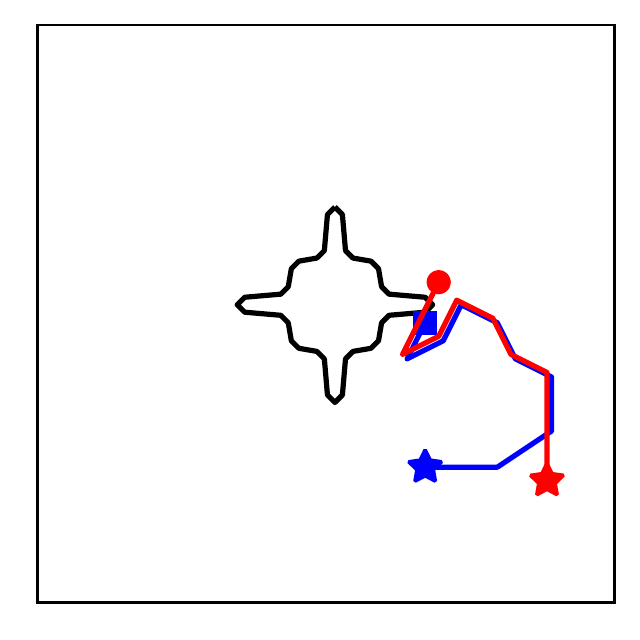}
\includegraphics[width=.45\textwidth]{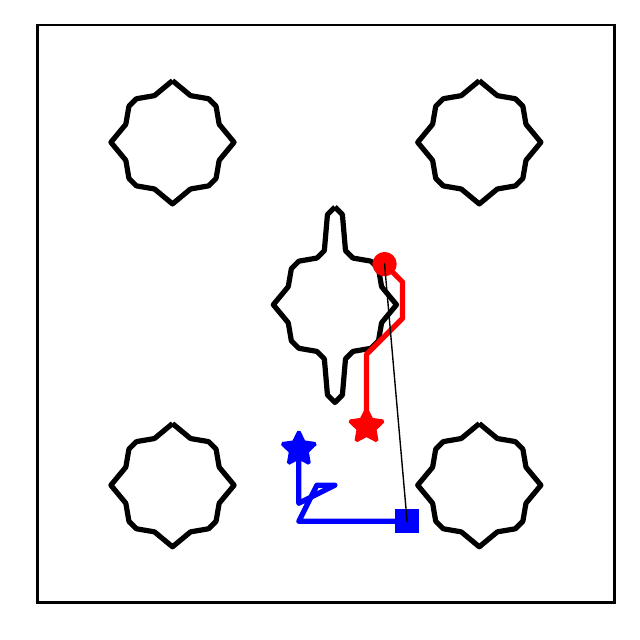}
\caption{Failure modes for the local strategies. Blindly using the distance
strategy (left) allows the evader to exploit the sharp concavities. The
shadow strategy (right) keeps the pursuer far away to reduce the size of shadows, but often,
the pursuer is too far away to catch the evader.}
\label{fig:traj-failure} \end{figure}

We present cases where the distance and shadow strategy fail in
Figure~\ref{fig:traj-failure}.  The evader tends to stay close to the obstacle,
since that enables the shortest path around the obstacle. Using the distance
strategy, the pursuer agressively follows the evader. The evader is able to
counter by quickly jumping behind sharp corners.  On the other hand, the shadow
strategy moves the pursuer away from obstacles to reduce the size of shadows.
As a consequence, the pursuer will generally be too far away from the evader
and eventually lose. In environments with many nonconvex obstacles, both
strategies will fail.

\begin{figure}[hptb]
\centering
\includegraphics[width=.45\textwidth]{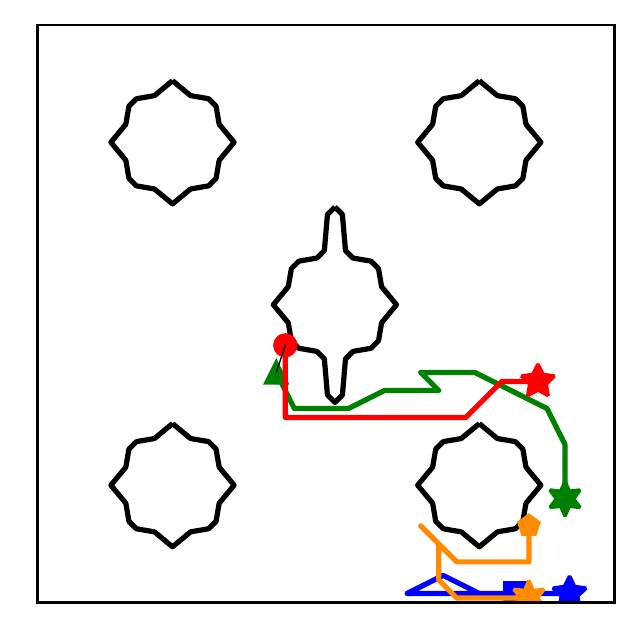}
\includegraphics[width=.45\textwidth]{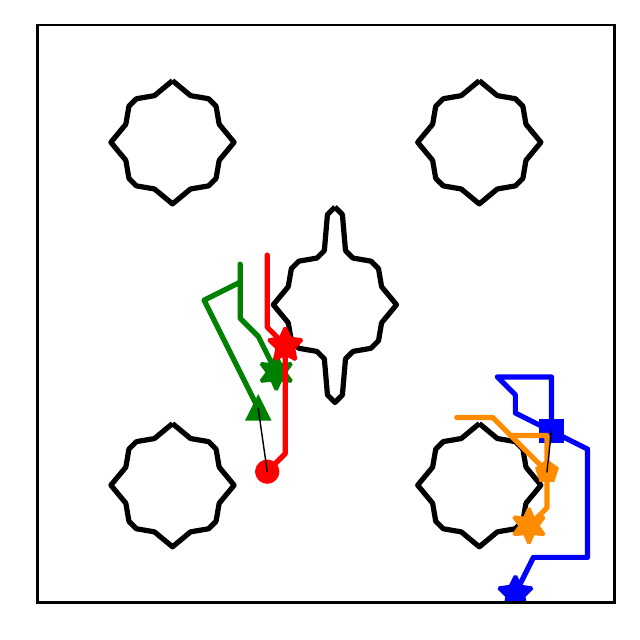}
\includegraphics[width=.45\textwidth]{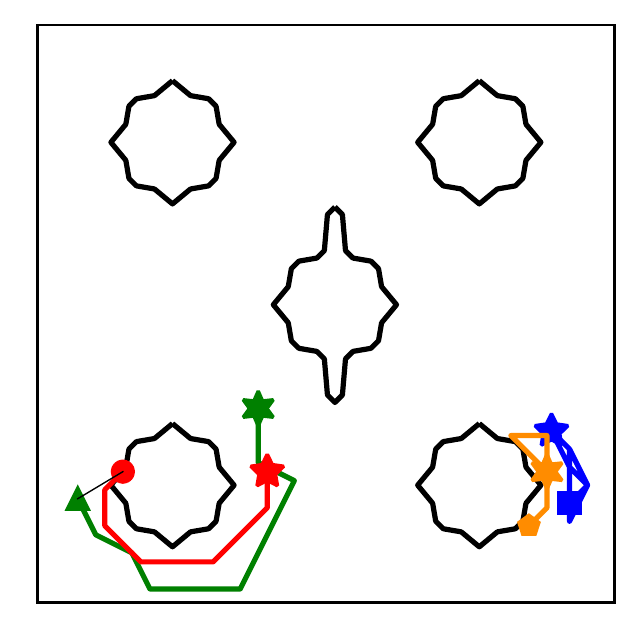}
\includegraphics[width=.45\textwidth]{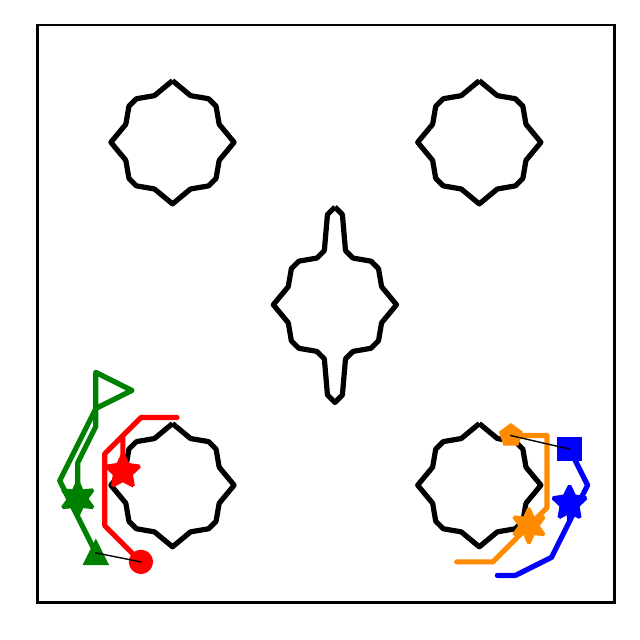}
\caption{The pursuers (blue and green) 
are able to win by combining the distance and shadow strategy. The pursuers stay close, while maintaining
enough distance to avoid creating large shadow regions. The pursuers are slightly faster than the evaders.
}
\label{fig:traj-blend-dice}
\end{figure}

Finally, we show that blending the shadow and distance strategies is very
effective in compensating for the shortcomings of each individual policy. 
The pursuers are able to efficiently track the evaders while maintain a safe distance.
Figure~\ref{fig:traj-blend-dice} shows an example with 2 pursuers and 2 evaders on
a map with multiple obstacles, where $\pspeed=3$ and $\espeed=2$.

% show how distance can fail around crescent,
% show how tstar fails many obstacles
% show how blend is good at both

\section{Learning the pursuer policy}
\label{sec:learning-policy}

We propose a method for learning optimal controls for the pursuer, though our
methods can be applied to find controls for the evader as well.  Again, we
consider a discrete game, where each player's position is restricted on a grid,
and at each turn, the player can move to a new position within a neighborhood
determined by their velocity. All players move simultaneously. The game
termination conditions are checked at the end of each turn.  

We initialize a neural network which takes as input any game state,
and produces a policy and value pair. The policy is probability distribution
over actions. Unlike in section~\ref{sec:hji}, the value, in this context, is
an estimate of the likely winner given the input state.

Initially the policy and value estimates are random.  We use Monte
Carlo tree search to compute refined policies.  We play the game using the
refined policies, and train the neural network to learn the refined policies.
By iterating in this feedback loop, the neural network continually learns to
improve its policy and value estimates. We train on games in various
environments and play games on-line on maps that were were not seen during the
training phase.

\subsection{Monte Carlo tree search}

In this section, we review the Monte Carlo tree search algorithm,
which allows the agent to plan ahead and refine policies. For clarity of notation,
we describe the single pursuer, single evader scenario, but the method applies to arbitrary
number of players.

Define the set of game states $\mathcal{S}:=\{(\xp,\xe)\in\free\times\free\}$ so that
each state $s\in\mathcal{S}$ characterizes the position of the players.
Let $\mathcal{A}\subseteq \free$ be the set of actions.
Let $\trans(s,a) : \mathcal{S}\times \mathcal{A} \to \mathcal{S}$ be the transition function
which outputs the state resulting from taking action $a$ at state $s$.
Let $\prior:\mathcal{S} \to \mathbb{R}^{m^d}\times[-1,1]$ be an evaluator function
which takes the current state as input and provides a policy and value estimate: $\prior(s)=(\vec{p},v)$.
Formally, Monte Carlo tree search is mapping takes as input the current state $s_0$, the evaluator function $\prior$, and a parameter $M$
indicating the number of search iterations: $\mcts(s_0,\prior;M)$.
It outputs a refined policy $\vec{\pi}^\ast$.

Algorithm \ref{alg:mcts} summarizes the MCTS algorithm.
At a high level, MCTS simulates game play starting from the current state,
keeping track of nodes it has visited during the search.
Each action is chosen according to a formula $U(s,a)$ which balances exploration and exploitation.
Simulation continues until the algorithm reaches a \emph{leaf node} $s_n$, a state which has not previously
been visited. At this point, we use the evaluator function $\prior(s_n)=(\vec{p},v)$ to estimate a policy
and value for that leaf node. The value $v$ is propagated to all parent nodes.
One iteration of MCTS ends when it reaches a leaf node.
MCTS keeps track of statistics that help guide the search. In particular
\begin{itemize}
\item $N(s,a)$: the number of times the action $a$ has been selected from state $s$
\item $W(s,a)$: the cumulative value estimate for each state-action pair
\item $Q(s,a)$: the mean value estimate for each state-action pair
\item $P(s,a)=(1-\varepsilon)p(a|s) + \varepsilon\eta$: the prior policy, computed by evaluating $\prior$. 
Dirichlet noise $\eta$ is added to allow a chance for each move to be chosen.
\item $U(s,a)=Q(s,a) + P(s,a) \frac{\sqrt{\sum_b N(s,b) }} {1+N(s,a)}$ is the \emph{upper confidence bound} \cite{rosin2011multi}. 
The first term exploits moves with high value, while the second term encourages moves that have not selected.
\end{itemize}

When all $M$ iterations are completed, the desired refined policy is proportional to $N(s_0,a)^{1/\tau}$,
where $\tau$ is a smoothing term.

\begin{algorithm}
\caption{Monte Carlo tree search: $\mcts(s_0,\prior,M)$} \label{alg:mcts}
\begin{algorithmic}
\State $N(s,a) \gets 0$ 
\State $Q(s,a) \gets 0$ 
\State $W(s,a) \gets 0$ 
\State visited = $\{ \emptyset \}$
\For{$i = 1,\dots,M $} 
  \State $n\gets0$
  \While{$s_n \notin$ visited}
	\State
	\If{ $\sum_b N(s_n,b)  > 0$}
    \State $a_n^\ast = \arg\max_a Q(s_n,a) + P(s_n,a) \frac{\sqrt{\sum_b N(s_n,b) }} {1+N(s_n,a)}$
	\Else
	\State  $a_n^\ast = \arg\max_a P(s_n,a)$
	\EndIf
    \State $s_{n+1} = \trans(s_n,a_n^\ast)$
    \State $n \gets n+1$
  \EndWhile

  \State $(p,v) = \prior(s_n)$
  \State $P(s_n,a) = (1-\varepsilon)p(a|s_n) + \varepsilon\eta$
  \State visited.append($s_n$)
  \For{$j = 0, \dots, n-1$}
    \State $N(s_j,a) \gets N(s_j,a_j^\ast) + 1$
    \State $W(s_j,a) \gets W(s_j,a_j^\ast) + v$
    \State $Q(s_j,a) \gets Q(s_j,a_j^\ast) / N(s_j,a_j^\ast)$
  \EndFor
\EndFor
\State $\pi^\ast(a|s_0) = N(s_0,a)^{1/\tau} / \sum_b N(s_0,b)^{1/\tau}$
\State \Return $\pi^\ast$
\end{algorithmic}
\end{algorithm}

\subsection{Policy and value network}

We use a convolutional neural network which takes in the game state
and produces a policy and value estimate.
Although the state can be completely characterized by the positions
of the players and the obstacles, the neural network requires more context in order to be
able to generalize to new environments.
We provide the following features as input to the neural network, each of which
is an $m\times m$ image:

\begin{itemize}
\item Obstacles as binary image
\item Player positions, a separate binary image for each player
\item Joint visibility of all pursuers, as a binary image
\item Joint shadow boundaries of all pursuers
\item Visibility from each pursuer's perspective, as a binary image
\item Shadow boundary from each pursuer's perspective
\item Valid actions for each player, as a binary image
\item Each evader's policy according to (\ref{eq:evader-action})
\end{itemize}

AlphaZero \cite{silver2017mastering2} suggests that training the policy and
value networks jointly improves performance.  We use a single network based on U-Net
\cite{ronneberger2015u}, which splits off to give output policy and value.

The input is $m\times m \times C_{\text{in}}$, where $C_{\text{in}}=2+4k_P+2k_E$ and $k_P,k_E$ are the number of
pursuers and evaders, respectively.
The U-Net consists of $\log_2(m)+1$ \emph{down-blocks}, followed by the same
number of \emph{up-blocks}.  
All convolution layers in the down-blocks and up-blocks use size 3 kernels.
Each down-block consists of input, conv, batch
norm, relu, conv, batch norm, residual connection from input, relu, followed by
downsampling with stride 2 conv, batch norm, and relu. A residual connection links
the beginning and end of each block, before downsampling.  The width
of each conv layer in the $l^{th}$ down-block is $l\cdot C_\text{in}$. Each
up-block is the same as the down-block, except instead of downsampling, we use
bilinear interpolation to upsample the image by a factor of 2. The upsampled
result is concatenated with the predownsampled output from the corresponding (same size)
down-block, followed by conv, batch norm, relu. The width of each conv layer in
the up-block is same as those in the down-block of corresponding size.

Then, the network splits into a policy and value head.
The policy head consists of $1\times 1$ conv with width 8, batch norm, relu, and
$1\times 1$ conv with width $k_P$. The final activation layer is a softmax to output $p\in \mathbb{R}^{m\times m \times k_P}$, a policy for each pursuer.
The value head is similar, with $1\times 1$ conv with width 8, batch norm, relu, and
$1\times 1$ conv with width 1. The result passes through a tanh activation and
average pooling to output a scalar $v\in[-1,1]$.

\subsection{Training procedure}
Since we do not have the true value and policy, we cannot train the networks
in the usual supervised fashion. Instead, we use MCTS to generate refined policies,
which serve as the training label for the policy network. 
Multiple games are played with actions selected according to MCTS refined policies.
The game outcomes act as the label for the value for each state in the game.
We train over various maps consisting of 2-7 obstacles, including circles, ellipses, squares,
and tetrahedrons.

More specifically,
let $\prior_\theta(s) $ be the neural network parameterized by $\theta$, which takes a state $s$ as input, and outputs a policy $\vec{\pi}_\theta(s)$ and value $v_\theta(s)$.
Let $s_j(0)$ be the initial positions.
For $j=1,\dots,J$, play the game using MCTS:
\begin{align}
\vec{\pi}_j(a|s_j(k)) &= \text{MCTS}(s_j(k), f_\theta; M)   \\
s_j(k+1) &= \arg\max_a \vec{\pi}_j^\ast(a|s_j(k))
\end{align}

for $k=0,\dots,K_j$. The game ends at 
\beq
K_j = \inf\{k | s_j(k) \in \mathcal{T}_\text{end}\}
\eeq

Then the "true" policy and value are
\begin{align}
\vec{\pi}_j^\ast(k) &= \vec{\pi}_j(\cdot|s_j(k))\\
v^\ast_j(k)&= 
\begin{cases} 
1  & K_j > K_\text{max}  \\
-1  & \text{otherwise}
\end{cases}
\end{align}

The parameters $\theta$ of the neural network
are updated by stochastic gradient descent (SGD) on the loss function:
\beq 
\min_\theta \sum_{j=1}^J \sum_{k=0}^{K_j}  L_{\text{policy}} & \Big(\vec{\pi}_\theta(s_j(k)) , \vec{\pi}_j^\ast(k) \Big) + L_\text{value} \Big(v_\theta(s_j(k)),v_j^\ast(k)\Big)\\
L_\text{policy}(\vec{p},\vec{q}) &= -\vec{p} \cdot \log \vec{q} \\
L_\text{value}(p,q) &= (p-q)^2 \\
\eeq

We use a learning rate of $0.001$ and the Adam optimizer \cite{kingma2014adam}.

\subsection{Numerical results}

A key difficulty in learning a good policy for the pursuer
is that it requires a good evader. If the evader is static,
then the pursuer can win with any random policy.

During training and evaluation, the game is played with the evader moving
according to (\ref{eq:evader-action}).  Although all players move
simultaneously, our MCTS models each team's actions sequentially, with the
pursuers moving first. This is conservative towards the pursuers, since the
evaders can counter.

We train using a single workstation with 2 Intel Xeon CPU E5-2620 v4 2.10GHz
processors and a single NVidia 1080-TI GPU.  For simplicity, $\pspeed$ and $\espeed$ are
constant, though it is straightforward to have spatially varying
velocity fields.  We use a gridsize of $m=16$. We set $K_{\text{max}}=100$ and
$M=1000$ MCTS iterations per move.  One step of training consists of playing
$J=64$ games and then training the neural network for 1 epoch based on training
data for the last 10 steps of training.  Self-play game data is generated in
parallel, while network training is done using the GPU with batch size 128. The total
training time is 1 day.

The training environments consist of between 2 to 6 randomly oriented obstacles,
each uniformly chosen from the set of ellipses, diamonds, and rectangles. We emphasize
that the environments shown in the experiments are not in the training set.

We compare our trained neural network against uniform random and dirichlet
noise-based policies, as well as the local policies from
section~\ref{sec:locally}.  In order to draw a fair comparison, we make sure
each action requires the same amount of compute time.  Each MCTS-based move in
the 2 player game takes 4 secs while the multiplayer game takes about 10 secs
per move, on average.  Since the noise-based policies require less overhead,
they are able to use more MCTS iterations.  The shadow strategies become very
expensive as more players are added. For the 1v1 game, we use $\hat{M}=1000$,
while the 2v2 game can only run for $\hat{M}=250$ in the same amount of time as
the Neural Net. 
Specifically,
\begin{itemize}
\item Distance strategy
\item Shadow strategy
\item Blend strategy
\item $\mcts(\cdot,\prior_\text{distance},1000)$ where $\prior_\text{distance}(s) = (p_\text{distance},0)$.
\item $\mcts(\cdot,\prior_\text{shadow},\hat{M})$ where $\prior_\text{shadow}(s) = (p_\text{shadow},0)$.
\item $\mcts(\cdot,\prior_\text{blend},\hat{M})$ where $\prior_\text{blend}(s) = (p_\text{blend},0)$.
\item $\mcts(\cdot,\prior_\mu,2000)$ where $\prior_\nu(s) = (\text{Uniform},0)$.
\item $\mcts(\cdot,\prior_\eta,2000)$ where $\prior_\eta(s) = (\text{Dir}(0.3),0)$.
\item $\mcts(\cdot,\prior_\theta,1000)$ where $\prior_\theta$ is the trained Neural Network
\end{itemize}

\subsection*{Two players}
\begin{figure}[hptb]
\centering
\includegraphics[width=.45\textwidth]{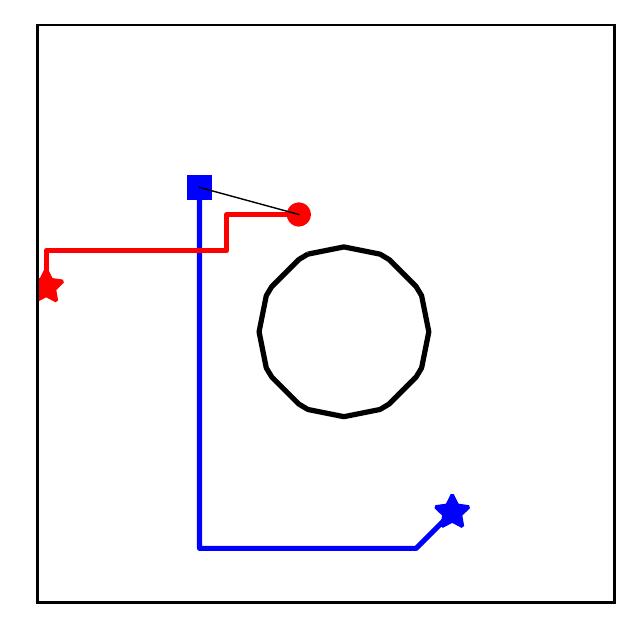}
\includegraphics[width=.45\textwidth]{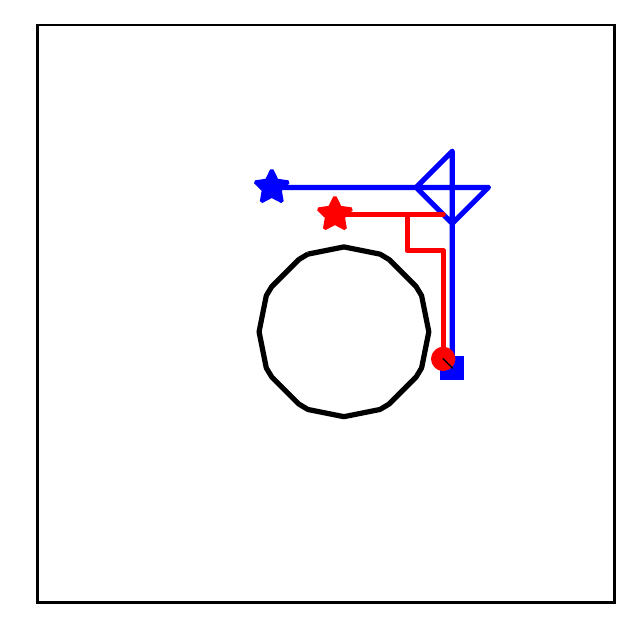}
\includegraphics[width=.45\textwidth]{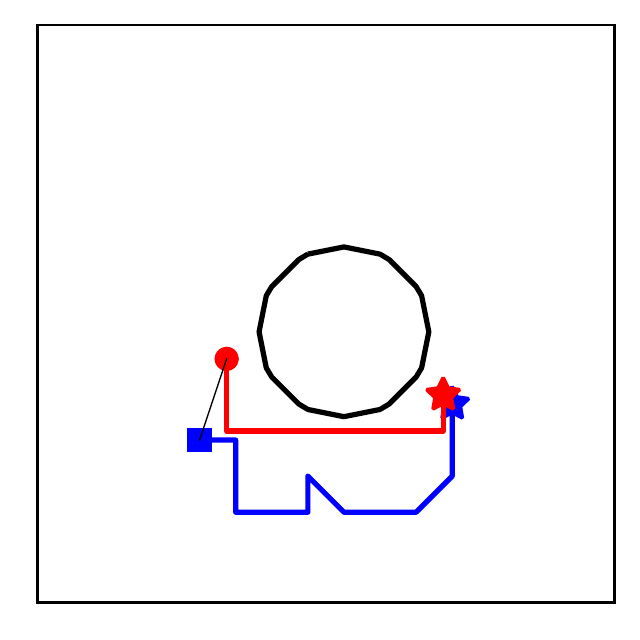}
\includegraphics[width=.45\textwidth]{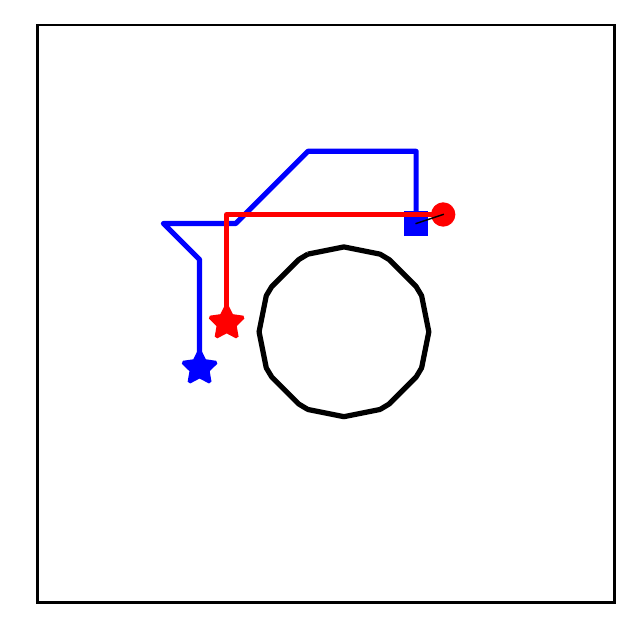}
\caption{Snapshots of the trajectory for the Neural Net pursuer around a circular obstacle. 
The pursuer (blue) tracks the evader (red) while maintaining a safe distance.
View from left to right, top to bottom.
Stars indicate the initial positions, and the black line (of sight) connects the players at the end of each time interval.} 
\label{fig:nnet-traj-circle}

\end{figure}

\begin{figure}[hptb]
\centering
\includegraphics[width=.45\textwidth]{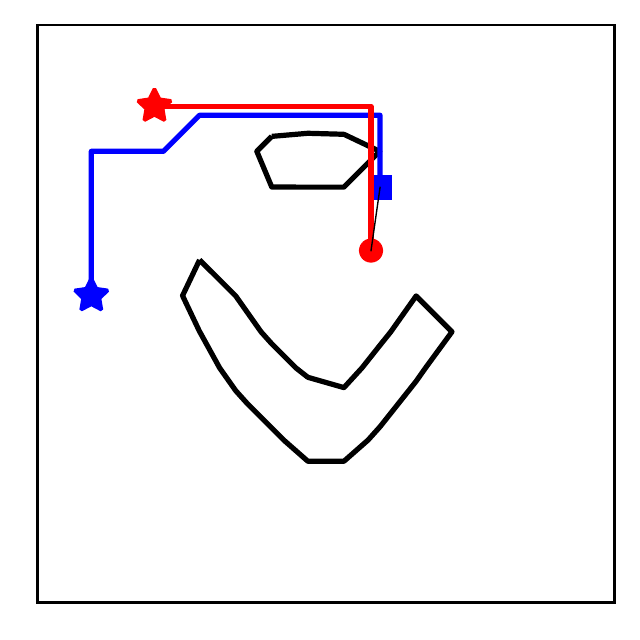}
\includegraphics[width=.45\textwidth]{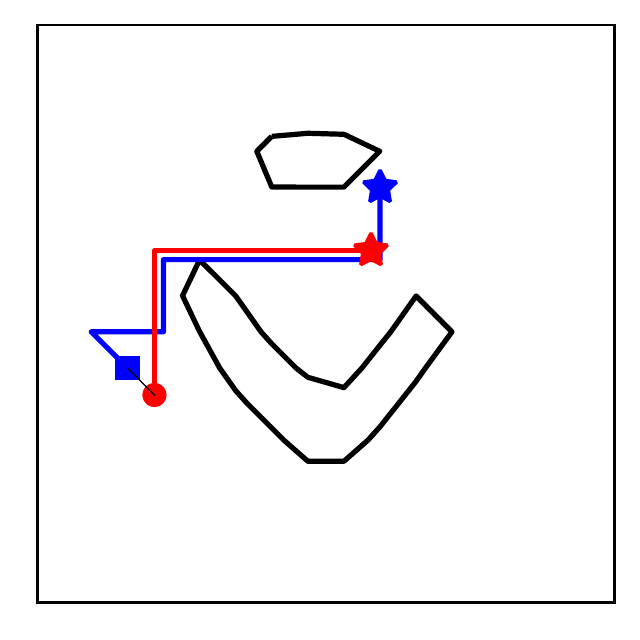}
\includegraphics[width=.45\textwidth]{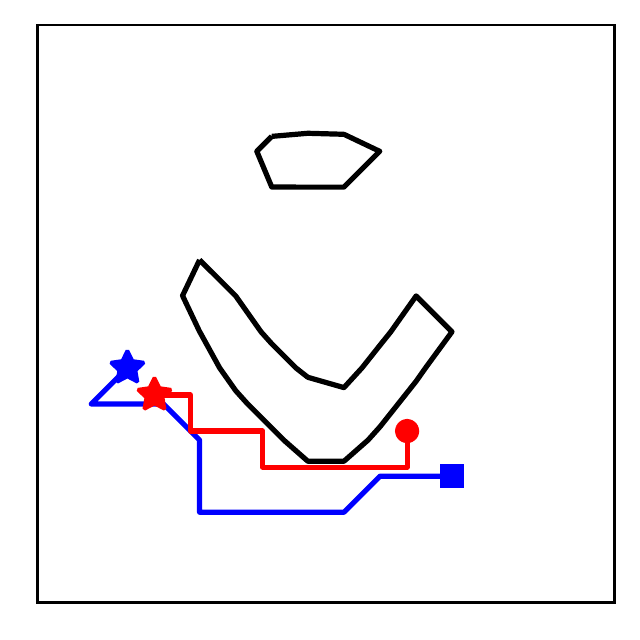}
\includegraphics[width=.45\textwidth]{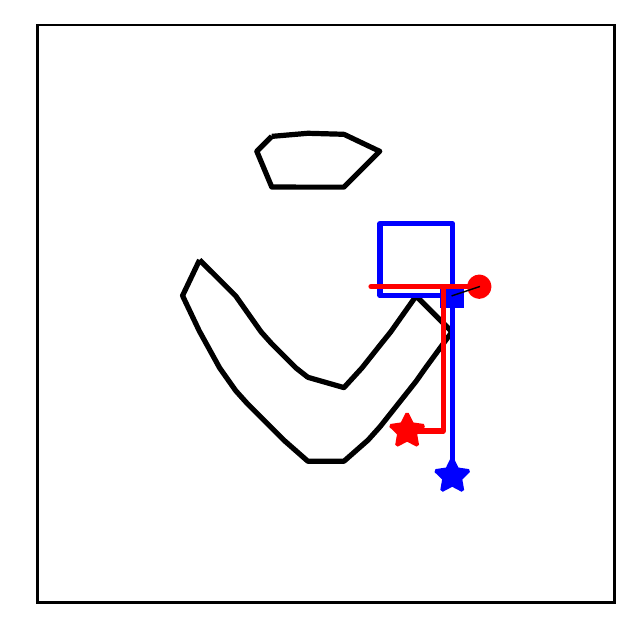}
\caption{Trajectory for the Neural Net pursuer against an adversarial human evader on a map with two obstacles. The pursuer
transitions between following closely, and leaving some space, depending on the shape of the obstacle.} 
\label{fig:nnet-traj-v}

\end{figure}

As a sanity check, we show an example on a single circular obstacle with a
single pursuer and single evader.  As we saw from the previous section, the
pursuer needs to be faster in order to have a chance at winning. We let
$\pspeed=2$ and $\espeed=1$.
Figure~\ref{fig:nnet-traj-circle} shows an example trajectory using Neural Net.
The neural network model gives reasonable policies. 
Figure~\ref{fig:nnet-traj-v} shows an adversarial human evader playing against the Neural Net pursuer,
on a map with two obstacles. The pursuer changes strategies depending on the shape of the obstacle.
In particular, near the corners of the "V" shape, it maintains a safe distance rather than blindly following
the evader.

%\begin{figure}[hptb]
%\centering
%\includegraphics[width=.45\textwidth]{figures/traj/traj_dice_seed_05_m_16_vP_2_vE_1_mcts_1000_nnet_vs_shadow_000.pdf}
%\includegraphics[width=.45\textwidth]{figures/traj/traj_dice_seed_05_m_16_vP_2_vE_1_mcts_1000_nnet_vs_shadow_001.pdf}
%\includegraphics[width=.45\textwidth]{figures/traj/traj_dice_seed_05_m_16_vP_2_vE_1_mcts_1000_nnet_vs_shadow_002.pdf}
%\includegraphics[width=.45\textwidth]{figures/traj/traj_dice_seed_05_m_16_vP_2_vE_1_mcts_1000_nnet_vs_shadow_003.pdf}
%\caption{Snapshots of the trajectory for the NNet pursuer around a circular obstacle. View from left to right, top to bottom.
%Stars indicate the initial positions, and the black line connects the players at the end of each time interval.} 
%\label{fig:nnet-traj-dice}
%
%\end{figure}

\begin{figure}[hptb]
\centering
\includegraphics[width=.45\textwidth]{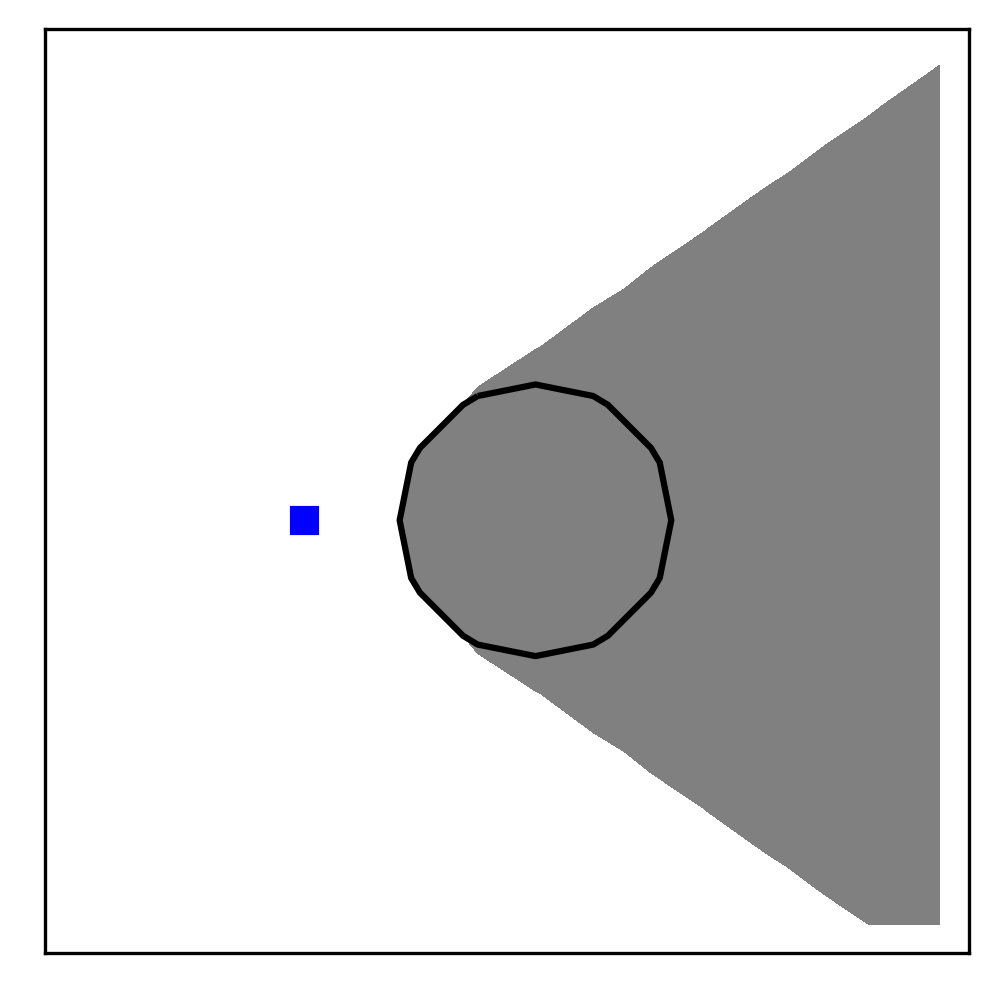}
\includegraphics[width=.45\textwidth]{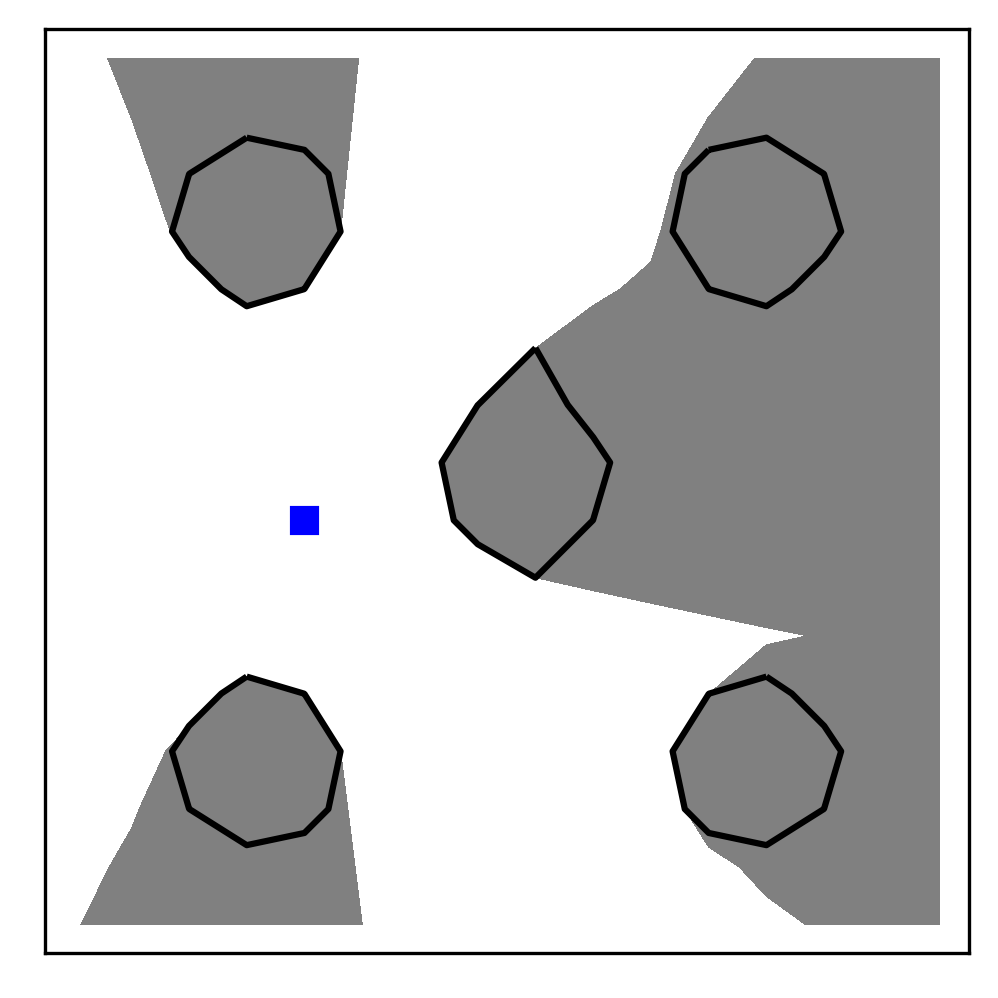}
\caption{Setup for computing a slice of the value function for the circular obstacle (left) and 5 obstacle map (right). The pursuer's initial position is fixed (blue) while the
evader's changes within the free space.} 
\label{fig:value-slice-setup}
\end{figure}

In order to do a more systematic comparison, we run multiple games
over the same map and report the game time statistics for each method.
We fix the pursuer's position at $(1/2,1/4)$ and vary the evader's initial location within the free space.
Figure~\ref{fig:value-slice-setup} shows the setup for the two maps considered for the statistical studies in this section.
One contains a single circle in the center, as we have seen previously. The other one contain 5 circular obstacles,
though the one in the center has some extra protrusions.

\begin{figure}[hptb]
\centering
\includegraphics[width=.3\textwidth]{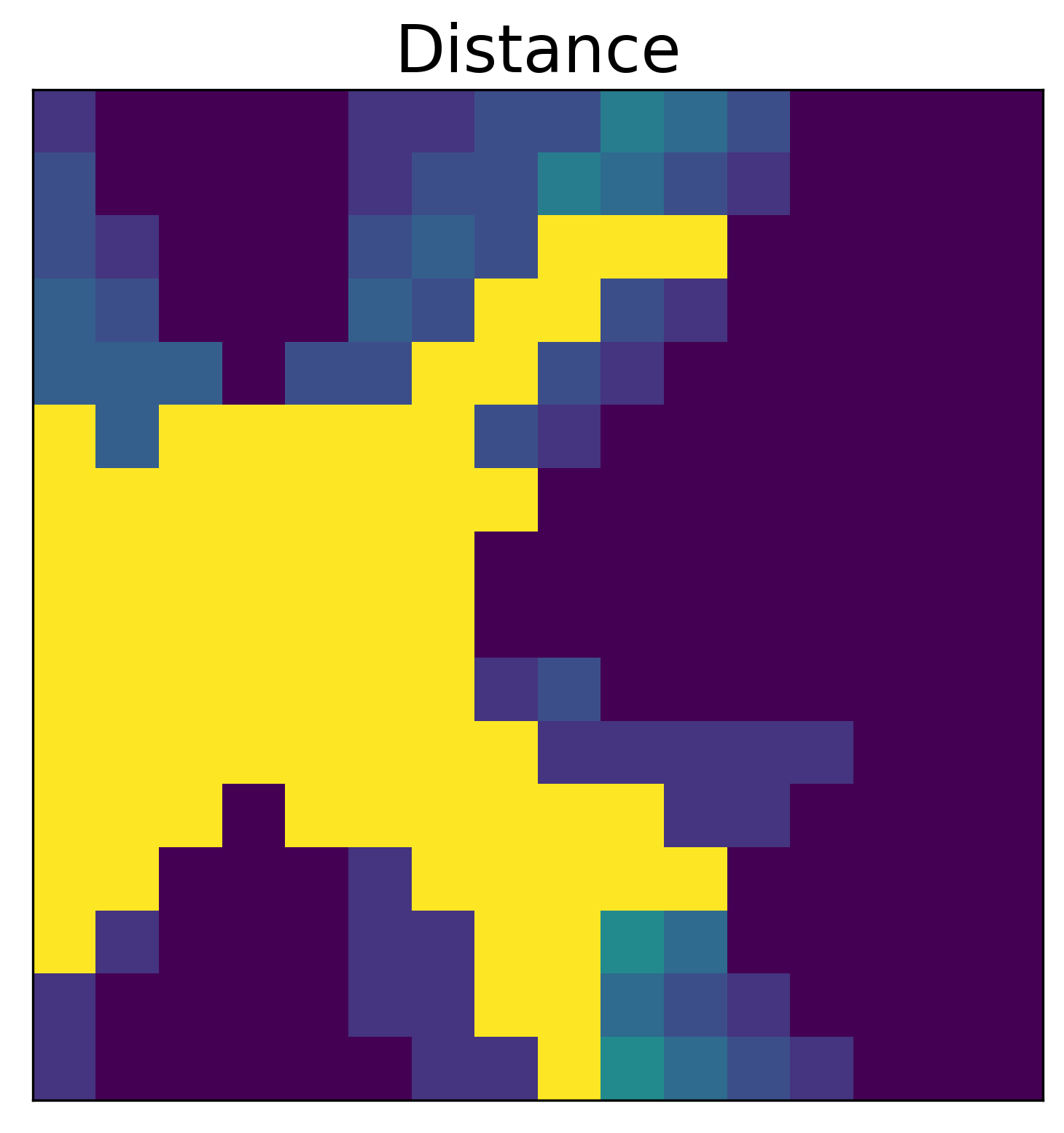}
\includegraphics[width=.3\textwidth]{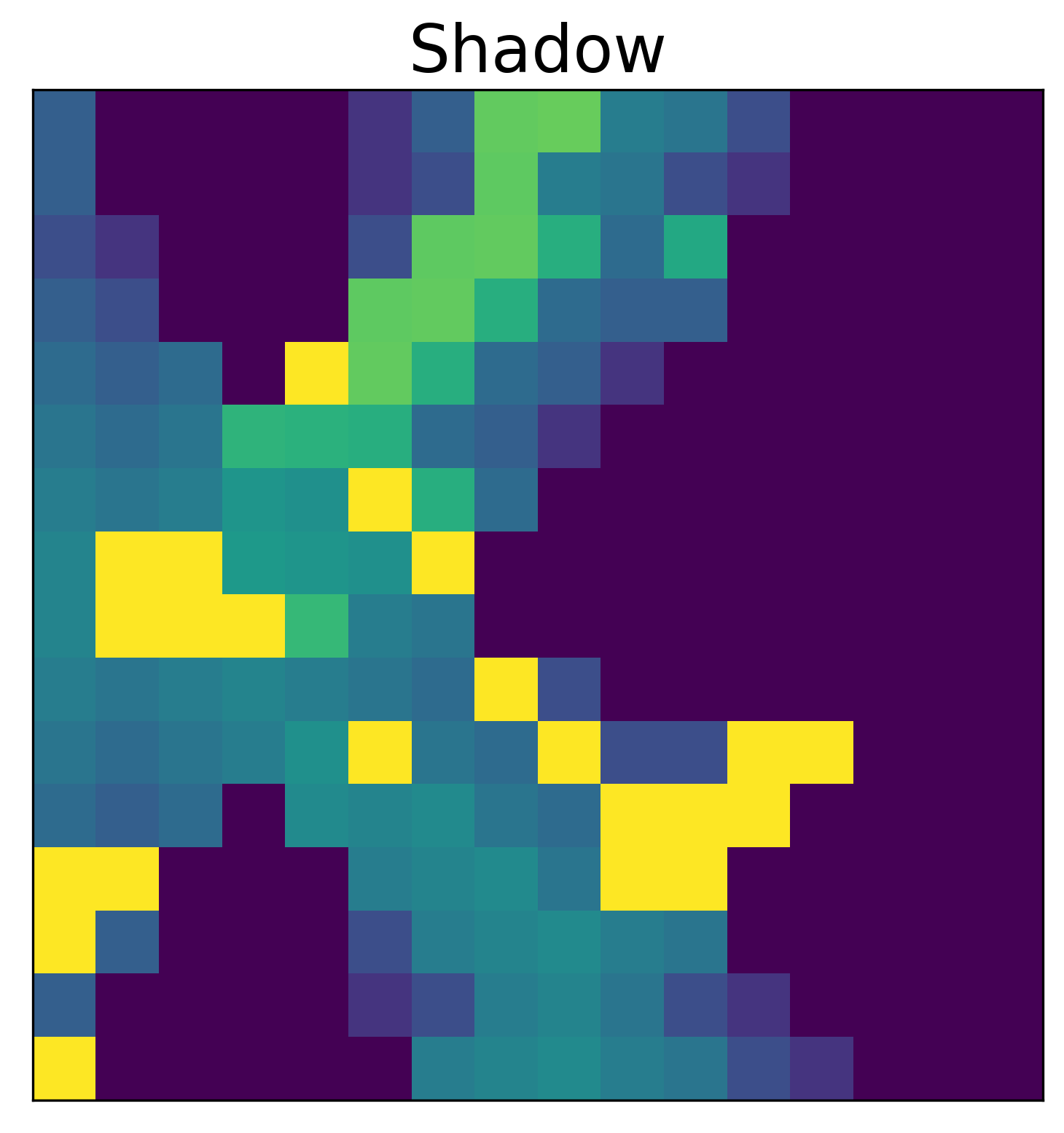}
\includegraphics[width=.3\textwidth]{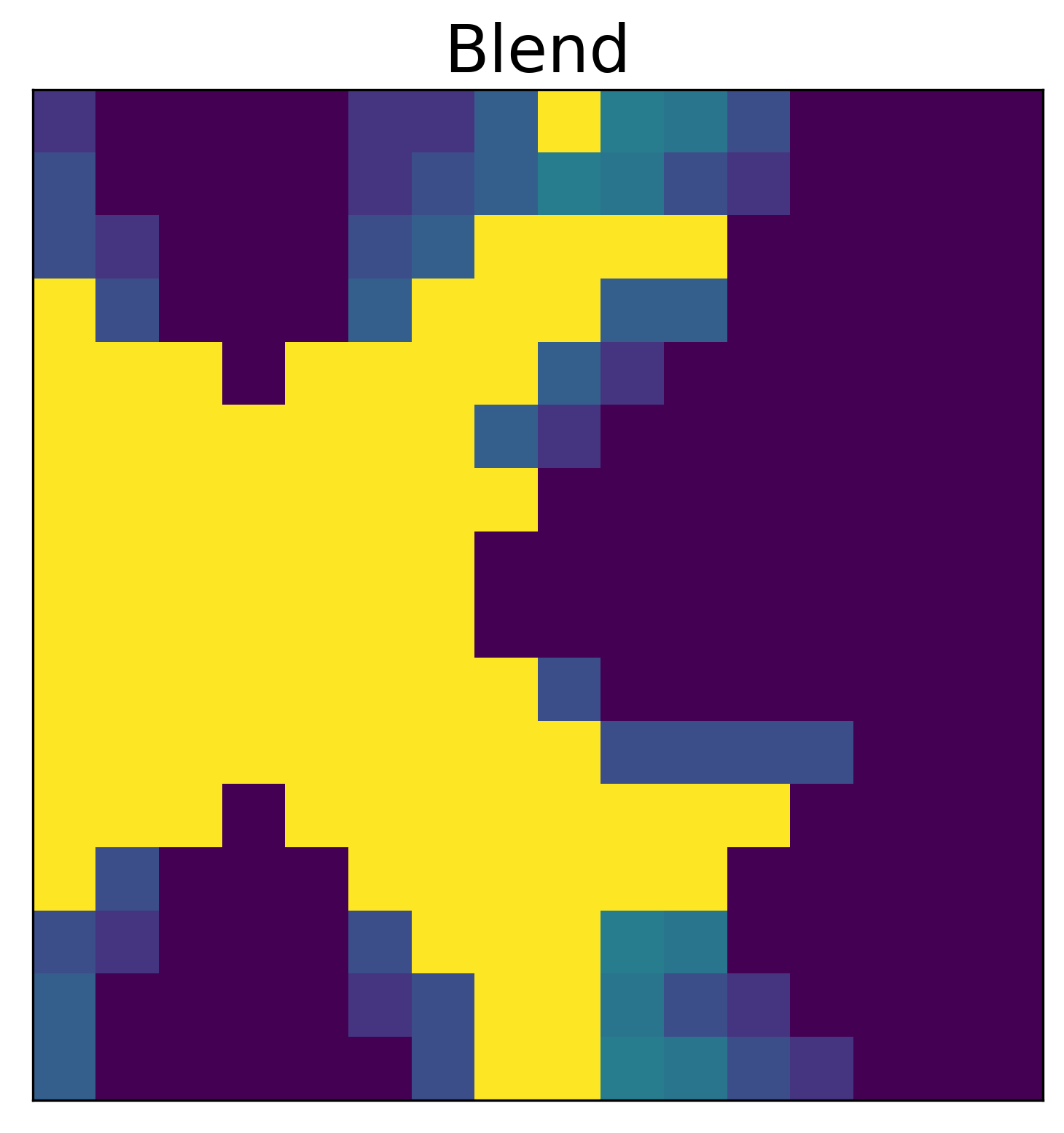}
\includegraphics[width=.3\textwidth]{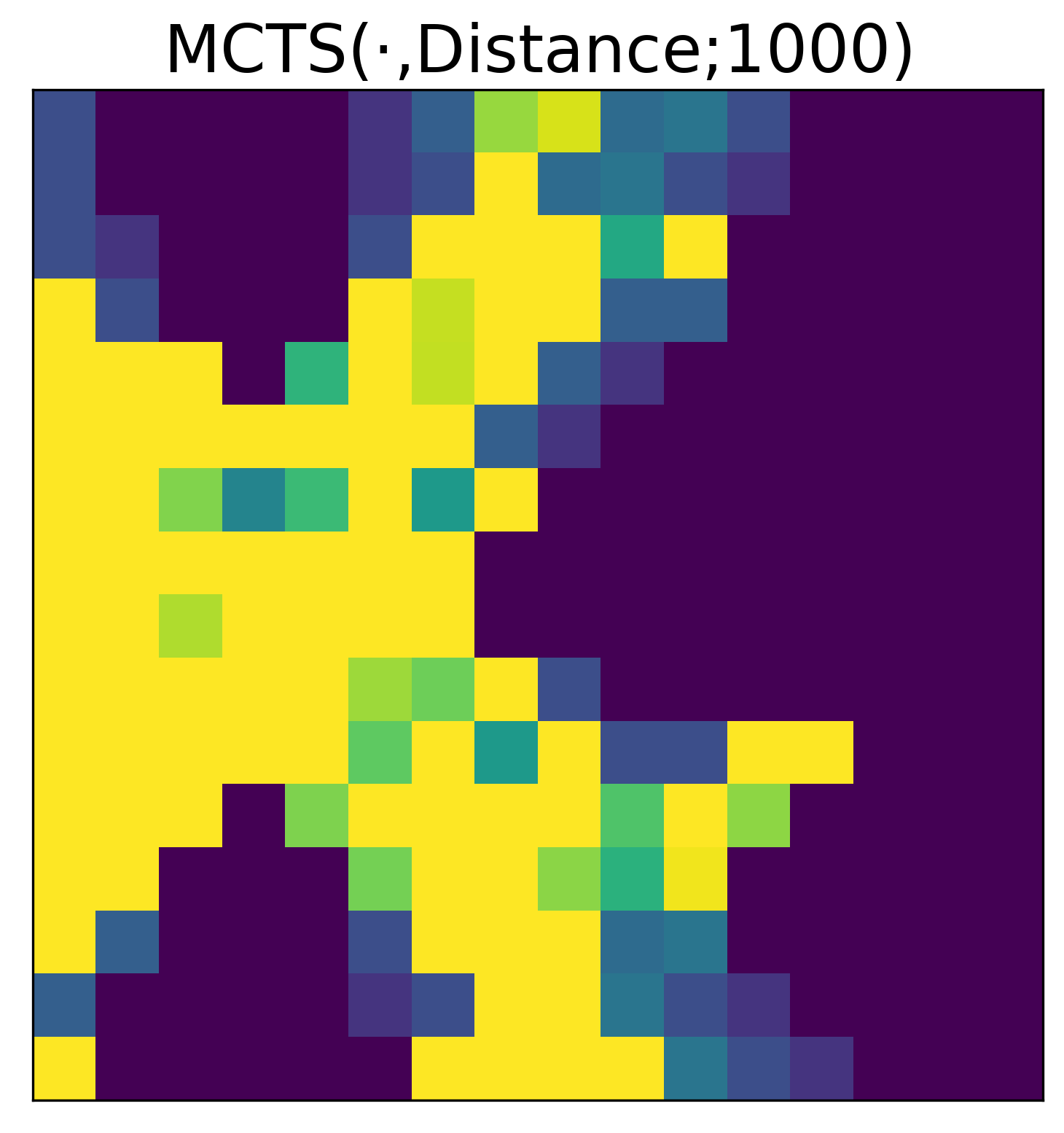}
\includegraphics[width=.3\textwidth]{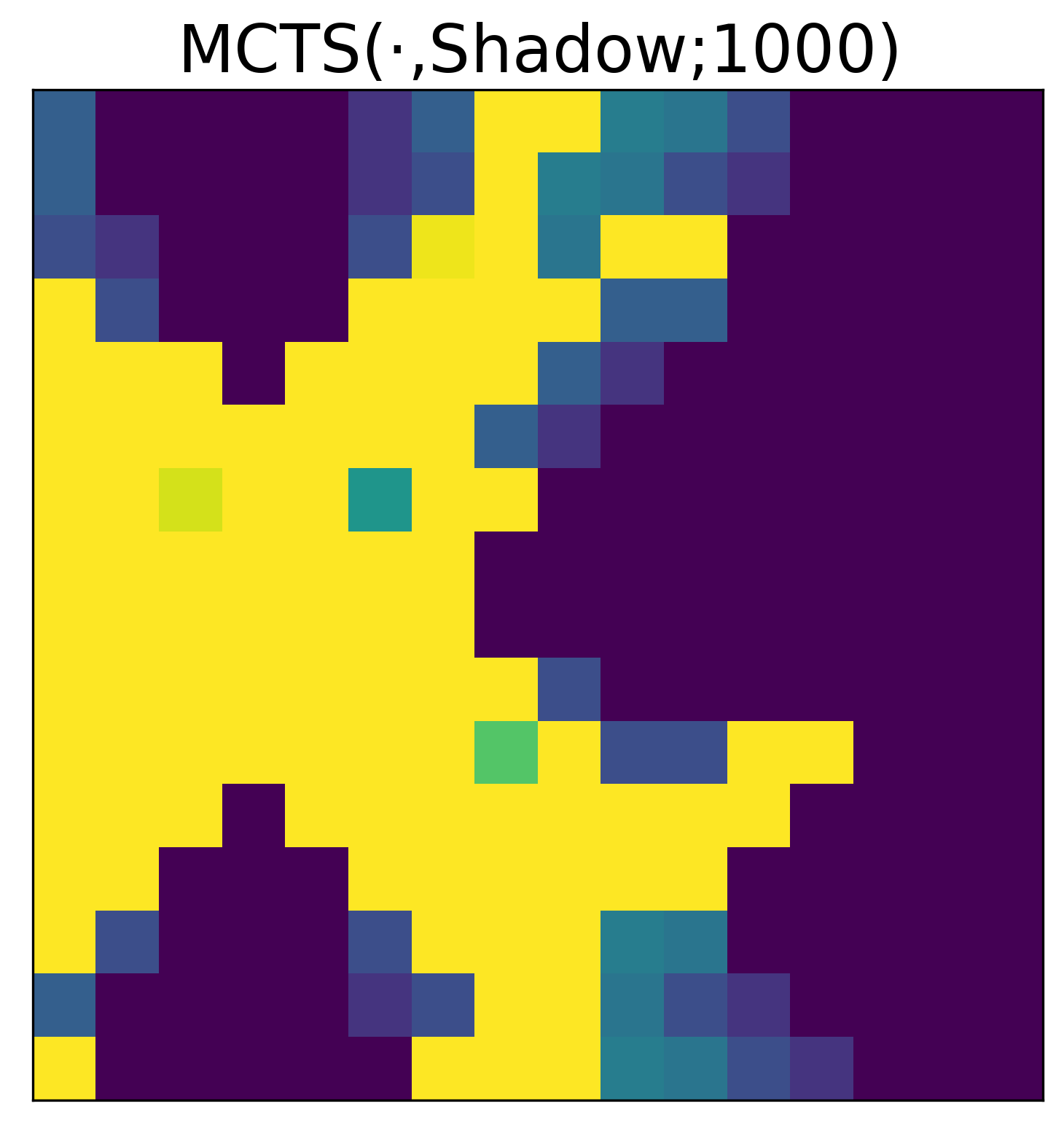}
\includegraphics[width=.3\textwidth]{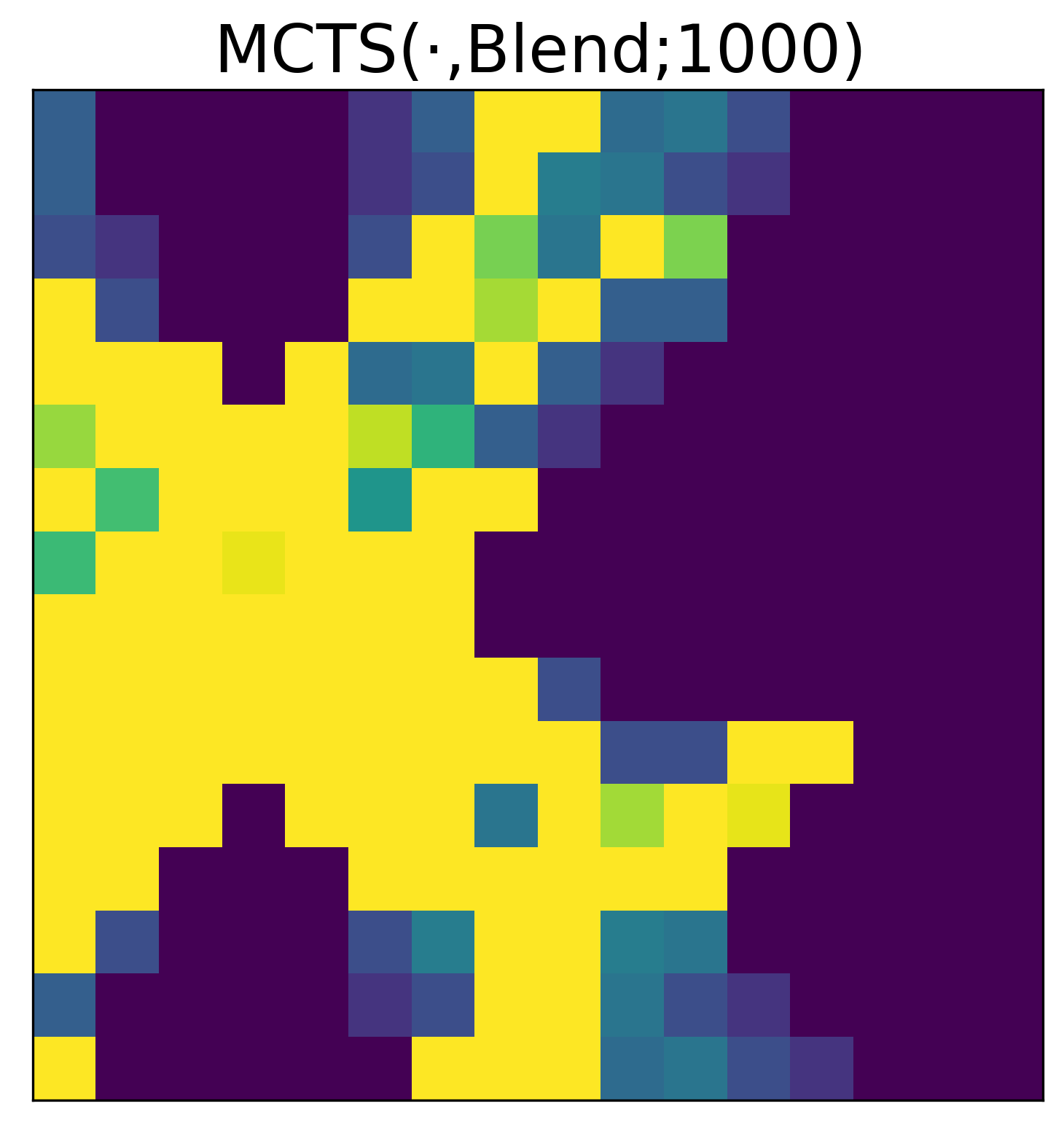}
\includegraphics[width=.3\textwidth]{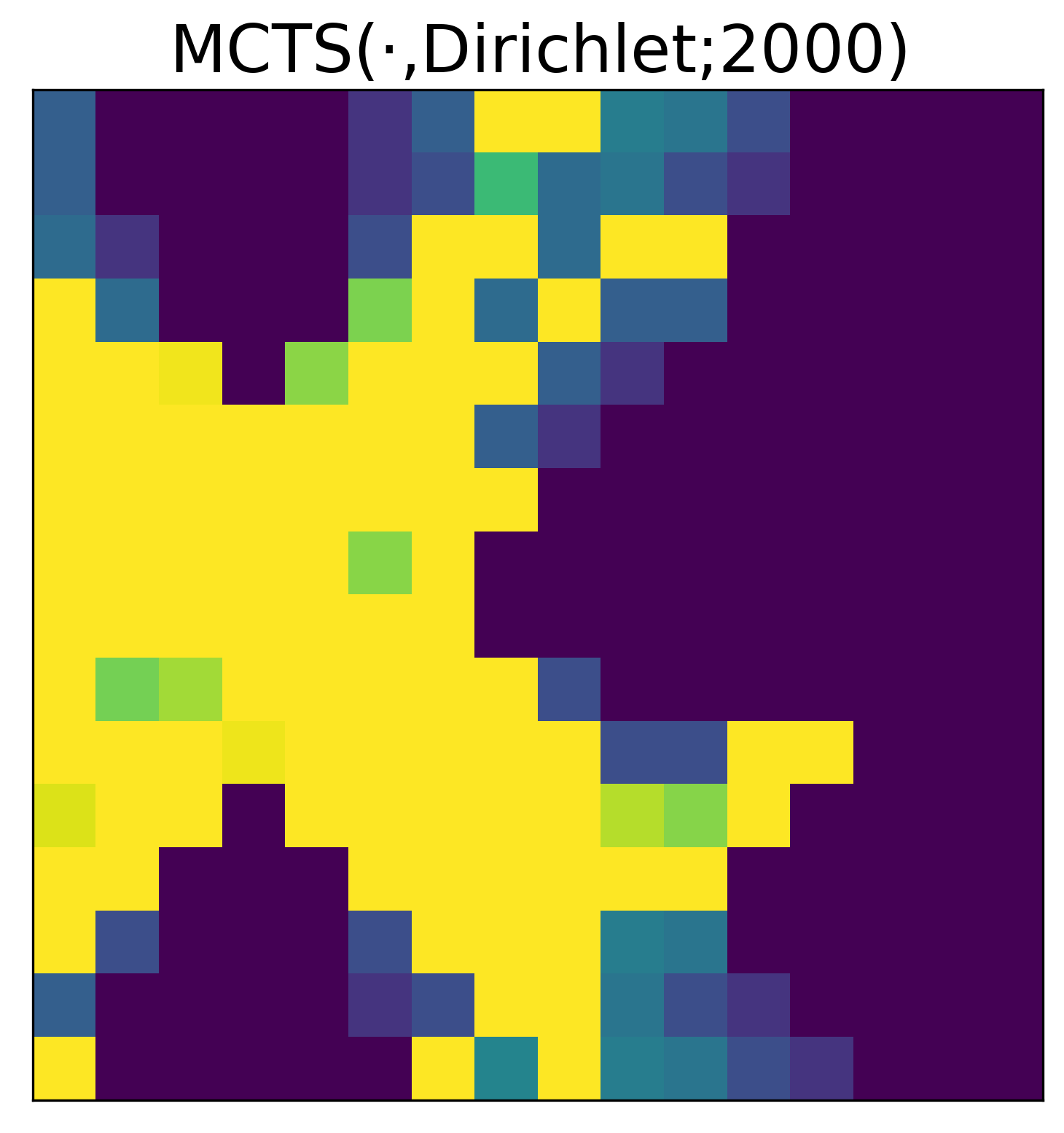}
\includegraphics[width=.3\textwidth]{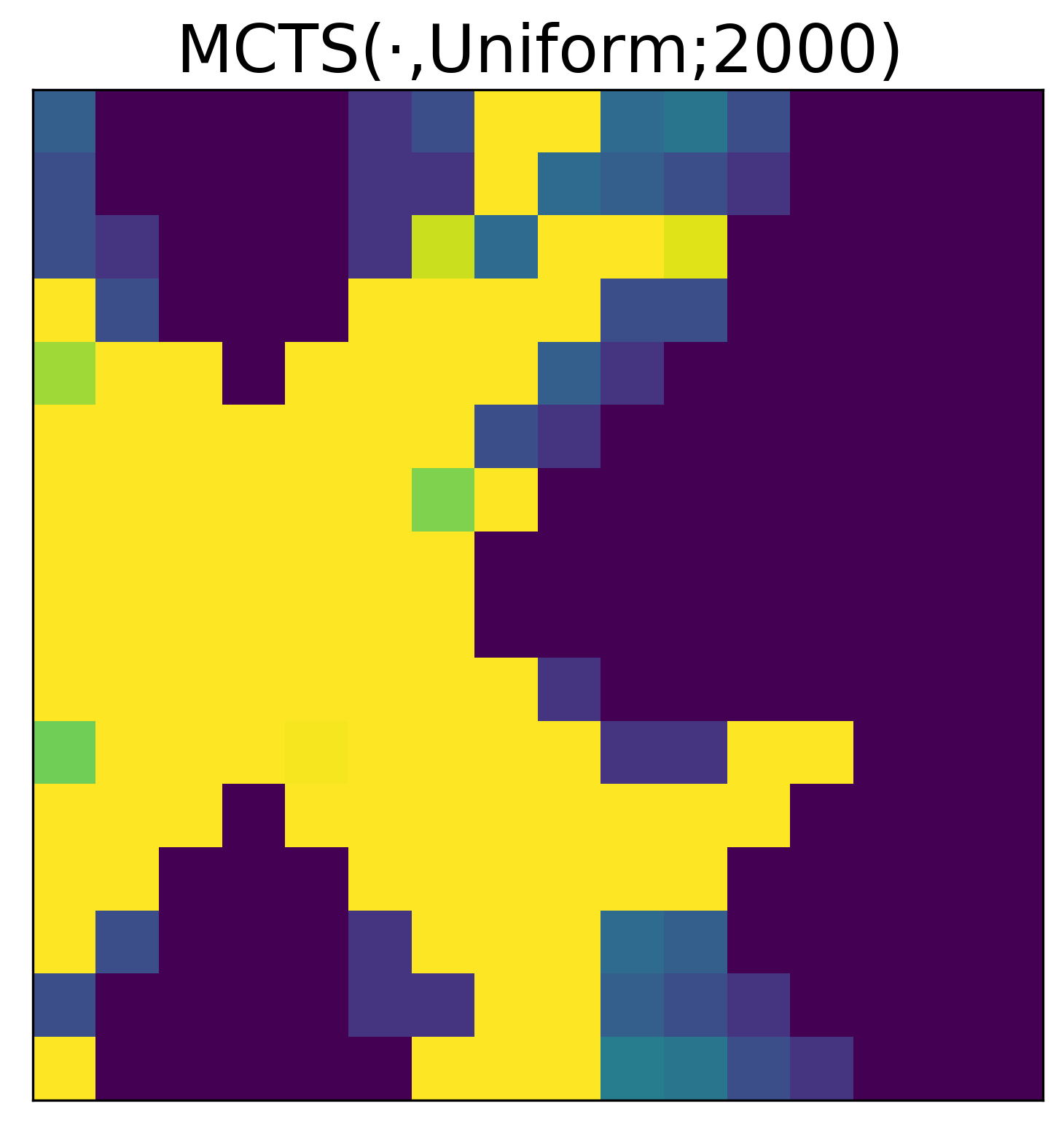}
\includegraphics[width=.3\textwidth]{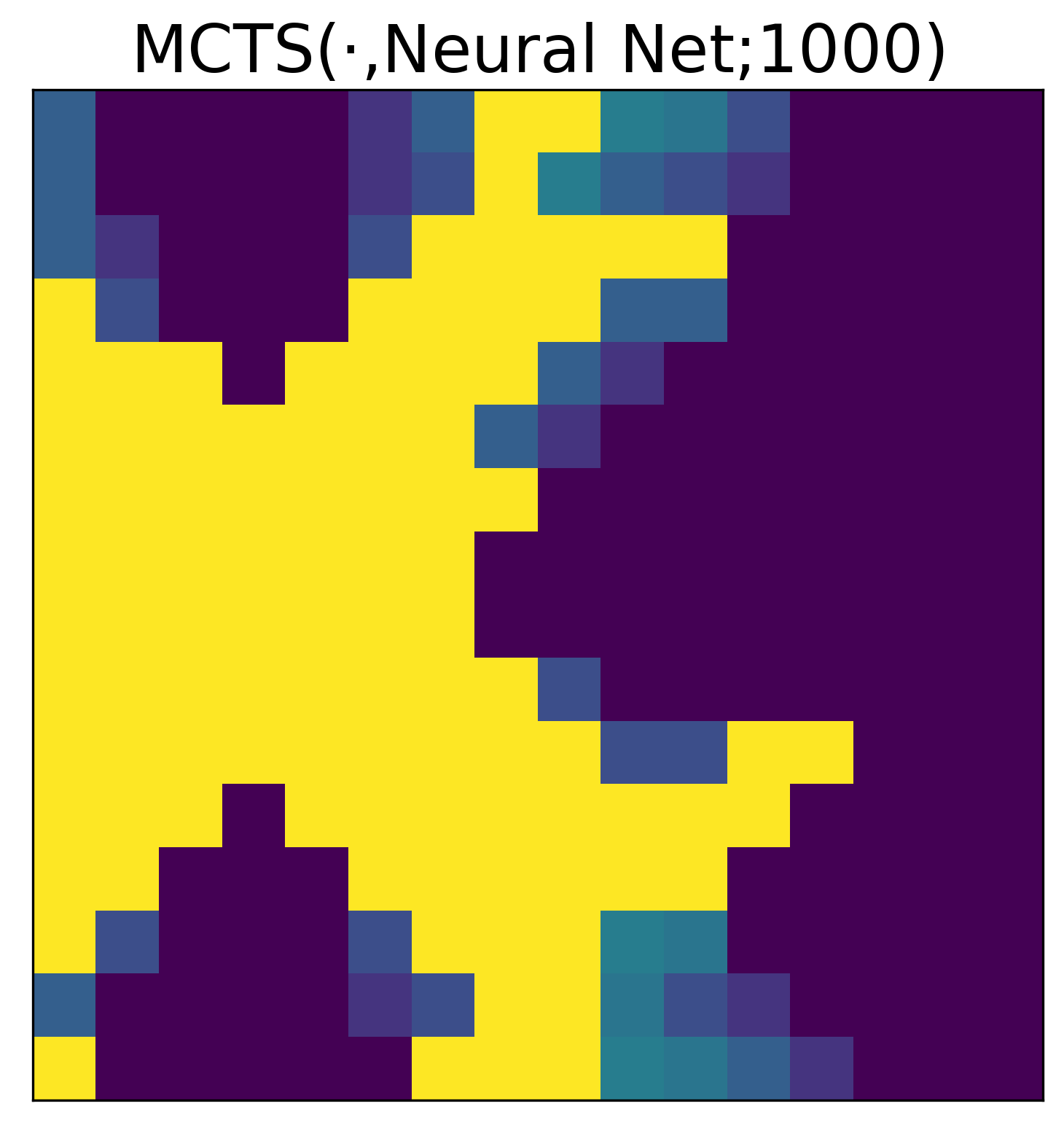}
\caption{One slice of the "value" function for single pursuer, single evader game with 5 obstacles. Bright spots
indicate that the pursuer won the game if that pixel was the evader's initial position.}
\label{fig:value-slice-1v1dice}
\end{figure}

Figure~\ref{fig:value-slice-1v1dice} shows an image corresponding the the length of the game
for each evader position; essentially, it is a single slice
of the value function for each method.
Table~\ref{tab:1v1dice} shows the number of games won. Shadow strategy particularly benefits
from using MCTS for policy improvements, going from 16\% to 67.15\% win rate. Our neural network
model outperforms the rest with a 70.8\% win rate.

\begin{table}[]
\caption{Game statistics for the 1 pursuer vs 1 evader game with 5 circular obstacles, where $\pspeed=2$ and $\espeed=1$.\vspace{1em}}
\label{tab:1v1dice}
\begin{tabular}{l|ccc}
Method          & Win \% (137 games) & Average game time                 \\   % & Average loss time \\
\hline
Distance                         &52.55       & 53.56         \\   % & 4.7             \\
Shadow                           &16.06       & 22.36         \\   % & 7.2             \\
Blend                            &63.50       & 64.45         \\   % & 7.3             \\
MCTS($\cdot$, Distance; 1000)    &55.47       & 62.40         \\   % & 12.2            \\
MCTS($\cdot$, Shadow; 1000)      &67.15       & 69.45         \\   % & 6.3             \\
MCTS($\cdot$, Blend; 1000)       &58.39       & 63.27         \\   % & 4.4             \\
MCTS($\cdot$, Uniform; 2000)     &60.58       & 65.84         \\   % & 5.3             \\
MCTS($\cdot$, Dirichlet; 2000)   &65.69       & 69.02         \\   % & 7.0             \\
MCTS($\cdot$, Neural Net; 1000)  &70.80       & 71.61         \\   % & 4.7            
\end{tabular}                     
\end{table}
% OLD, somehow incorrect
%66.42                     & 67.30
%5.84                      & 20.50
%73.72                     & 74.45
%75.18                     & 78.69
%80.29                     & 82.02
%73.72                     & 77.18
%60.58                     & 65.83
%65.69                     & 69.02
%83.21                     & 83.65

\subsection*{Multiple players}
\begin{figure}[phtb]
\centering
\includegraphics[width=.45\textwidth]{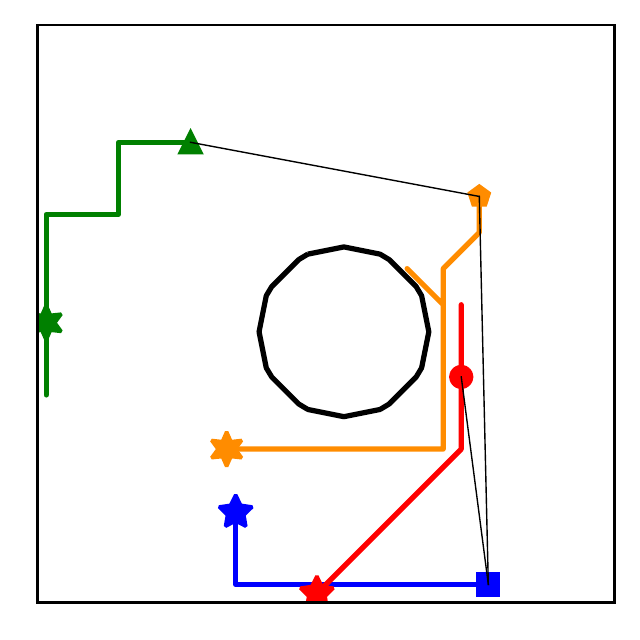}
\includegraphics[width=.45\textwidth]{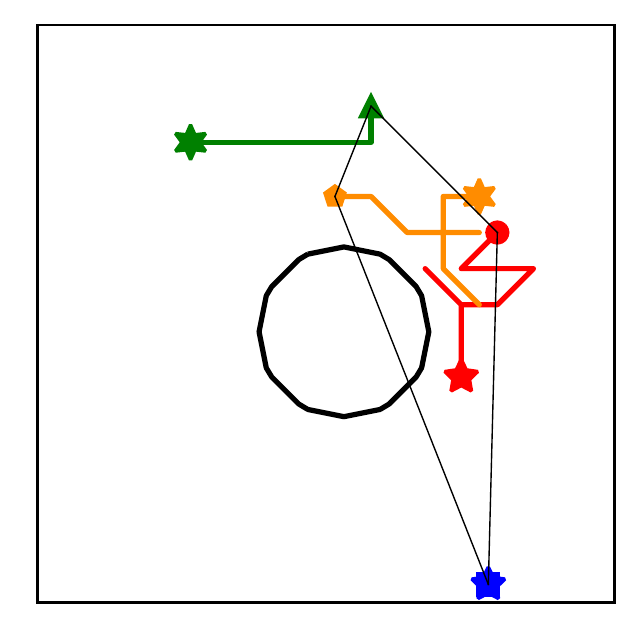}
\includegraphics[width=.45\textwidth]{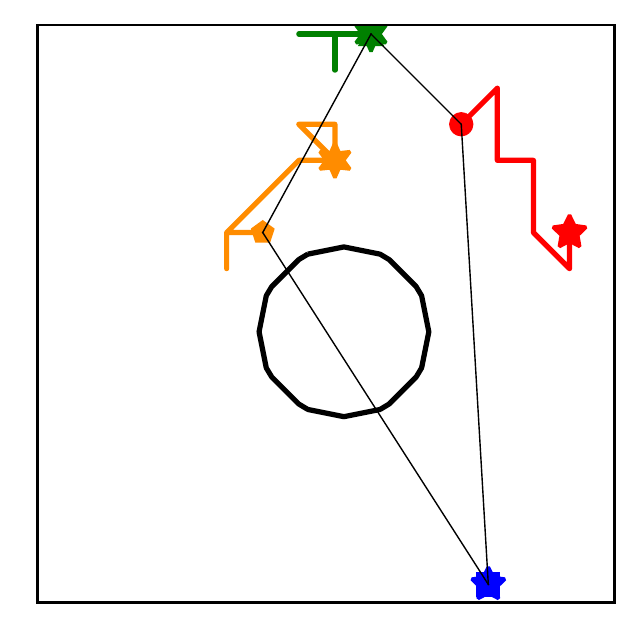}
\includegraphics[width=.45\textwidth]{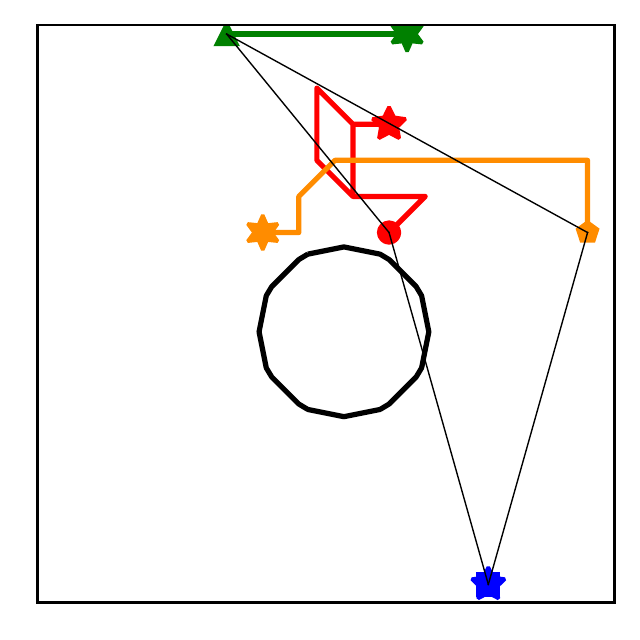}
\caption{Trajectories for the multiplayer game played using NNet around a circle. Pursuers are blue and green, while evaders are red and orange.
Blue has learned the tactic of remaining stationary in the corner, while green manages the opposite side.
The evaders movements are sporadic because there is no chance of winning; there are no shadows in which to hide.} 
\label{fig:nnet-traj-circle-multi}

\end{figure}

Next, we consider the multiplayer case with 2 pursuers and 2 evaders on a circular obstacle map
where $\pspeed=2$ and $\espeed=2$. 
Even on a $16\times16$ grid, the computation of the corresponding feedback value function 
would take several days.
Figure~\ref{fig:nnet-traj-circle-multi} shows a sample trajectory. Surprisingly, the neural network has learned
a smart strategy. Since there is only a single obstacle, it is sufficient for each
pursuer to guard one opposing corner of the map. Although all players have the same speed, it is possible
to win.

\begin{figure}[phtb]
\centering
\includegraphics[width=.3\textwidth]{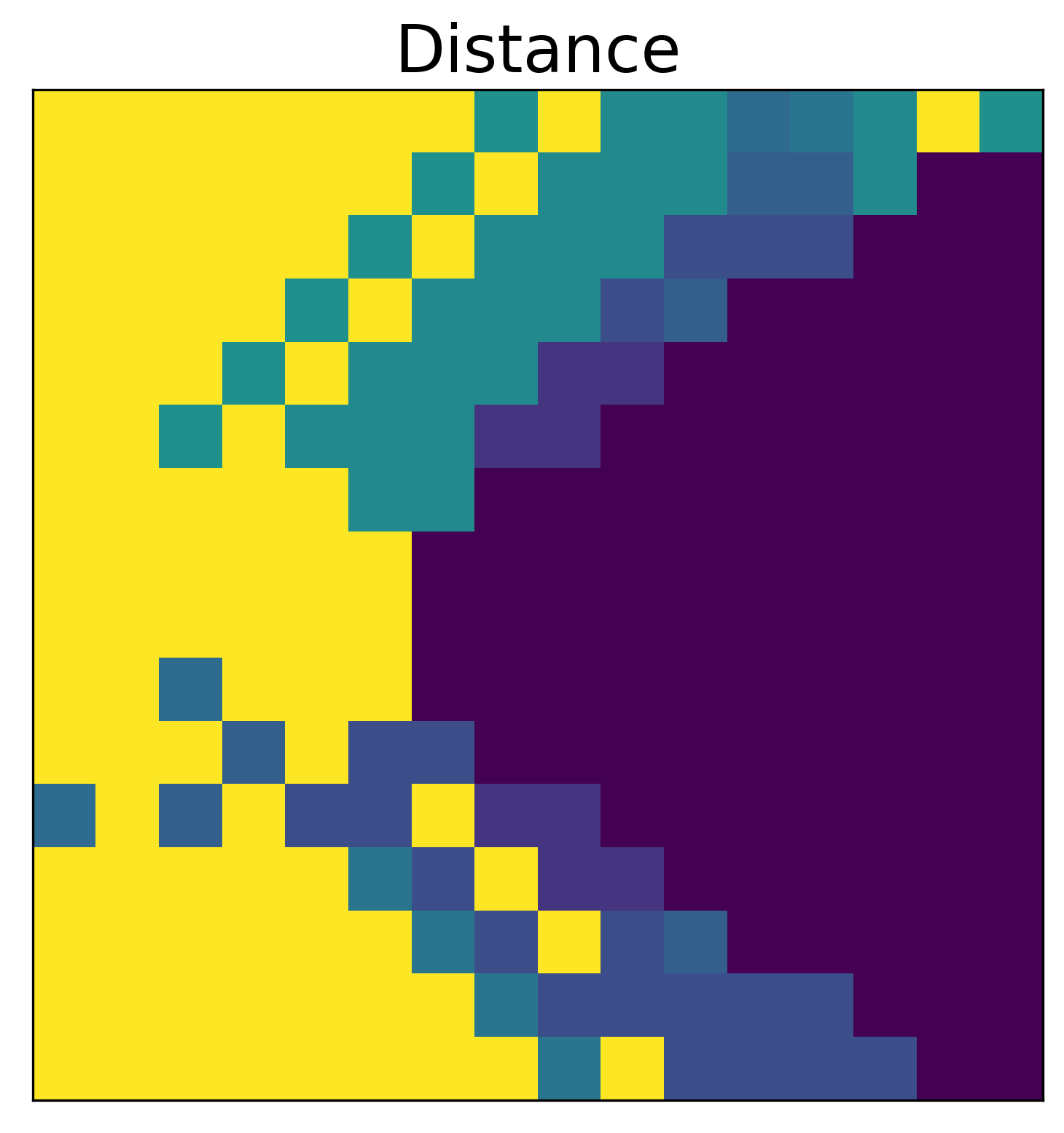}
\includegraphics[width=.3\textwidth]{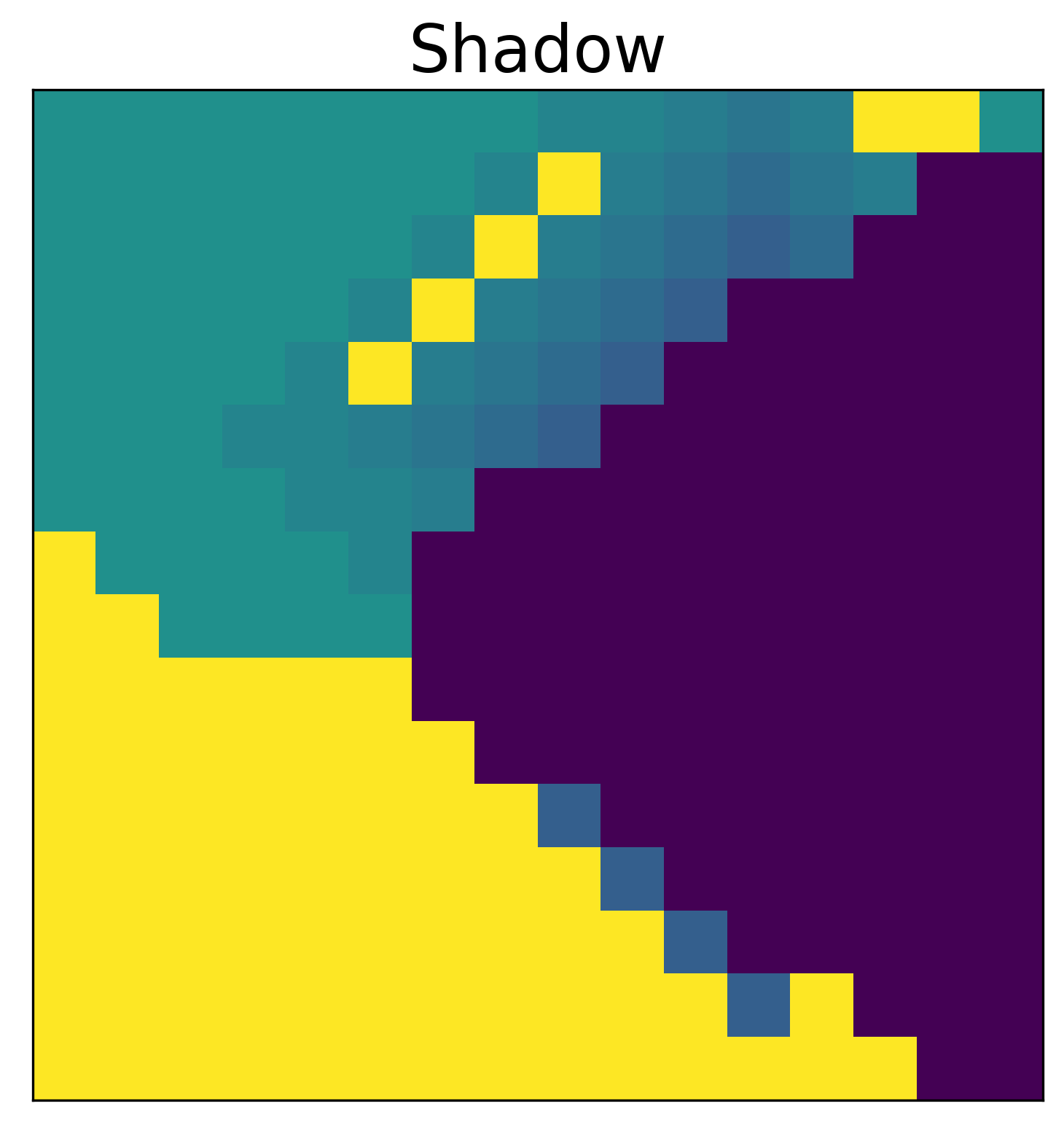}
\includegraphics[width=.3\textwidth]{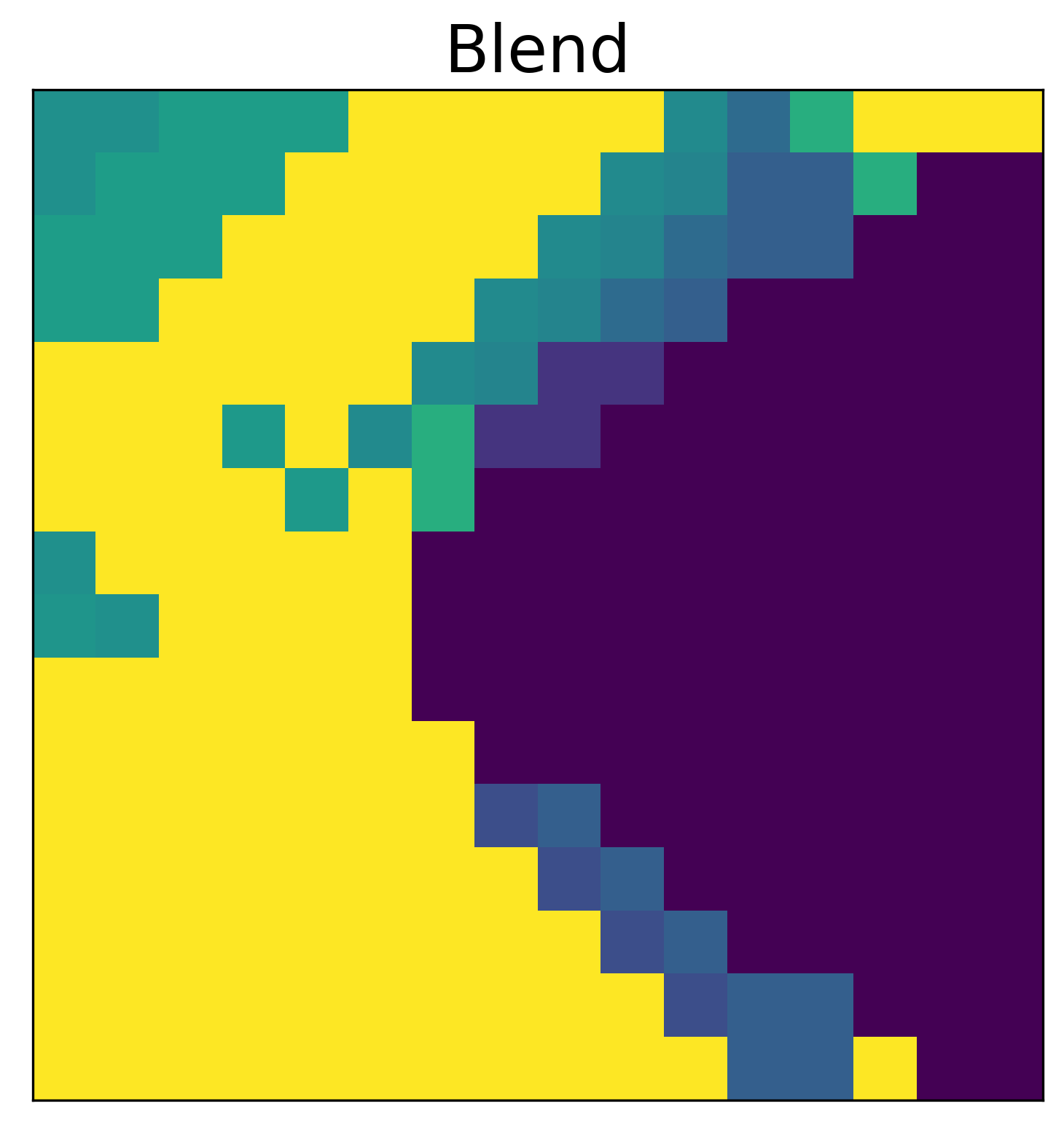}
\includegraphics[width=.3\textwidth]{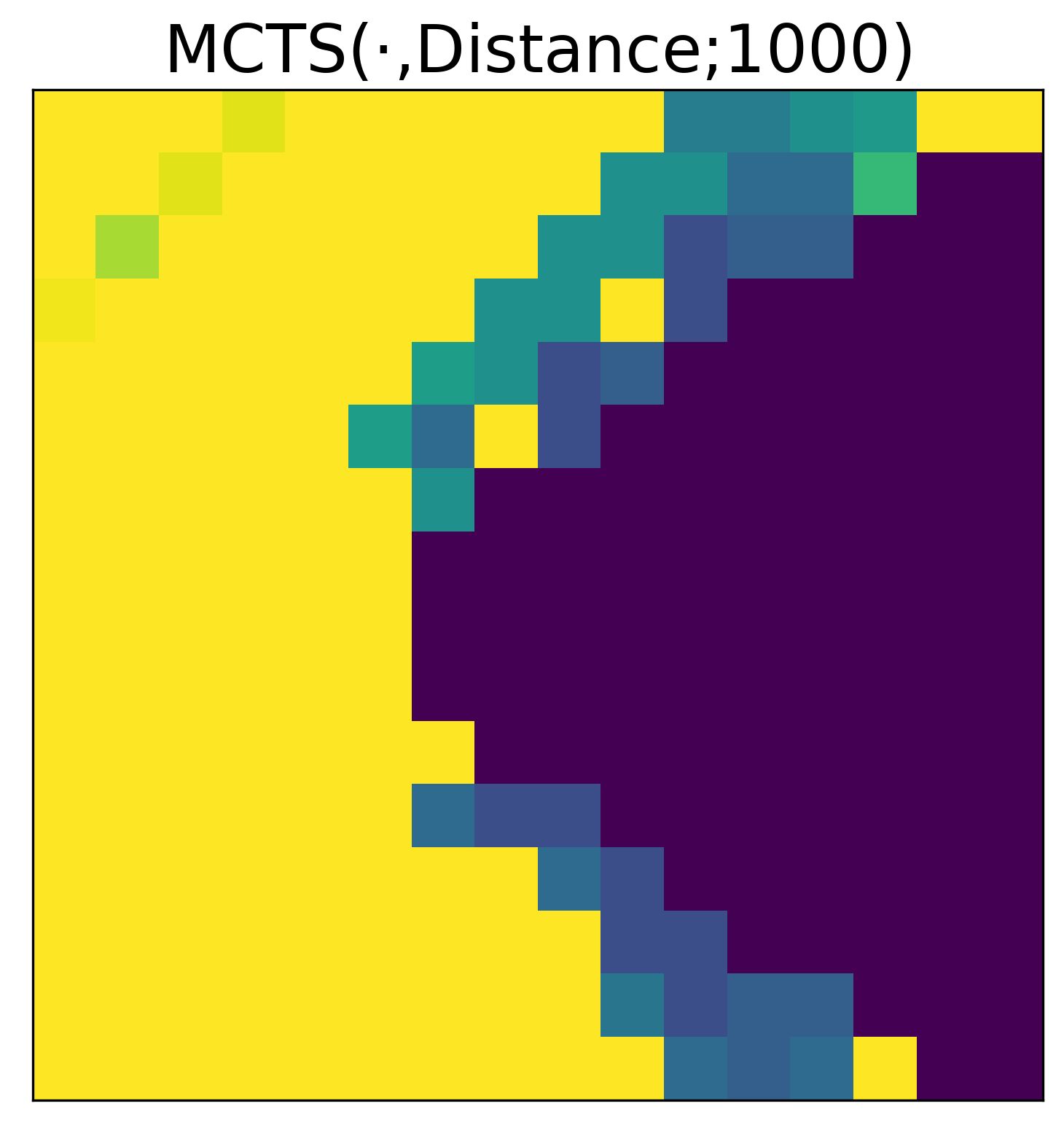}
\includegraphics[width=.3\textwidth]{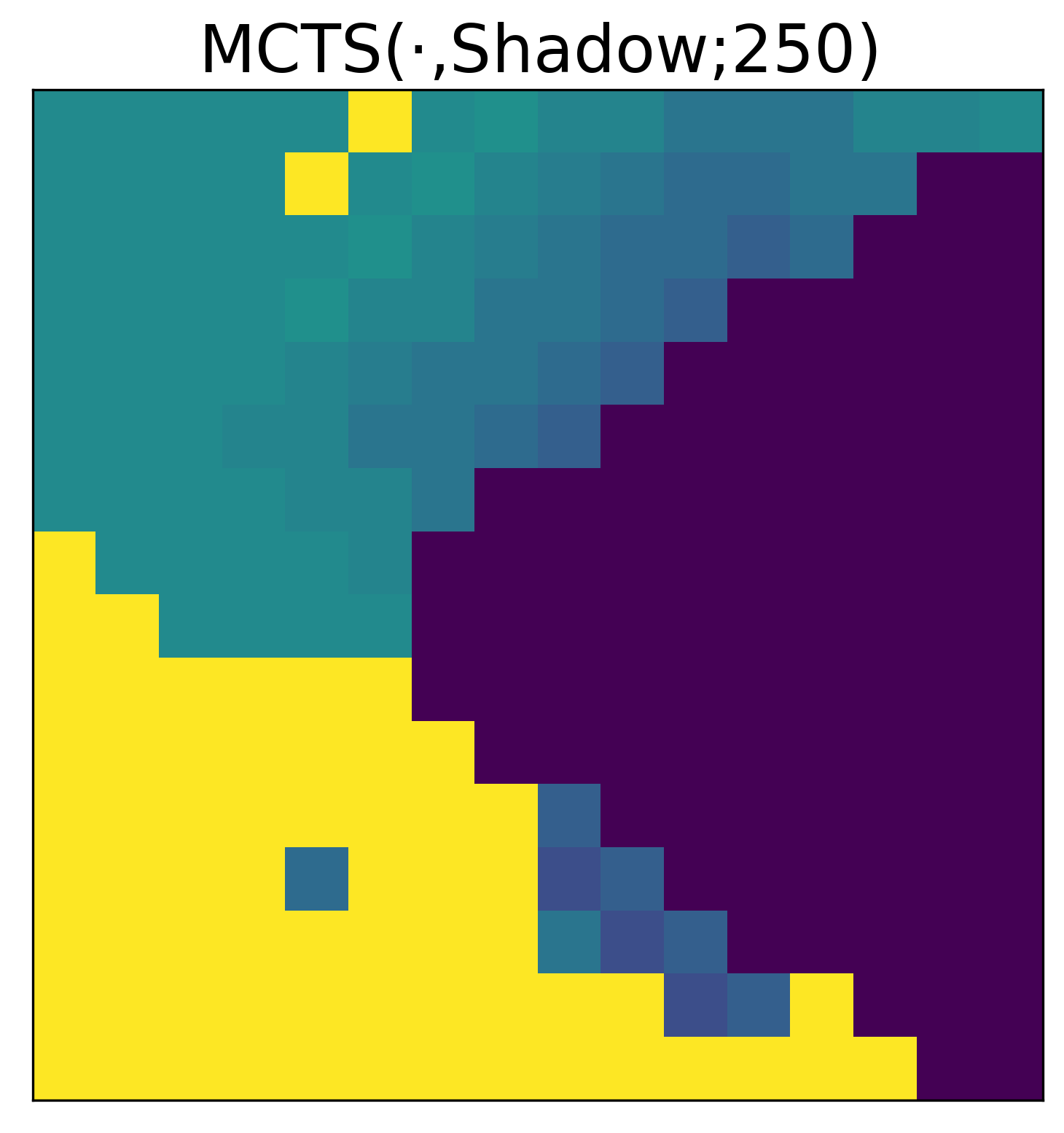}
\includegraphics[width=.3\textwidth]{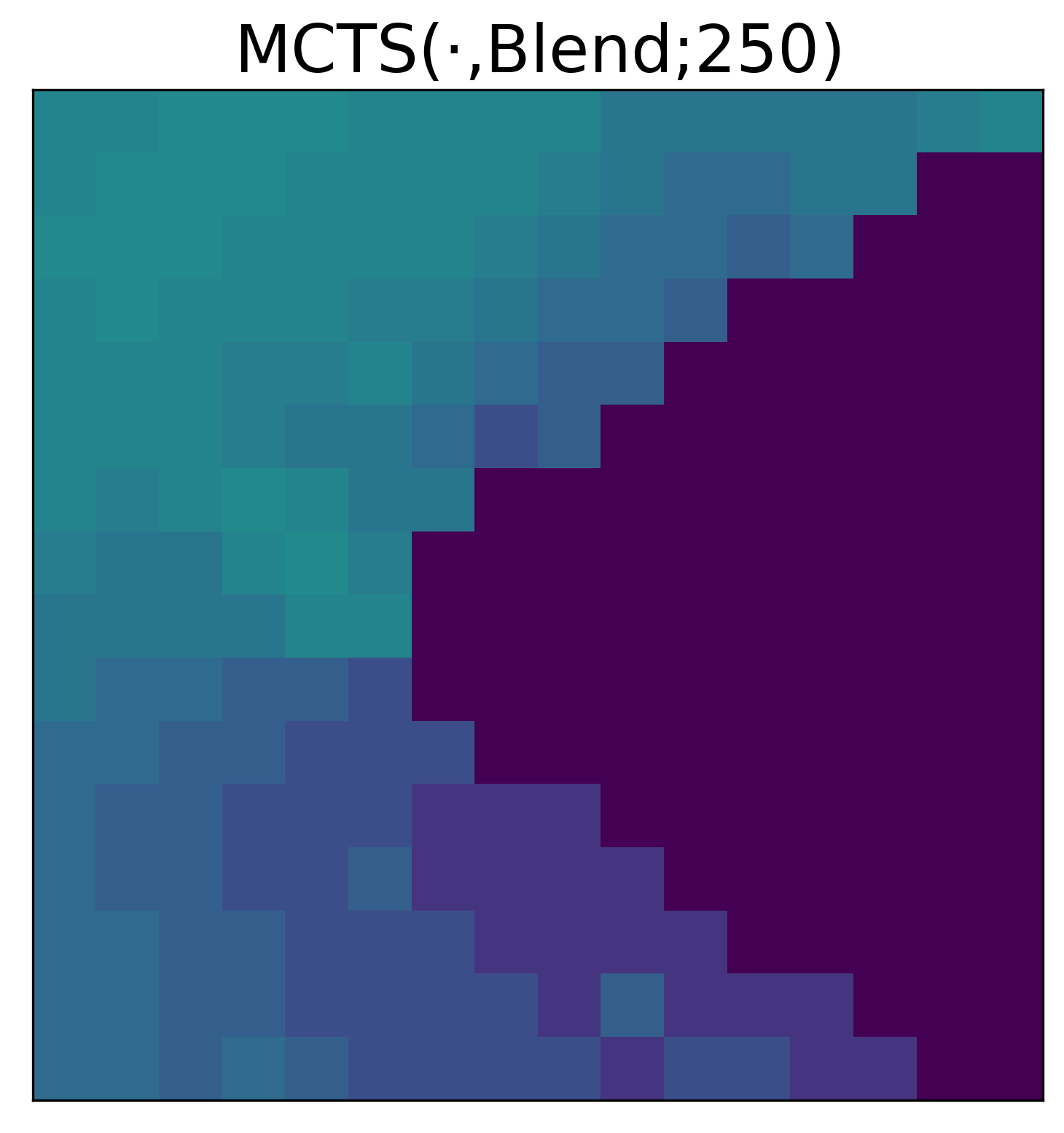}
\includegraphics[width=.3\textwidth]{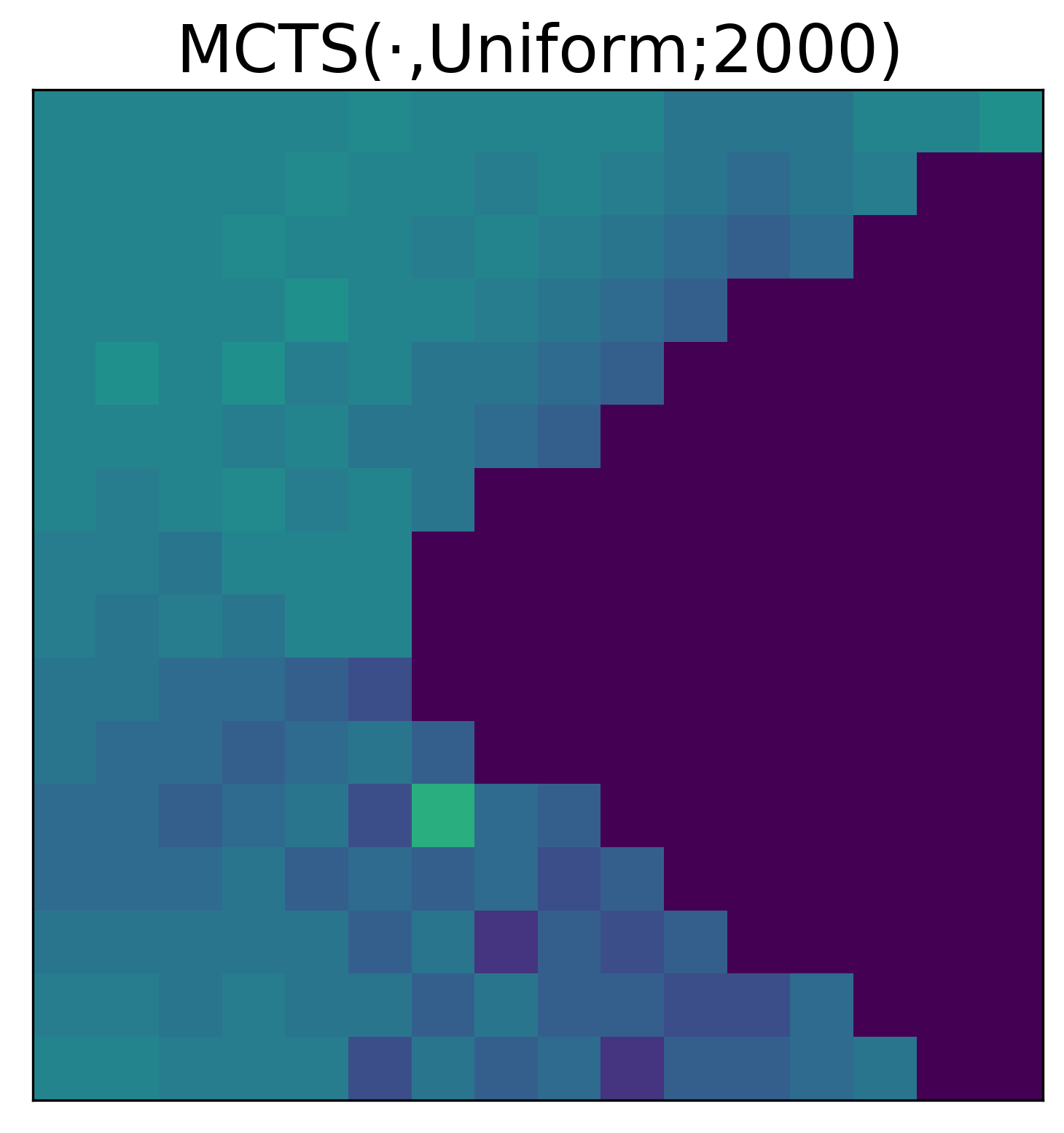}
\includegraphics[width=.3\textwidth]{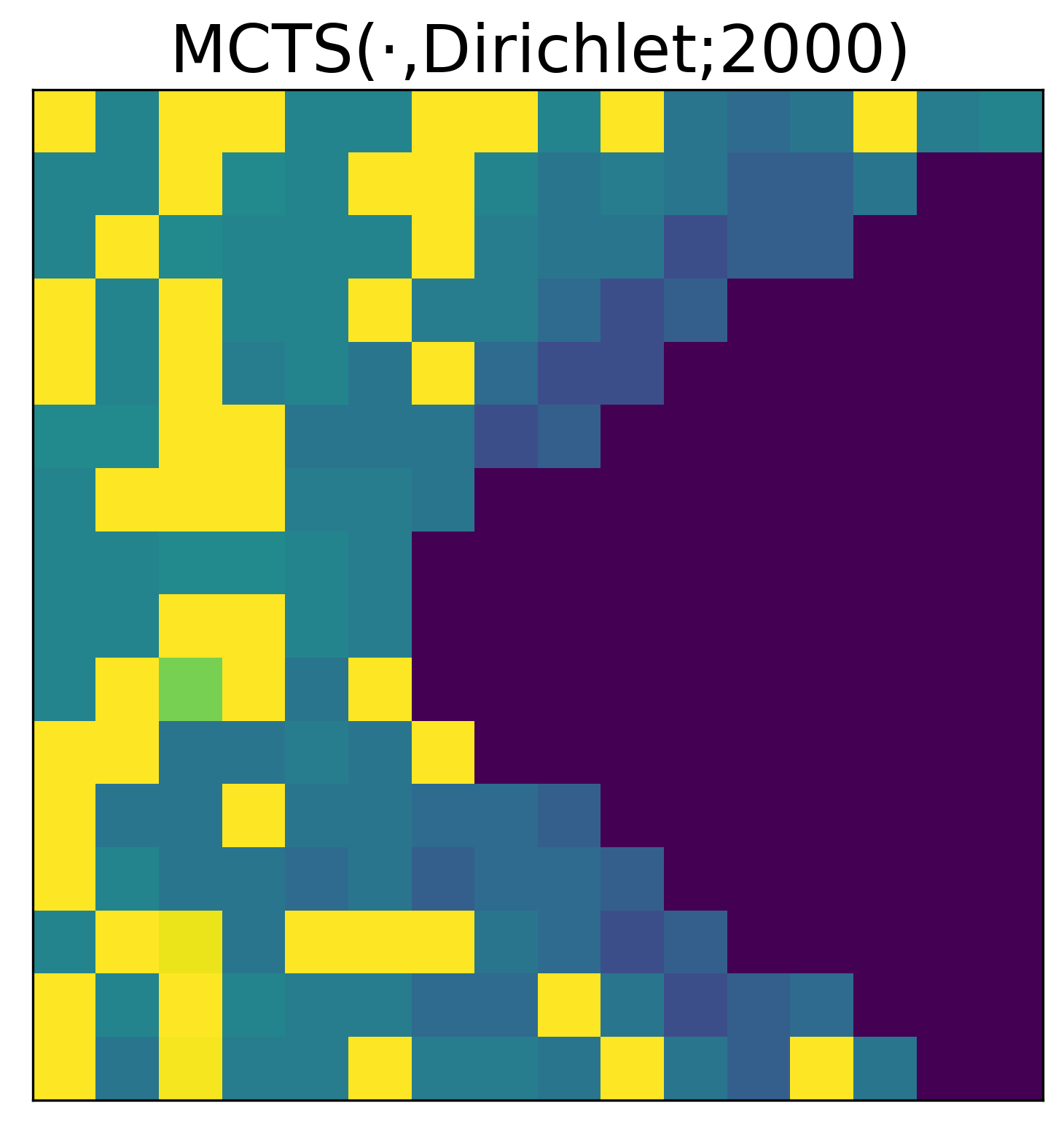}
\includegraphics[width=.3\textwidth]{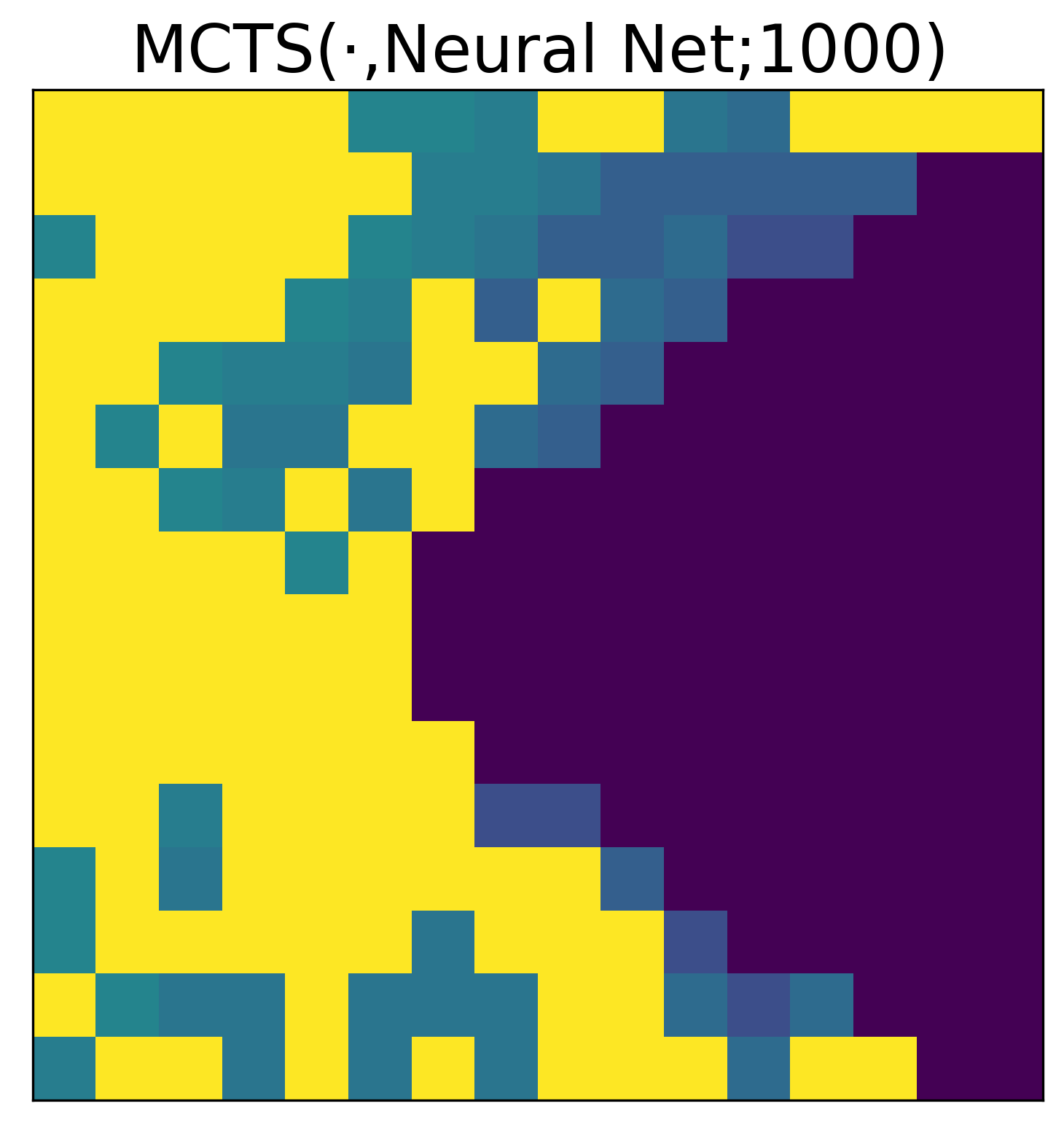}
\caption{One slice of the value function for 2 pursuer, 2 evader game on the circular obstacle.}
\label{fig:value-slice-2v2circle}
\end{figure}

Figure~\ref{fig:value-slice-2v2circle} shows a slice of the value function, where 3 players' positions are fixed,
and one evader's position varies.
Table~\ref{tab:2v2circle} shows the game statistics.
Here, we see some deviation from the baseline. As the number of players increase,
the number of actions increases. It is no longer sufficient to use random sampling.
The neural network is learning useful strategies to help guide the
Monte Carlo tree search to more significant paths. The distance and blend strategies
are effective by themselves. MCTS helps improve performance for Distance. However, 250 iterations is
not enough to search the action space, and actually lead to poor performance for Blend and Shadow.
For this game setup, MCTS(Distance,1000) performs the best with a 73.5\% win rate, followed by 
Blend with 65.4\% and Neural Net with 59.9\%. Although the trained network is not the best
in this case, the results are very promising. We want to emphasize that the model was trained with no prior
knowledge. Given enough offline time and resources, we believe the proposed approach can scale to larger grids and
learn more optimal policies than the local heuristics.

\begin{table}[]
\caption{Game statistics for the 2 pursuer vs 2 evader game with a circular obstacle.\vspace{1em}}
\label{tab:2v2circle}
\begin{tabular}{l|ccc}
Method          & Win \% (162 games) & Average game time                 \\   % & Average loss time \\
\hline
Distance                        & 56.8                     & 58.8        \\   % & 4.7             \\
Shadow                          & 46.3                     & 50.1        \\   % & 7.2             \\
Blend                           & 65.4                     & 67.9        \\   % & 7.3             \\
MCTS($\cdot$, Distance; 1000)    & 73.5                     & 76.6        \\   % & 12.2            \\
MCTS($\cdot$, Shadow; 250)       & 40.7                     & 44.5        \\   % & 6.3             \\
MCTS($\cdot$, Blend; 250)        & 00.0                     &  4.4        \\   % & 4.4             \\
MCTS($\cdot$, Uniform; 2000)     & 00.0                     &  5.3        \\   % & 5.3             \\
MCTS($\cdot$, Dirichlet; 2000)   & 27.8                     & 32.8        \\   % & 7.0             \\
MCTS($\cdot$, Neural Net; 1000)  & 59.9                     & 61.7        \\   % & 4.7            
\end{tabular}
\end{table}

\begin{figure}[hptb]
      \centering
      \includegraphics[width=.45\textwidth,trim={0 0.15in 0 0},clip]{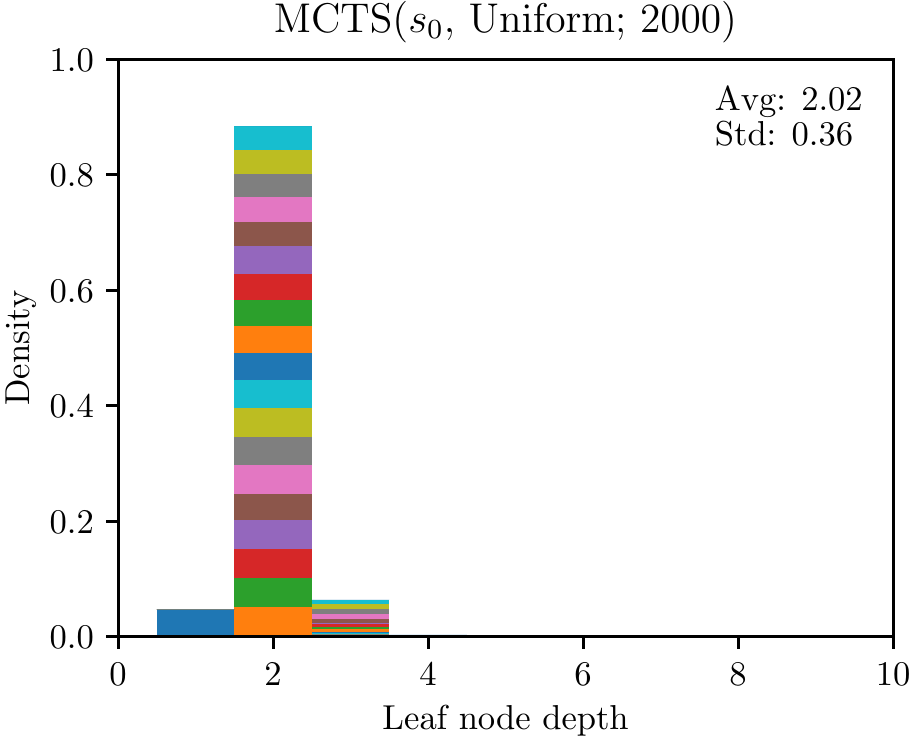} \quad
      \includegraphics[width=.45\textwidth,trim={0 0.15in 0 0},clip]{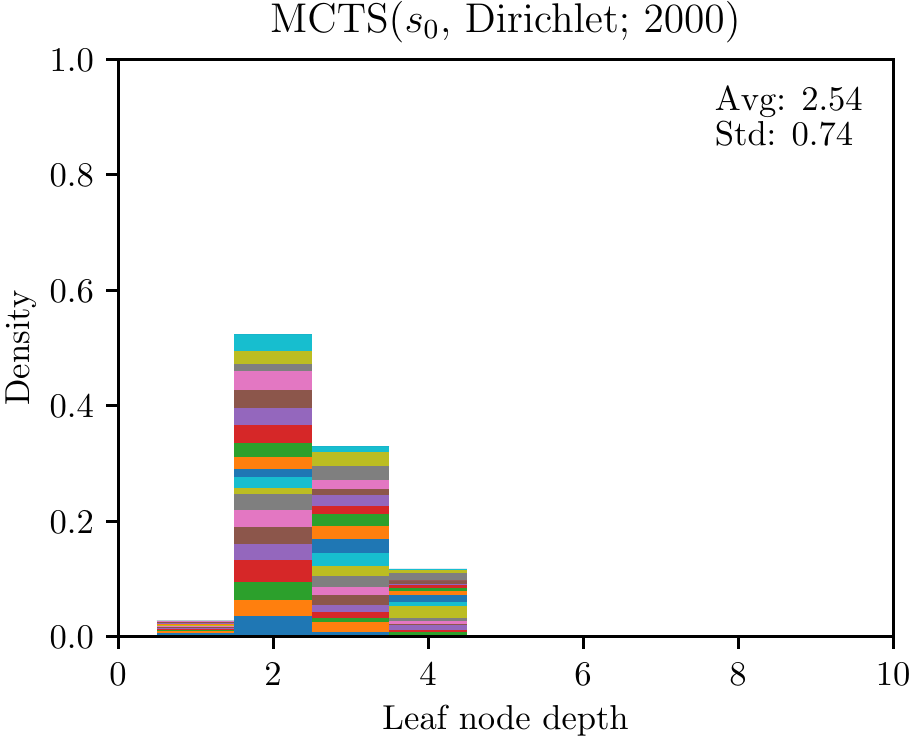}\\[.5em]
      \includegraphics[width=.45\textwidth,trim={0 0.15in 0 0},clip]{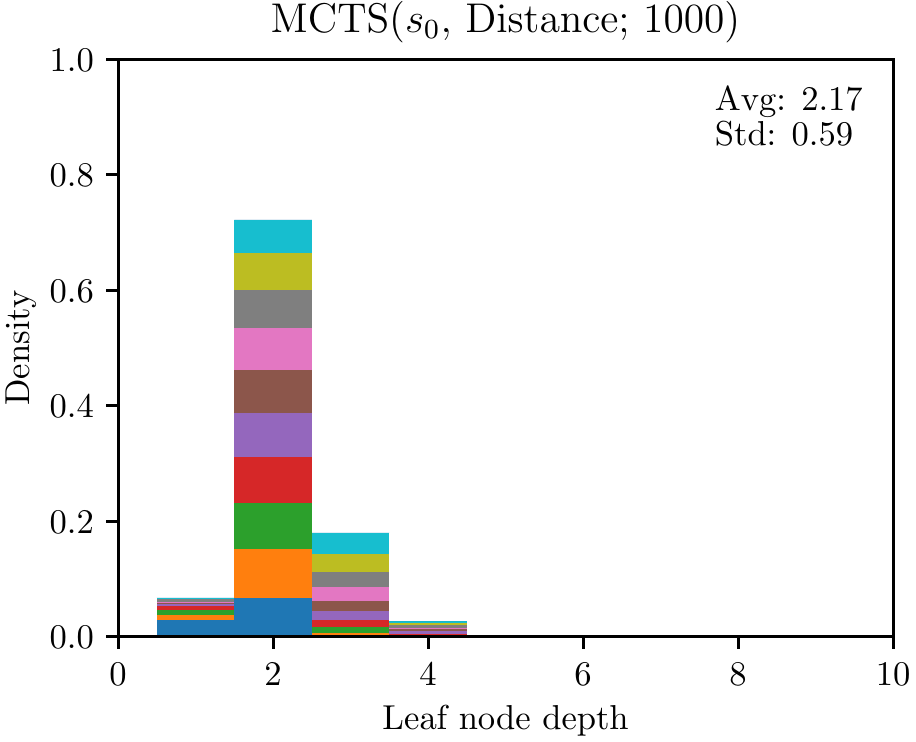} \quad
      \includegraphics[width=.45\textwidth,trim={0 0.15in 0 0},clip]{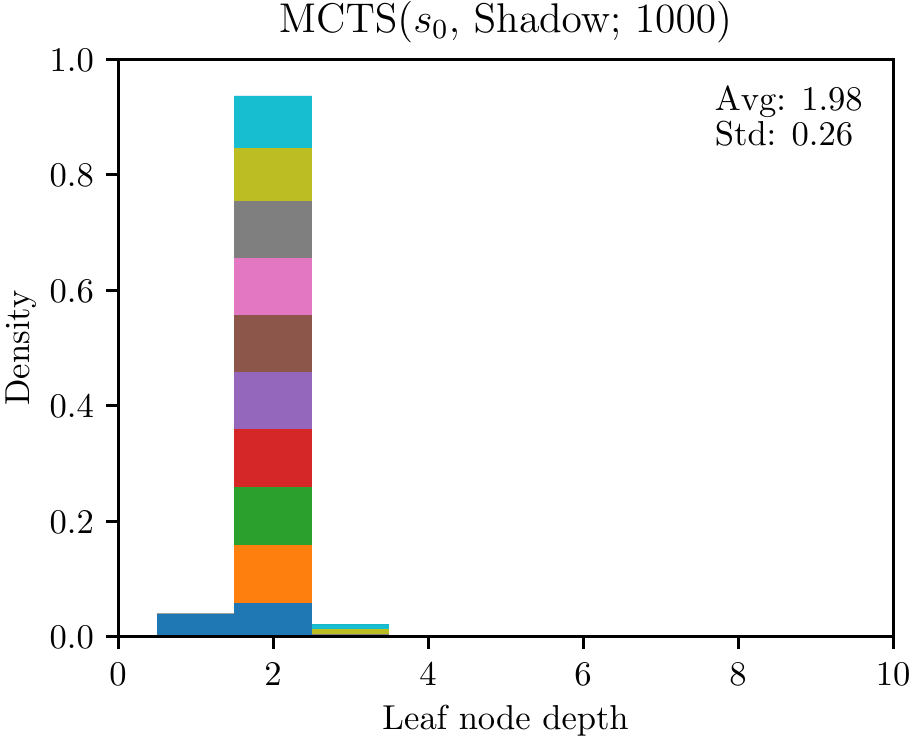} \\[.5em]
      \includegraphics[width=.45\textwidth,trim={0 0in 0 0},clip]{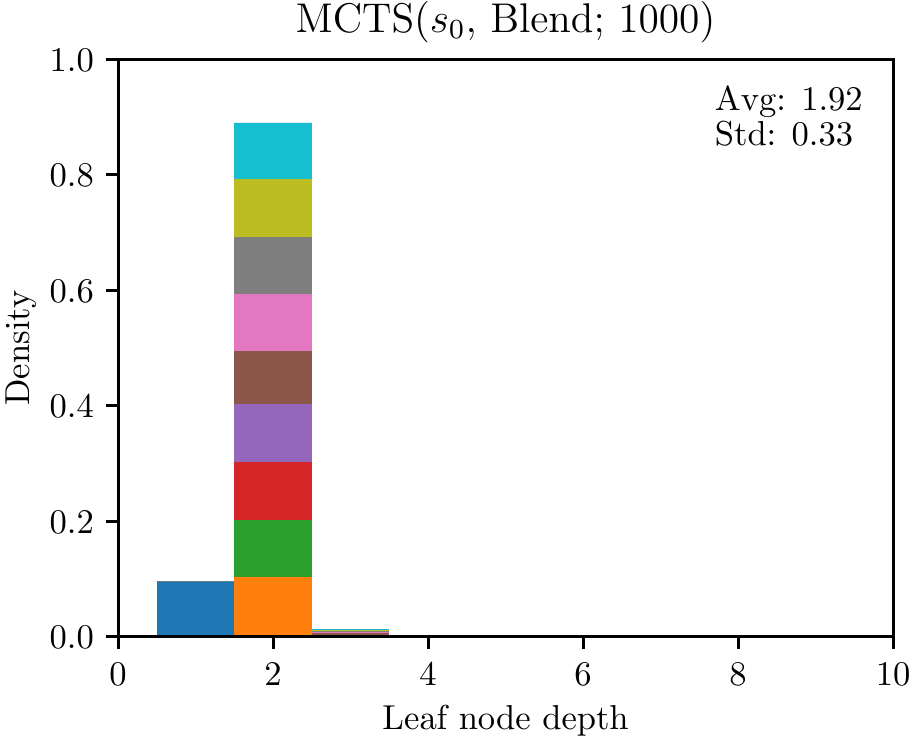} \quad
      \includegraphics[width=.45\textwidth,trim={0 0in 0 0},clip]{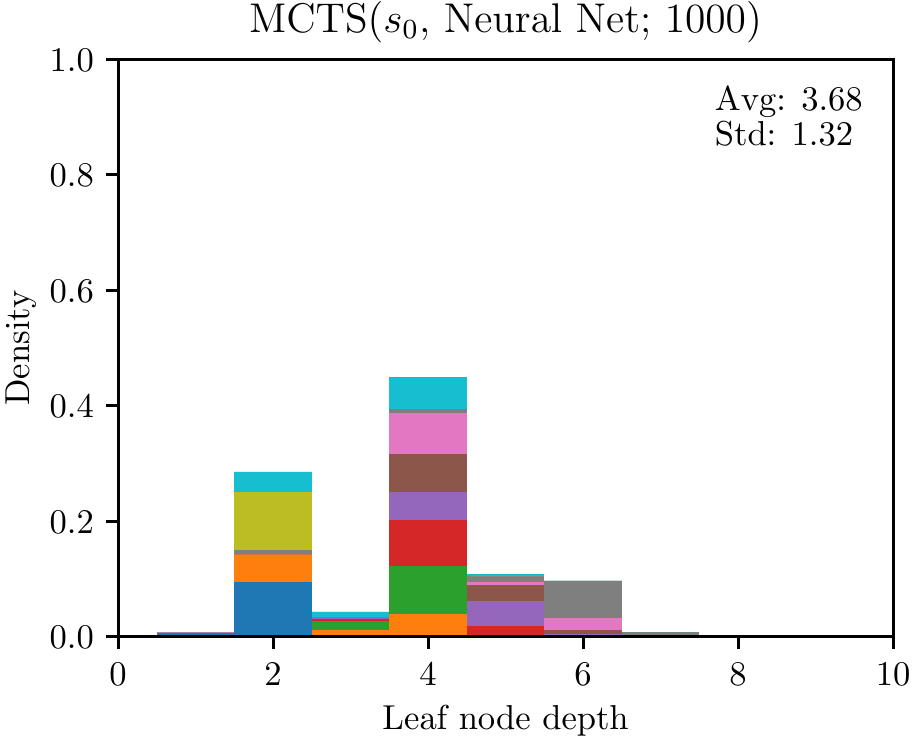} \quad

	  \caption{Histogram of leaf node depth for MCTS using various evaluator
functions for the multiplayer game around a circular obstacle.  The colors show
increments of 100 iterations. The multiplayer game has a much larger action
space, making tree search difficult. The neural network appears to search deeper
into the tree.} 

\label{fig:mcts-stats-multi}

\end{figure}

Figure~\ref{fig:mcts-stats-multi} shows a comparison of the depth of search for
$M=1000$ MCTS iterations.  Specifically, we report depth of each leaf node, as
measured by game time. To be fair, we allow the uniform and dirichlet baselines 
to run for 2000 MCTS iterations to match the runtime needed for 1 move.  Also,
the shadow strategies are extremely costly, and can only run 250 MCTS
iterations in the same amount of time. However, we show the statistics for
$M=1000$ to gain better insights.  Ideally, a good search would balance breadth
and depth.  The neural network appears to search further than the baselines. Of
course, this alone is not sufficient to draw any conclusions. For example, a
naive approach could be a depth-first search. 

\begin{figure}[hptb]
      \centering
      \includegraphics[width=.45\textwidth,trim={0 0.15in 0 0},clip]{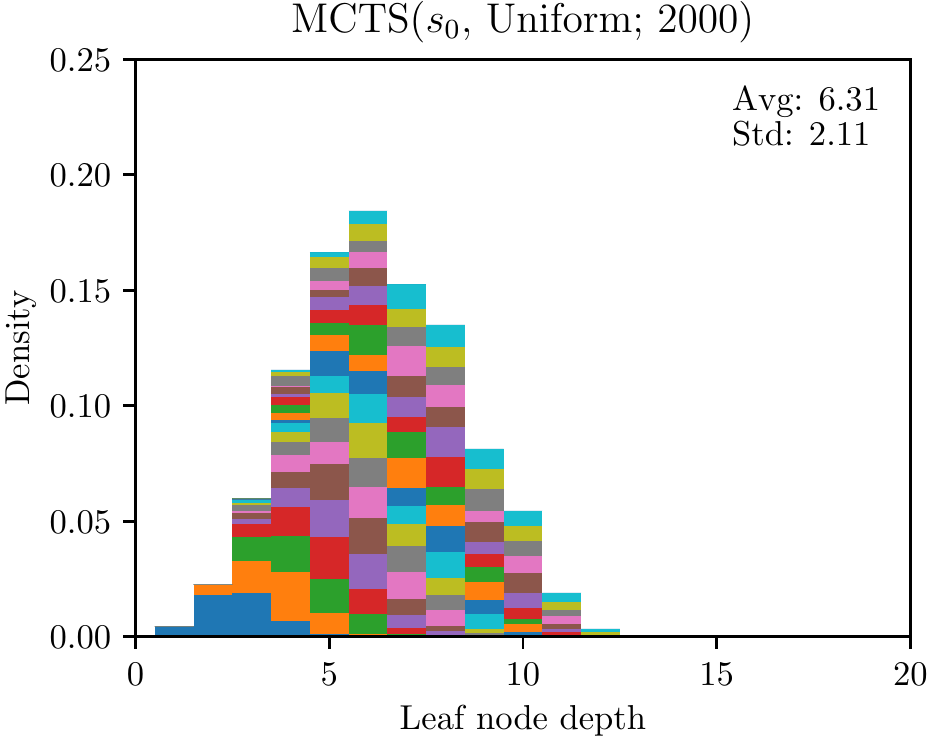} \quad
      \includegraphics[width=.45\textwidth,trim={0 0.15in 0 0},clip]{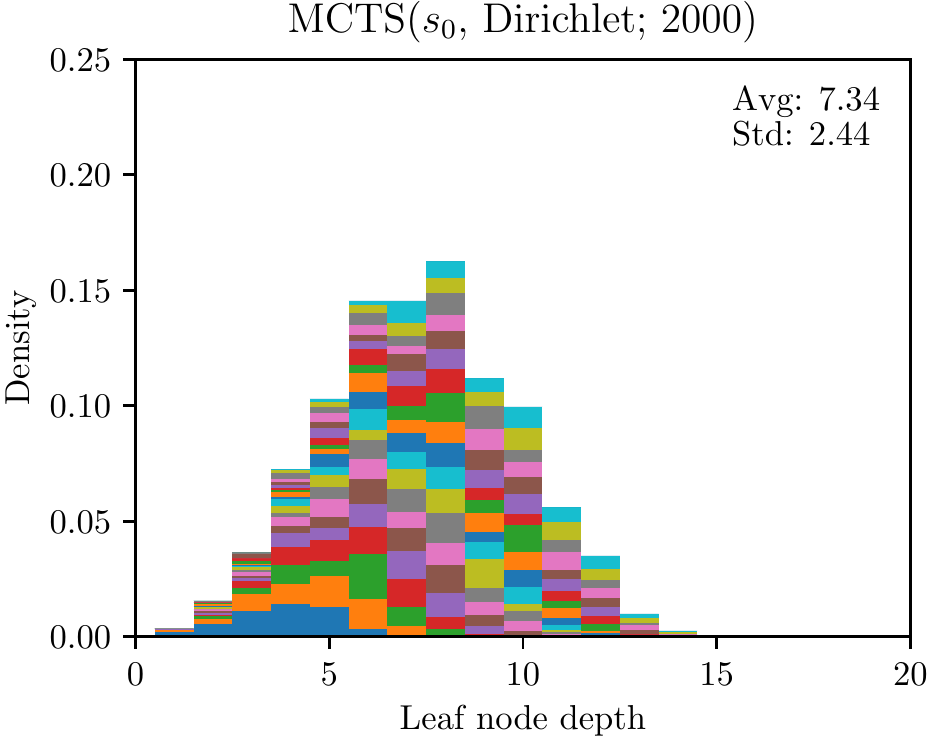}\\[.5em]
      \includegraphics[width=.45\textwidth,trim={0 0.15in 0 0},clip]{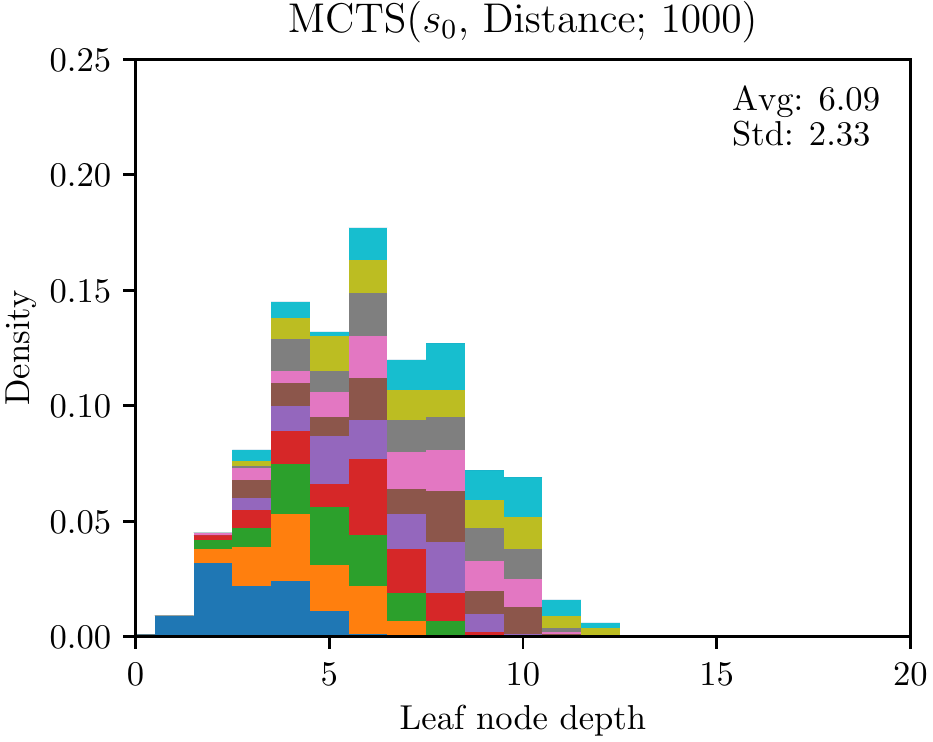} \quad
      \includegraphics[width=.45\textwidth,trim={0 0.15in 0 0},clip]{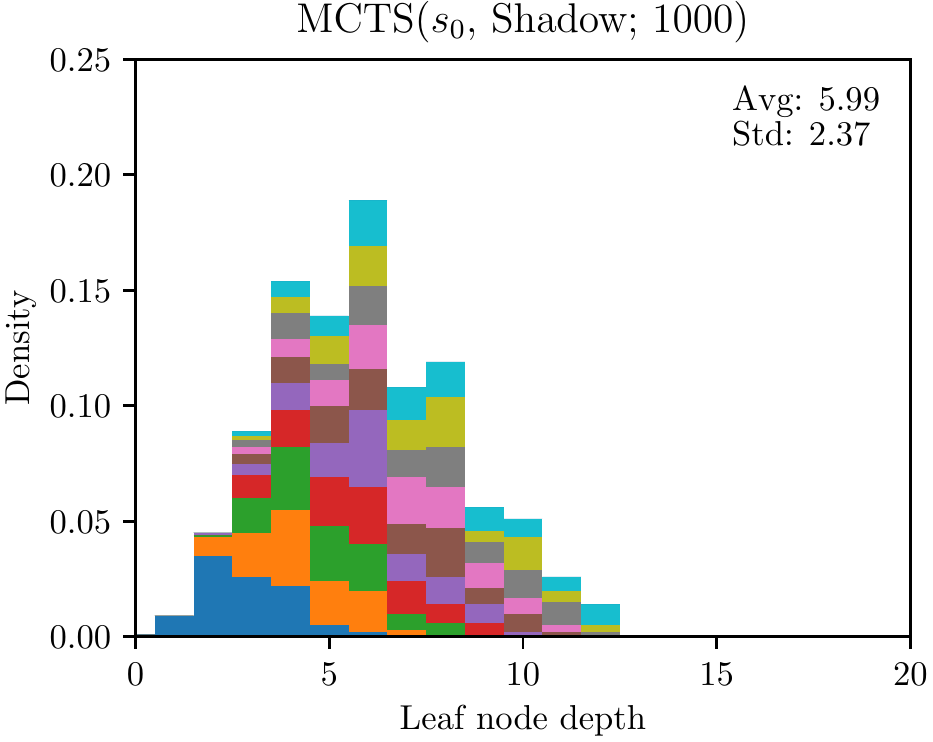} \\[.5em]
      \includegraphics[width=.45\textwidth,trim={0 0.00in 0 0},clip]{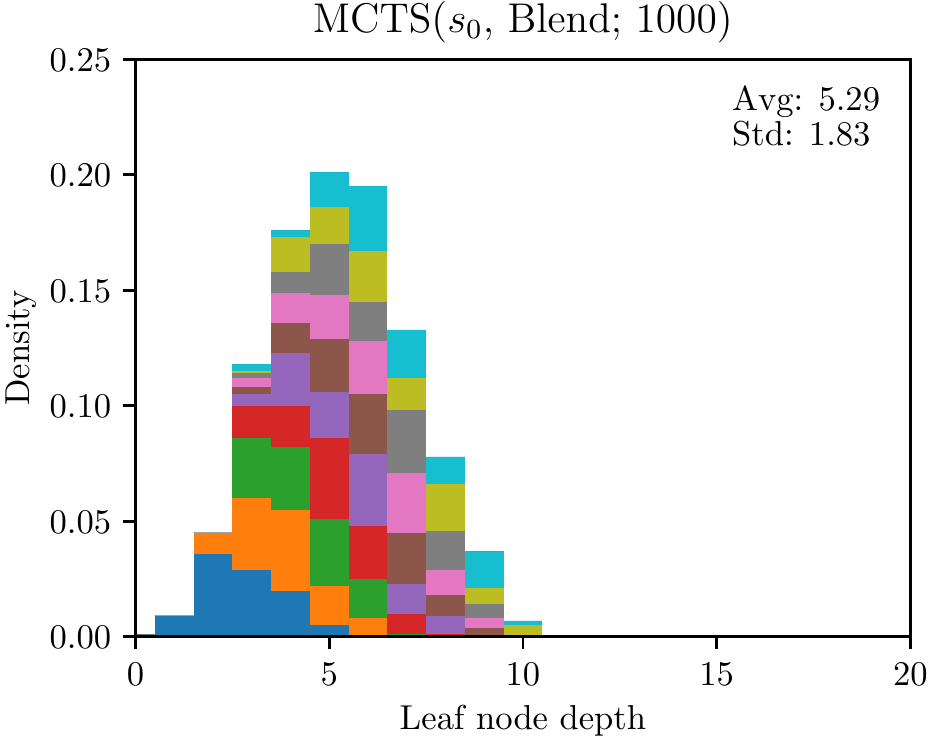} \quad
      \includegraphics[width=.45\textwidth,trim={0 0.00in 0 0},clip]{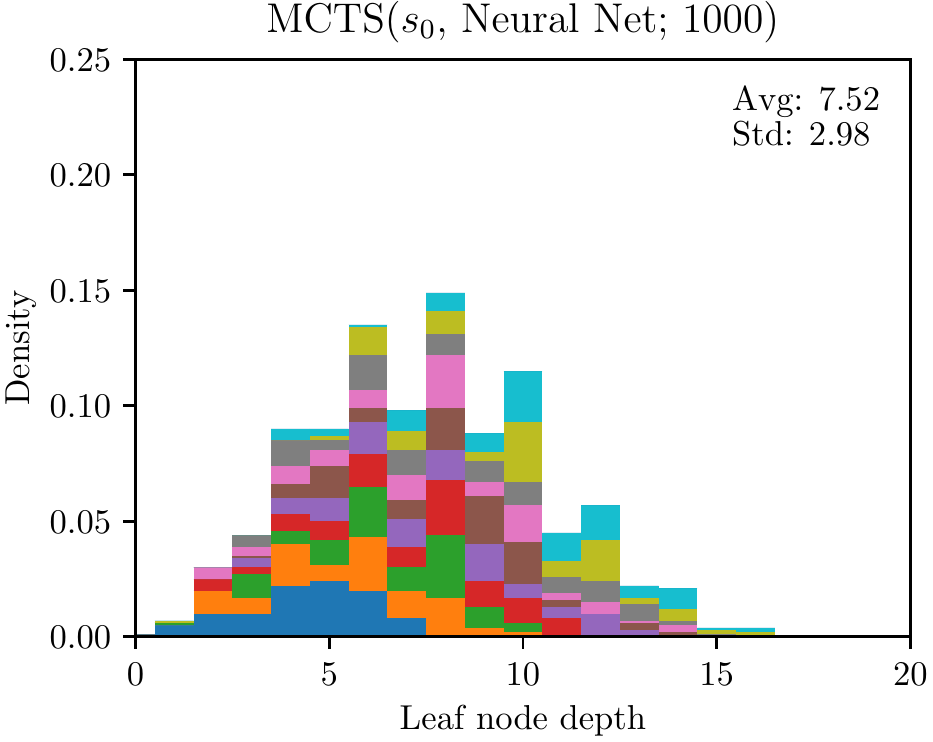} \quad

	  \caption{Histogram of leaf node depth for MCTS using various evaluator
functions for the single pursuer vs single evader game around a circular
obstacle.  The colors show increments of 100 iterations. The game is relatively easy
and thus all algorithms appear comparable. Note that Uniform and Dirichlet are allowed 2000 MCTS iterations,
since they require less overhead to run.} 

\label{fig:mcts-stats}
\end{figure}

In Figure~\ref{fig:mcts-stats}, we show a similar chart for the single pursuer,
single evader game with a circular obstacle. In this case, the game is
relatively easy, and all evaluator functions are comparable.

\section{Conclusion and future work}
We proposed three approaches for approximating optimal controls for the
surveillance-evasion game. When there are few players and the grid size is
small, one may compute the value function via the Hamilton-Jacobi-Isaacs
equations. The offline cost is immense, but on-line game play is very
efficient. The game can be played on the continuously in time and space, since
the controls can be interpolated from the value function. However, the value
function must be recomputed if the game settings, such as the obstacles or player
velocities, change.

When there are many players, we proposed locally optimal strategies for the
pursuer and evader. There is no offline preprocessing. All computation is done
on-line, though the computation does not scale well as the velocities or number
of pursuers increases. The game is discrete in time and space.

Lastly, we proposed a reinforcement learning approach for the 
multiplayer game. The offline training time can be enormous, but on-line game play
is very efficient and scales linearly with the number of players. The game is played
in discrete time and space, but the neural network model generalizes to
maps not seen during training. Given enough computational resources,
the neural network has the potential to approach the optimal controls
afforded by the HJI equations, while being more efficient than the local strategies.

\begin{figure}[hptb]
	\vspace{1em}
      \centering
      %\fbox{\includegraphics[width=.65\textwidth]{figures/3d_seg_crop.png}}
      \includegraphics[width=.65\textwidth]{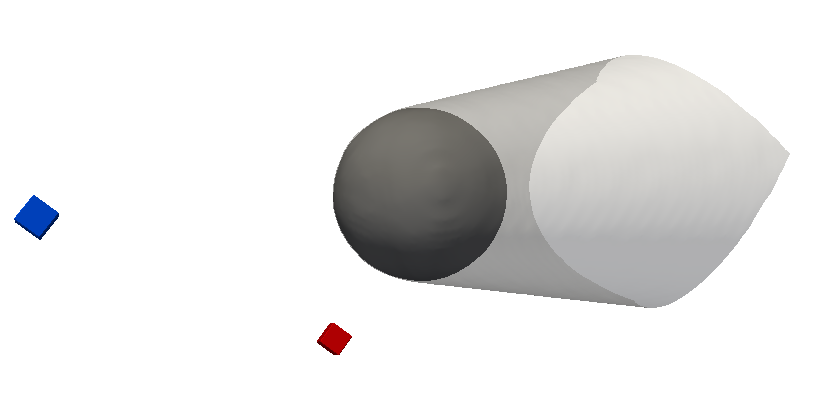}
	  \caption{A snapshot of a 3D surveillance-evasion game around a sphere.} 
\label{fig:3dseg}
\end{figure}

There are many avenues to explore for future research.  
We are working on the extension of our reinforcement learning approach to 3D,
which is straight-forward, but requires more computational resources.
Figure~\ref{fig:3dseg} shows an example surveillance-evasion game in 3D. Along
those lines, a multi-resolution scheme is imperative for scaling to higher
dimensions and resolutions.  One may also consider different game objectives,
such as seeking out an intially hidden evader, or allowing brief moments of
occlusion.

\section*{Acknowledgment}
This work was partially supported by NSF grant DMS-1913209.

%%%%%%%%%%%%%%%%%%%%%%%%%%%%%%%%%%%%%%%%%%%%%%%%%%%%%%%%%%%%%%%%%%%%%%
% Appendix/Appendices                                                %
%%%%%%%%%%%%%%%%%%%%%%%%%%%%%%%%%%%%%%%%%%%%%%%%%%%%%%%%%%%%%%%%%%%%%%
%
% If you have only one appendix, use the command \appendix instead
% of \appendices.
%

%\appendix
%\index{Appendix@\emph{Appendix}}%

%%%%%%%%%%%%%%%%%%%%%%%%%%%%%%%%%%%%%%%%%%%%%%%%%%%%%%%%%%%%%%%%%%%%%%
% Generate the bibliography.					     %
%%%%%%%%%%%%%%%%%%%%%%%%%%%%%%%%%%%%%%%%%%%%%%%%%%%%%%%%%%%%%%%%%%%%%%
%								     %
% NOTE: For master's theses and reports, NOTHING is permitted to     %
%	come between the bibliography and the vita. The command      %
%	to generate the index (if used) MUST be moved to before      %
%	this section.						     %
%								     %
%\nocite{*}      % This command causes all items in the 		     %
                % bibliographic database to be added to 	     %
                % the bibliography, even if they are not 	     %
                % explicitly cited in the text. 		     %
		%						     %
\bibliographystyle{plain}  % Here the bibliography 		     %
\bibliography{references}        % is inserted.			     %

\begin{thebibliography}{10}

\bibitem{bardi2008optimal}
Martino Bardi and Italo Capuzzo-Dolcetta.
\newblock {\em Optimal control and viscosity solutions of
  Hamilton-Jacobi-Bellman equations}.
\newblock Springer Science \& Business Media, 2008.

\bibitem{bardi1999numerical}
Martino Bardi, Maurizio Falcone, and Pierpaolo Soravia.
\newblock Numerical methods for pursuit-evasion games via viscosity solutions.
\newblock In {\em Stochastic and differential games}, pages 105--175. Springer,
  1999.

\bibitem{bharadwaj2018synthesis}
Suda Bharadwaj, Rayna Dimitrova, and Ufuk Topcu.
\newblock Synthesis of surveillance strategies via belief abstraction.
\newblock In {\em 2018 IEEE Conference on Decision and Control (CDC)}, pages
  4159--4166. IEEE, 2018.

\bibitem{bhattacharya2008approximation}
Sourabh Bhattacharya and Seth Hutchinson.
\newblock Approximation schemes for two-player pursuit evasion games with
  visibility constraints.
\newblock In {\em Robotics: Science and Systems}, 2008.

\bibitem{bhattacharya2009existence}
Sourabh Bhattacharya and Seth Hutchinson.
\newblock On the existence of nash equilibrium for a two player pursuit-evasion
  game with visibility constraints.
\newblock In {\em Algorithmic Foundation of Robotics VIII}, pages 251--265.
  Springer, 2009.

\bibitem{bhattacharya2011cell}
Sourabh Bhattacharya and Seth Hutchinson.
\newblock A cell decomposition approach to visibility-based pursuit evasion
  among obstacles.
\newblock {\em The International Journal of Robotics Research},
  30(14):1709--1727, 2011.

\bibitem{cartee2019time}
Elliot {Cartee}, Lexiao {Lai}, Qianli {Song}, and Alexander {Vladimirsky}.
\newblock Time-dependent surveillance-evasion games.
\newblock In {\em 2019 IEEE 58th Conference on Decision and Control (CDC)},
  pages 7128--7133, 2019.

\bibitem{chow2017algorithm}
Yat~Tin Chow, J{\'e}r{\^o}me Darbon, Stanley Osher, and Wotao Yin.
\newblock Algorithm for overcoming the curse of dimensionality for
  time-dependent non-convex hamilton--jacobi equations arising from optimal
  control and differential games problems.
\newblock {\em Journal of Scientific Computing}, 73(2-3):617--643, 2017.

\bibitem{chow2018algorithm}
Yat~Tin Chow, J{\'e}r{\^o}me Darbon, Stanley Osher, and Wotao Yin.
\newblock Algorithm for overcoming the curse of dimensionality for certain
  non-convex hamilton--jacobi equations, projections and differential games.
\newblock {\em Annals of Mathematical Sciences and Applications},
  3(2):369--403, 2018.

\bibitem{chow2019algorithm}
Yat~Tin Chow, J{\'e}r{\^o}me Darbon, Stanley Osher, and Wotao Yin.
\newblock Algorithm for overcoming the curse of dimensionality for
  state-dependent hamilton-jacobi equations.
\newblock {\em Journal of Computational Physics}, 387:376--409, 2019.

\bibitem{crandall1983viscosity}
Michael~G Crandall and Pierre-Louis Lions.
\newblock Viscosity solutions of hamilton-jacobi equations.
\newblock {\em Transactions of the American mathematical society},
  277(1):1--42, 1983.

\bibitem{evans1984differential}
Lawrence~C Evans and Panagiotis~E Souganidis.
\newblock Differential games and representation formulas for solutions of
  hamilton-jacobi-isaacs equations.
\newblock {\em Indiana University mathematics journal}, 33(5):773--797, 1984.

\bibitem{gilles2019evasive}
Marc~Aur{\`e}le Gilles and Alexander Vladimirsky.
\newblock Evasive path planning under surveillance uncertainty.
\newblock {\em Dynamic Games and Applications}, pages 1--26, 2019.

\bibitem{guibas1997visibility}
Leonidas~J Guibas, Jean-Claude Latombe, Steven~M LaValle, David Lin, and Rajeev
  Motwani.
\newblock Visibility-based pursuit-evasion in a polygonal environment.
\newblock In {\em Workshop on Algorithms and Data Structures}, pages 17--30.
  Springer, 1997.

\bibitem{ho1965differential}
Yu-Chi Ho, Arthur Bryson, and Sheldon Baron.
\newblock Differential games and optimal pursuit-evasion strategies.
\newblock {\em IEEE Transactions on Automatic Control}, 10(4):385--389, 1965.

\bibitem{isaacs1965differential}
Rufus Isaacs.
\newblock {\em Differential games}.
\newblock John Wiley and Sons, 1965.

\bibitem{kang2017mitigating}
Wei Kang and Lucas~C Wilcox.
\newblock Mitigating the curse of dimensionality: sparse grid characteristics
  method for optimal feedback control and hjb equations.
\newblock {\em Computational Optimization and Applications}, 68(2):289--315,
  2017.

\bibitem{karnad2009lion}
Nikhil Karnad and Volkan Isler.
\newblock Lion and man game in the presence of a circular obstacle.
\newblock In {\em 2009 IEEE/RSJ International Conference on Intelligent Robots
  and Systems}, pages 5045--5050. IEEE, 2009.

\bibitem{kingma2014adam}
Diederik~P Kingma and Jimmy Ba.
\newblock Adam: A method for stochastic optimization.
\newblock {\em arXiv preprint arXiv:1412.6980}, 2014.

\bibitem{lavalle1997motion}
Steven~M LaValle, Hector~H Gonz{\'a}lez-Banos, Craig Becker, and J-C Latombe.
\newblock Motion strategies for maintaining visibility of a moving target.
\newblock In {\em Proceedings of International Conference on Robotics and
  Automation}, volume~1, pages 731--736. IEEE, 1997.

\bibitem{lavalle2001visibility}
Steven~M LaValle and John~E Hinrichsen.
\newblock Visibility-based pursuit-evasion: The case of curved environments.
\newblock {\em IEEE Transactions on Robotics and Automation}, 17(2):196--202,
  2001.

\bibitem{lewin1975surveillance}
Joseph Lewin and John Breakwell.
\newblock The surveillance-evasion game of degree.
\newblock {\em Journal of Optimization Theory and Applications},
  16(3-4):339--353, 1975.

\bibitem{merz1974homicidal}
Antony~W Merz.
\newblock The homicidal chauffeur.
\newblock {\em AIAA Journal}, 12(3):259--260, 1974.

\bibitem{osher2006level}
Stanley Osher and Ronald Fedkiw.
\newblock {\em Level set methods and dynamic implicit surfaces}, volume 153.
\newblock Springer Science \& Business Media, 2006.

\bibitem{osher1988fronts}
Stanley Osher and James~Albert Sethian.
\newblock Fronts propagating with curvature-dependent speed: algorithms based
  on hamilton-jacobi formulations.
\newblock {\em Journal of computational physics}, 79(1):12--49, 1988.

\bibitem{ronneberger2015u}
Olaf Ronneberger, Philipp Fischer, and Thomas Brox.
\newblock U-net: Convolutional networks for biomedical image segmentation.
\newblock In {\em International Conference on Medical image computing and
  computer-assisted intervention}, pages 234--241. Springer, 2015.

\bibitem{rosin2011multi}
Christopher~D Rosin.
\newblock Multi-armed bandits with episode context.
\newblock {\em Annals of Mathematics and Artificial Intelligence},
  61(3):203--230, 2011.

\bibitem{sachs2004visibility}
Shai Sachs, Steven~M LaValle, and Stjepan Rajko.
\newblock Visibility-based pursuit-evasion in an unknown planar environment.
\newblock {\em The International Journal of Robotics Research}, 23(1):3--26,
  2004.

\bibitem{sethian1999level}
James~Albert Sethian.
\newblock {\em Level set methods and fast marching methods: evolving interfaces
  in computational geometry, fluid mechanics, computer vision, and materials
  science}, volume~3.
\newblock Cambridge university press, 1999.

\bibitem{sgall2001solution}
Ji{\v{r}}{\'\i} Sgall.
\newblock Solution of david gale's lion and man problem.
\newblock {\em Theoretical Computer Science}, 259(1-2):663--670, 2001.

\bibitem{silver2017mastering2}
David Silver, Thomas Hubert, Julian Schrittwieser, Ioannis Antonoglou, Matthew
  Lai, Arthur Guez, Marc Lanctot, Laurent Sifre, Dharshan Kumaran, Thore
  Graepel, et~al.
\newblock Mastering chess and shogi by self-play with a general reinforcement
  learning algorithm.
\newblock {\em arXiv preprint arXiv:1712.01815}, 2017.

\bibitem{silver2017mastering}
David Silver, Julian Schrittwieser, Karen Simonyan, Ioannis Antonoglou, Aja
  Huang, Arthur Guez, Thomas Hubert, Lucas Baker, Matthew Lai, Adrian Bolton,
  et~al.
\newblock Mastering the game of go without human knowledge.
\newblock {\em Nature}, 550(7676):354, 2017.

\bibitem{stiffler2014complete}
Nicholas~M Stiffler and Jason~M O'Kane.
\newblock A complete algorithm for visibility-based pursuit-evasion with
  multiple pursuers.
\newblock In {\em 2014 IEEE International Conference on Robotics and Automation
  (ICRA)}, pages 1660--1667. IEEE, 2014.

\bibitem{suzuki1992searching}
Ichiro Suzuki and Masafumi Yamashita.
\newblock Searching for a mobile intruder in a polygonal region.
\newblock {\em SIAM Journal on computing}, 21(5):863--888, 1992.

\bibitem{takei2015optimal}
Ryo Takei, Weiyan Chen, Zachary Clawson, Slav Kirov, and Alexander Vladimirsky.
\newblock Optimal control with budget constraints and resets.
\newblock {\em SIAM Journal on Control and Optimization}, 53(2):712--744, 2015.

\bibitem{takei2014efficient}
Ryo Takei, Richard Tsai, Zhengyuan Zhou, and Yanina Landa.
\newblock An efficient algorithm for a visibility-based surveillance-evasion
  game.
\newblock {\em Communications in Mathematical Sciences}, 12(7):1303--1327,
  2014.

\bibitem{tovar2008visibility}
Benjam{\'\i}n Tovar and Steven~M LaValle.
\newblock Visibility-based pursuit—evasion with bounded speed.
\newblock {\em The International Journal of Robotics Research},
  27(11-12):1350--1360, 2008.

\bibitem{tsai2002rapid}
Yen-Hsi~Richard Tsai.
\newblock Rapid and accurate computation of the distance function using grids.
\newblock {\em J. Comput. Phys.}, 2002.

\bibitem{tsai2004visibility}
Yen-Hsi~Richard Tsai, Li-Tien Cheng, Stanley Osher, Paul Burchard, and
  Guillermo Sapiro.
\newblock Visibility and its dynamics in a pde based implicit framework.
\newblock {\em Journal of Computational Physics}, 199(1):260--290, 2004.

\bibitem{tsai2003fast}
Yen-Hsi~Richard Tsai, Li-Tien Cheng, Stanley Osher, and Hong-Kai Zhao.
\newblock Fast sweeping algorithms for a class of hamilton--jacobi equations.
\newblock {\em SIAM journal on numerical analysis}, 41(2):673--694, 2003.

\bibitem{tsitsiklis1995efficient}
John~N Tsitsiklis.
\newblock Efficient algorithms for globally optimal trajectories.
\newblock {\em IEEE Transactions on Automatic Control}, 40(9):1528--1538, 1995.

\bibitem{zhang2016multi}
Mengzhe Zhang and Sourabh Bhattacharya.
\newblock Multi-agent visibility-based target tracking game.
\newblock In {\em Distributed Autonomous Robotic Systems}, pages 271--284.
  Springer, 2016.

\bibitem{zou2016visibility}
Rui Zou and Sourabh Bhattacharya.
\newblock Visibility-based finite-horizon target tracking game.
\newblock {\em IEEE Robotics and Automation Letters}, 1(1):399--406, 2016.

\bibitem{zou2018optimal}
Rui Zou and Sourabh Bhattacharya.
\newblock On optimal pursuit trajectories for visibility-based target-tracking
  game.
\newblock {\em IEEE Transactions on Robotics}, 35(2):449--465, 2018.

\bibitem{zou2016optimal}
Rui Zou, Hamid Emadi, and Sourabh Bhattacharya.
\newblock On the optimal policies for visibility-based target tracking.
\newblock {\em arXiv preprint arXiv:1611.04613}, 2016.

\end{thebibliography}
\index{Bibliography@\emph{Bibliography}}%			     %
%%%%%%%%%%%%%%%%%%%%%%%%%%%%%%%%%%%%%%%%%%%%%%%%%%%%%%%%%%%%%%%%%%%%%%

%%%%%%%%%%%%%%%%%%%%%%%%%%%%%%%%%%%%%%%%%%%%%%%%%%%%%%%%%%%%%%%%%%%%%%
% Generate the index.						     %
%%%%%%%%%%%%%%%%%%%%%%%%%%%%%%%%%%%%%%%%%%%%%%%%%%%%%%%%%%%%%%%%%%%%%%
%								     %
% NOTE: For master's theses and reports, NOTHING is permitted to     %
%	come between the bibliography and the vita. This section     %
%	to generate the index (if used) MUST be moved to before      %
%	the bibliography section.				     %
%								     %
%\printindex%    % Include the index here. Comment out this line      %
%		% with a percent sign if you do not want an index.   %
%%%%%%%%%%%%%%%%%%%%%%%%%%%%%%%%%%%%%%%%%%%%%%%%%%%%%%%%%%%%%%%%%%%%%%

%%%%%%%%%%%%%%%%%%%%%%%%%%%%%%%%%%%%%%%%%%%%%%%%%%%%%%%%%%%%%%%%%%%%%%
% Vita page.							     %
%%%%%%%%%%%%%%%%%%%%%%%%%%%%%%%%%%%%%%%%%%%%%%%%%%%%%%%%%%%%%%%%%%%%%%

\end{document}